\documentclass[twoside]{article}

\usepackage[normalem]{ulem}

\usepackage[accepted]{aistats2024}
\usepackage[round]{natbib}

\usepackage{amsmath, amsthm, amstext, amsfonts, amssymb, mathrsfs}
\newtheorem{theorem}{Theorem}
\newtheorem{assumption}{Assumption}
\newtheorem{lemma}{Lemma}
\newtheorem{proposition}{Proposition}
\newtheorem{definition}{Definition}

\usepackage{thmtools}

\usepackage{algorithm, algorithmic}
\usepackage{graphicx, subfigure}
\usepackage{xcolor}
\usepackage{hyperref}
\usepackage[capitalize]{cleveref}
\usepackage{wrapfig}

\Crefname{thm}{Theorem}{Theorems}
\begin{document}

%

%

\twocolumn[

\aistatstitle{Provable Policy Gradient Methods for Average-Reward Markov Potential Games}

\aistatsauthor{ Min Cheng \And Ruida Zhou \And P. R. Kumar \And Chao Tian }

\aistatsaddress{ Texas A\&M University \And  Univeristy of California, LA \And Texas A\&M University \And Texas A\&M University } ]

\begin{abstract}
  We study Markov potential games under the infinite horizon average reward criterion. Most previous studies have been for discounted rewards. We prove that both algorithms based on independent policy gradient and independent natural policy gradient converge globally to a Nash equilibrium for the average reward criterion. To set the stage for gradient-based methods, we first establish that the average reward is a smooth function of policies and provide sensitivity bounds for the differential value functions, under certain conditions on ergodicity and the second largest eigenvalue of the underlying Markov decision process (MDP).  We prove that three algorithms, policy gradient, proximal-Q, and natural policy gradient (NPG),  converge to an $\epsilon$-Nash equilibrium with time complexity $O(\frac{1}{\epsilon^2})$, given a gradient/differential Q function oracle. When policy gradients have to be estimated, we propose an algorithm with $\tilde{O}(\frac{1}{\min_{s,a}\pi(a|s)\delta})$ sample complexity to achieve $\delta$ approximation error w.r.t~the $\ell_2$ norm. Equipped with the estimator, we derive the first sample complexity analysis for a policy gradient ascent algorithm, featuring a sample complexity of $\tilde{O}(1/\epsilon^5)$.
  Simulation studies are presented.
\end{abstract}

\section{INTRODUCTION}
Multi-agent reinforcement learning (MARL) \citep{busoniu2008comprehensive, zhang2021multi} features interactions among multiple agents, with each agent having its own objective and decision-making process. It finds applications in various domains, such as video games \citep{vinyals2019grandmaster, samvelyan2019starcraft}; robotics \citep{yang2004multiagent,perrusquia2021multi}; economics \citep{zheng2022ai}; and networked system control \citep{chu2020multi}.
 Unlike single-agent RL, interactions among agents create a dynamic and non-stationary environment, making the learning process more challenging. Under the criterion of discounted reward, theoretical investigations have examined Markov general-sum games \citep{song2021can}, zero-sum games \citep{NEURIPS2020_0cc6ee01}, and Markov potential games \citep{leonardos2021global, zhang2021gradient,ding2022independent}. 

In infinite-horizon tasks, it is more natural to use the average reward over the entire life-span for continuing tasks where optimizing stable, long-term performance becomes crucial, e.g., resource allocation in data centers, congestion games, and control problems \citep{xu2014reinforcement}. As shown in \citep{zhang2020average}, algorithms designed for discounted reward criterion can lead to unsatisfactory performance under the long-term average cost criterion. 
However, the average reward criterion which suits long-term strategic games remains largely unexplored. This paper delves into the challenges of employing the average reward criterion in the realm of MARL, specifically for Markov potential games.

Among different approaches to reinforcement learning, policy-based methods are appealing due to the ease of applying function approximation for large state and action spaces. However, most of the existing literature on average reward RL is either based on model-based methods \citep{auer2008near,azar2017minimax}, value-based methods \citep{wei2020model}, or based on reduction to discounted MDPs \citep{jin2021towards}, with relatively fewer works dedicated to exploring policy-based methods \citep{Li,wei2020model}. 
This paper examines policy-based methods in the context of average reward Markov potential games and demonstrates the convergence to a Nash policy, which is the primary goal of theoretical investigations in MARL \citep{leonardos2021global, ding2022independent, zhang2022global}.

\subsection{Contributions}

\begin{itemize}
    \item We address the problem of average reward Markov potential games and analyze three algorithms, policy gradient ascent, proximal-Q, and natural policy gradient. We show that with access to a gradient oracle, they converge to an $\epsilon$-Nash equilibrium with time complexity $O(\frac{1}{\epsilon^2})$.
    \item When the policy gradient has to be estimated, we propose a single-trajectory policy gradient estimator that estimates the policy gradient with $\tilde{O}(\frac{1}{\pi(a|s)\delta})$ sample complexity and $\delta$ approximation error w.r.t. the $\ell_2$ norm. We also provide the first sample complexity bound $\Tilde{O}(\frac{1}{\epsilon^5})$ for the projected policy gradient ascent algorithm.
    \item On the technical side, we rigorously show that the average reward is an $L$-smooth function of the policy under an ergodicity assumption. This is the first theoretical analysis of policy gradient for the average reward. We note that the concurrent work of \cite{bai2023regret} assumes $L$-smoothness without proving it. We also establish sensitivity bounds for differential value functions for general single-agent average reward MDPs, which play an important role in providing regret bound independent of the size of the action set (Section 5). These bounds can potentially be used in smoothness analysis of other parameterized policy parameter classes and function approximation analysis, under a further assumption of \cite[Assumption 1]{xu2020improving}.
\end{itemize}

\paragraph{Comparison with Previous Works} To obtain a sample complexity bound for a projected policy gradient algorithm, existing work \citep{leonardos2021global} attempts to establish such a bound for a policy projected from a true deterministic gradient, instead of one projected from a gradient estimated from samples (details in \cref{B.4}). The only work that analyzes the estimation error of policy gradient under the average reward setting that we are aware of is a recent work \citep{bai2023regret} which separately estimates two parts of policy gradient. Even though they have the same sample complexity, our algorithm calls the estimation algorithm $O(1/\lfloor\frac{1}{\delta} \rfloor)$ times less than theirs, thus reducing the computational burden.

\subsection{Related Works}
\paragraph{Markov Potential Games} originate from \cite{monderer1996potential}, who proposed static potential games. Later, \cite{dechert2006stochastic} addressed the problem of stochastic lake water usage modeling it as a Markov potential game with known transition probabilities. With the emergence of reinforcement learning, \cite{leonardos2021global} and \cite{zhang2021gradient} extended the Markov potential game to the unknown dynamics setting, where they analyzed the convergence to Nash equilibrium by extending the policy gradient techniques developed by \cite{agarwal2021theory} and \cite{mei2020global} to the multi-agent setting. Later \cite{ding2022independent} proposed a policy ascent algorithm, projecting from the Q-function instead of direct policy gradient. \cite{cen2022independent}, \cite{zhang2022global}, and \cite{sun2024provably} have studied the natural policy gradient algorithm in the static and Markov settings. Recently, variants of networked Markov potential games and $\alpha$-Markov potential games have been studied by \cite{zhou2023convergence} and \cite{guo2023markov}. However, all these results are restricted to the discounted reward setting.

\paragraph{Average Reward MDPs} date back to the classic results of \cite{how60,bla62,puter,kakade2001natural,sutton1999policy}. When system dynamics are known, average reward MDPs can be solved by linear programming, value iteration, or policy iteration \citep{how60,puter}. A survey can be found in \cite{dewanto2020average}. When the dynamics need to be learned, model-based methods like UCRL2 and UCBVI were proposed in \cite{auer2008near} and \cite{azar2017minimax}, and the reward-biased method originally proposed by \cite{kumar1982new} has been recently reexamined in \cite{mete2021reward}. For model-free algorithms, \cite{wei2020model} proposed optimistic mirror ascent, \cite{zhang2021finite} studied TD($\lambda$) and Q learning, and \cite{Li} analyzed the policy gradient methods under the mirror descent framework. \cite{zhang2023sharper} and \cite{jin2021towards} proposed algorithms utilizing a reduction to the discounted reward setting.

\section{PRELIMINARIES}

In this section, we introduce the average reward Markov potential game (AMPG), Nash equilibria, and differential value functions for the average reward MDP.

\subsection{Average Reward Markov Potential Games}

An $N$-agent infinite-horizon tabular average-reward Markov game (AMG) is represented by a tuple $\mathrm{AMG}(\mathcal{S},\{\mathcal{A}_i\}_{i=1}^N,P,\{r_i\}_{i=1}^N)$.
$\mathcal{S}$ denotes a finite state space, $\mathcal{A}_i$ is a finite action space for agent $i$, and $\mathcal{A}=\mathcal{A}_1\times \mathcal{A}_2\times ...\times \mathcal{A}_N$ is the joint action space for all agents. We denote by $S$, $A_1$, ...,$A_N$ their respective cardinalities. $P$ denotes the transition probabilities, i.e., $P(\cdot|s,\mathbf{a})\in\triangle(\mathcal{S})$ is the probability distribution of the next state under a joint action profile $\textbf{a}=(a_1,a_2,...,a_N) \in \mathcal{A}$ when the current state is $s$, and $\triangle(\mathcal{S})$ denotes the probability simplex over set $\mathcal{S}$. 
$r_i: \mathcal{S}\times\mathcal{A}\to [0,1]$ is the one-step reward function for agent $i$. 

A randomized stationary policy for an agent is defined by a map $\pi_i:\mathcal{S}\to\triangle(\mathcal{A}_i)$, i.e., $\pi_i(\cdot|s)\in\triangle(\mathcal{A}_i)$. 
Denote by $\Pi_i:=(\triangle(\mathcal{A}_i))^{\mathcal{S}}$ the set of all randomized policies for agent $i$. We use $\pi=\{\pi_i\}_{i=1}^N$ to represent the joint policy of all agents, and $\pi_{-i}=\{\pi_j\}_{j\neq i}$ to represent all policies but $i$. $\Pi=\Pi_1\times\Pi_2\times\hdots\times\Pi_N$ is the independent joint policy set. Similarly, we denote $a_{-i}:=\{a_j\}_{j\neq i}$ and $\Pi_{-i}:=\Pi_1\times\hdots\Pi_{i-1}\times\Pi_{i+1}\hdots\times\Pi_{N}$. Under a given joint policy $\pi$, the long-term average reward for agent $i$ starting from initial state $s$ is
\begin{equation}
       \rho_i^{\pi}(s):=\underset{T \to \infty}{\lim\inf}\frac{1}{T}\mathbb{E}_{\pi} \left[\sum_{t=0}^{T-1}r_i(s^{(t)},a^{(t)})|s^{(0)}=s \right].
\end{equation}

In this work, we will restrict our attention to an ergodic underlying MDP:
\begin{assumption}\label{asp:1}
    For any joint policy $\pi \in \Pi$, the induced Markov chain is irreducible and aperiodic.
\end{assumption}

Under Assumption 1, there exists a unique stationary distribution $\nu^{\pi}\in\triangle(\mathcal{S})$  independent of the initial state $s$ \citep{puter} i.e., $\nu^{\pi}(s')= \underset{T \to \infty}{\lim}\frac{1}{T}\mathbb{E}_{\pi}\left[\sum_{t=0}^{T-1}\mathbb{I}(s^{(t)}=s')|s^{(0)}=s \right]={\lim}_{T\to\infty}\mathbb{E}_{\pi}\left[\mathbb{I}(s^{(T)}=s')|s^{(0)}=s \right]$ for any $s\in\mathcal{S}$. As a result, 
the average reward $\rho_i^\pi(s) = \langle\nu^{\pi}, r_i^{\pi}\rangle$ is also independent of the initial state $s$ and we write it as $\rho^{\pi}_i$ for simplicity.

\begin{definition}[Average reward Markov potential games]
The average reward Markov potential game (ARMPG) is a special case of AMG, where there exists a potential function $\Phi(\pi):\Pi_1\times\Pi_2\times...\Pi_N\to \mathbb{R}$, such that for any $i$, $\pi_i,\pi_i'\in\Pi_i$ and $\pi_{-i}\in\Pi_{-i}$, 
\begin{equation*}
    \Phi(\pi_i,\pi_{-i})-\Phi(\pi_i',\pi_{-i})=\rho_i^{\pi_i,\pi_{-i}}-\rho_i^{\pi_i',\pi_{-i}}.
\end{equation*}
\end{definition}

Denote by $C_{\Phi}:=max_{\pi,\pi'}|\Phi(\pi)-\Phi(\pi')|$ the span of the potential function $\Phi$. Note that $C_{\Phi} \leq N$ since for any joint policies $\pi$ and $\pi'$, $|\Phi(\pi)-\Phi(\pi')| =|\rho^{\pi}- \rho^{\pi_1',\pi_{2:N}} 
    + \rho^{\pi_1',\pi_{2: N}}-\rho^{\pi_{1,2}',\pi_{3:N}}+ \hdots + 
    \rho^{\pi_{1:N-1}',\pi_{N}}-\rho^{\pi'}|\leq N $.


\begin{definition}[Nash and $\epsilon$-Nash equilibrium]
    A policy $\pi^*$ is a Nash equilibrium if for each agent $i$,
\begin{equation*}
    \rho^{\pi^*_i, \pi^*_{-i}}\geq \rho^{\pi_i, \pi^*_{-i}},\quad \forall \pi_i \in\Pi_i,
\end{equation*}
or an $\epsilon$-Nash equilibrium if
\begin{equation*}
    \rho^{\pi^*_i, \pi^*_{-i}}\geq \rho^{\pi_i, \pi^*_{-i}}-\epsilon,\quad \forall \pi_i\in\Pi_i.
\end{equation*}
\end{definition}

It may be noted that the maximizer of the potential function is a Nash equilibrium, while the converse may not be true since there could be multiple Nash equilibria.



\subsection{Value Functions and Their Properties}

The differential value function
\begin{equation}
    V^{\pi}_i(s) := \mathbb{E}_{\pi} \left[\sum_{t=0}^{\infty}(r_i(s^{(t)},\textbf{a}^{(t)})-\rho^{\pi}_i)|s^{(0)}=s \right]
\end{equation}
captures the accumulated deviation from the stationary performance. The differential Q function and differential advantage function are defined respectively as: 
\begin{align*}
    &Q^{\pi}_i(s,\textbf{a}) := \mathbb{E}_{\pi}\left[\sum_{t=0}^{\infty}(r_i(s^{(t)},\textbf{a}^{(t)})-\rho^{\pi}_i)|s^{(0)}=s, \textbf{a}^{(0)}=\textbf{a}\right], \\
    &A^{\pi}_i(s,\textbf{a}) := Q^{\pi}_i(s,\textbf{a})-V^{\pi}(s,\textbf{a}).
\end{align*}

Taking the expectation with respect to other policies except $j$, we can describe how agent $j$ affects $Q_i^{\pi}$ by:
\begin{equation}
    \overline{Q}^{\pi}_{j;i}(s,a_j) := \underset{a_{-j}\in\mathcal{A}_{-j}}{\sum} \pi_{-j}(a_{-j}|s)Q^{\pi}_i(s,a_j,a_{-j}).
\end{equation}
To simplify notation, we will use $\overline{Q}^{\pi}_i$ to represent $\overline{Q}^{\pi}_{i;i}$, and define $r_i^{\pi}(s):=E_{\mathbf{a}\sim\pi(\cdot|s)}r_i(s,\mathbf{a})$, $r_i^{\pi_{-j}}(s,a_j):=E_{a_{-j}\sim\pi_{-j}(\cdot|s)}r_i(s,a_j,a_{-j})$, and $\overline{A}^{\pi}_{i}(s,a_i) := \underset{a_{-i}\in\mathcal{A}_{-i}}{\sum} \pi_{-i}(a_{-i}|s)A^{\pi}_i(s,a_i,a_{-i})$.

With $P_{\pi}\in\mathbb{R}^{S\times S}$ denoting the state transition probability matrix induced by policy $\pi$, and $P^{\pi}\in\mathbb{R}^{SA\times SA}$ denoting the induced state-action transition probability matrix, the differential value function $\mathbf{V_i^{\pi}}$ and differential Q function $\mathbf{Q_i^{\pi}}$ are solutions of the following equations, up to a constant shift \citep{puter}:
\begin{equation}
\begin{aligned}
    \mathbf{V^{\pi}} &= \mathbf{r}_i^{\pi}-\rho_i^{\pi}\mathbf{1}_{S}+P_{\pi}\mathbf{V^{\pi}}, \\
    \mathbf{Q^{\pi}} &= \mathbf{r}_i - \rho_i^{\pi}\mathbf{1}_{SA} + P^{\pi}\mathbf{Q^{\pi}}.
\end{aligned}
\end{equation}

\begin{lemma}[Performance difference lemma] \label{lm:2}
For any policies $\pi_j$, $\pi_j'$ $\in\Pi_j$, and $\pi_{-j}\in\Pi_{-j}$, the difference between the average rewards for each agent $i$ is
    \begin{align*}
        & \rho_i^{\pi_j,\pi_{-j}}-\rho_i^{\pi'_j,\pi_{-j}} \\
        =& \mathbb{E}_{s\sim\nu^{\pi_j, \pi_{-j}}}\langle\overline{Q}_{j:i}^{\pi'_j,\pi_{-j}}(s,\cdot), \pi_j(\cdot|s)-\pi'_j(\cdot|s)\rangle \\ 
        =& \mathbb{E}_{s\sim\nu^{\pi_j, \pi_{-j}}}\sum_{a_j}\pi_j(a_j|s)\overline{A}_{j:i}^{\pi'_j,\pi_{-j}}(s,a_j).
    \end{align*}
\end{lemma}

Let $\overrightarrow{\mathbf{e}}(s,a_j)\in\{0,1\}^{S\times A_j}$ denote the unit vector where the only non-zero term has the index $(s,a_j)$. With the definition of Fréchet derivative \citep{dieudonne2011foundations}, we can verify that there exists a linear operator $\mathbf{A}:=\sum_s\nu^{\pi}(s)\sum_{a_j}\overrightarrow{\mathbf{e}}(s,a_j)\overline{Q}_{j;i}^{\pi}(s,a_j)$ mapping the set $U=\{u\in\mathbb{R}^{S\times A_j}:\sum_{a\in \mathcal{A}_j} u(s,a_j)=0,\ \forall s\in\mathcal{S}\}$ to $\mathbb{R}^{S\times A_j}$ s.t.~$\lim_{\lVert u\rVert_2\to0}\frac{\lVert\rho^{\pi_j+u,\pi_{-j}}-\rho_i^{\pi_j,\pi_{-j}}-\langle\mathbf{A},u\rangle\rVert_2}{\lVert u\rVert_2}=0$. This can be shown by the performance difference lemma.

\begin{lemma}[Partial derivative]\label{lm:3}
    For any $i,j$, and any policies $\pi_j\in\Pi_j$, $\pi_{-j}\in\Pi_{-j}$,
    \begin{align*}
        \frac{\partial \rho_i^{\pi}}{\partial \pi_j(a_j|s)} = \overline{Q}_{j;i}^{\pi}(s,a_j)\nu^{\pi}(s),\\
        \frac{\partial \Phi(\pi)}{\partial \pi_j(a_j|s)} = \overline{Q}_{j}^{\pi}(s,a_j)\nu^{\pi}(s).
    \end{align*}
\end{lemma}

Let $\{\lambda_i(\mathbf{M})\}_{i=1,...,n}$ be the eigenvalues of matrix $\mathbf{M}\in\mathcal{R}^{n\times n}$, where $|\lambda_1(\mathbf{M})|\geq|\lambda_2(\mathbf{M})|\geq...\geq |\lambda_n(\mathbf{M})|$. The largest eigenvalue of $P^{\pi}$ is 1 corresponding to the eigenvector $\nu^{\pi}P^{\pi}=\nu^{\pi}$ \cite{puter}. The second largest eigenvalue is strictly less than $1$ and is related to the mixing time of the Markov chain induced by $\pi$ \citep[Lemma 2.1]{kale2013eigenvalues}.

\begin{definition}[\cite{Li}]\label{def:3}
    Let $1-\Gamma$ be the probability of the least visited state of the MDP, with $\Gamma:=1 - \min_{\pi\in\Pi,s\in\mathbb{S}}\nu^{\pi}(s)$.
   Note that $\Gamma\in(0,1)$.
\end{definition}

\begin{definition}
\label{def:4}
    The mixing coefficient of the MDP is defined as:
    \begin{equation*}
        \kappa_0:=\underset{\pi\in\Pi}{\max}\frac{1}{1-|\lambda_2(P^{\pi})|}.
    \end{equation*}
\end{definition}
Unlike our definitions of $\Gamma$ and $\kappa_0$, some literature in the field of average reward MDP employs the concepts of hitting time $t_{hit}:=\max_{\pi}\max_s\frac{1}{\nu^{\pi}(s)}$ and mixing time $t_{mix}:=\max_{\pi}\min\{t\geq1|\lVert (P^{\pi})^t(\cdot|s)-\nu^{\pi} \rVert_1\leq \frac{1}{4}, \forall s\in\mathcal{S}\}$ of the underlying MDP, \citep{wei2020model,bai2023regret}. We can establish a close relationship between these two sets of definitions. Specifically $\Gamma$ is related to the hitting time as $t_{hit}=\frac{1}{1-\Gamma}$, and from \cref{lm:4} $\kappa_0$ is related to the mixing time as $t_{mix}=O(\frac{1}{\log(1-\frac{1}{\kappa_0})})$. It is also evident that $\kappa_0$ is finite under Assumption \ref{asp:1} in the tabular setting. 

\begin{lemma}\label{lm:4}
     Let $C_p:=\min\{\sqrt{\frac{S}{1-\Gamma}},\frac{1}{1-\Gamma}\}$ and $\varrho:=1-\frac{1}{\kappa_0}$. Then for any policy $\pi$,
    \begin{equation*}
        \sup_s\lVert (P^{\pi})^t(\cdot|s_0=s)-\nu^{\pi}\rVert_1\leq C_p\varrho^t, \forall\ t>0.
    \end{equation*}
\end{lemma}

\begin{definition}\label{def:5}
Define $\kappa:=\max_i\max_{\pi}\min_{b\in\mathbb{R}}\lVert\overline{Q}_i^{\pi}+b\mathbf{1}\rVert_{\infty}$ for any reward function $r\in[0,1]$. 
\end{definition}

By \cref{lm:4}, the span of the differential Q function can be bounded as $\kappa\leq\lVert \overline Q_i^{\pi}\rVert_{\infty}\leq 1+C_p\sum_{t=1}^{\infty}\varrho^t=1+\frac{C_p\varrho}{1-\varrho}\leq C_p\kappa_0$. Details are in Appendix \cref{prop:3}.

\subsection{Performance Metric}
We study “independent” policy optimization. By this we mean that at step $t$, each agent $i$ updates its policy $\pi_i^t$ to $\pi_i^{t+1}$ based on the information it can collect locally without coordinating with other agents. The goal is to find a Nash equilibrium, and we quantitatively measure the closeness of the joint policy $\pi^t$ to a Nash equilibrium by the $\text{Nash-gap}(t):=\max_{i}\max_{p\in\Pi_i}(\rho_i^{p,\pi_{-i}^t}-\rho_i^{\pi_i^t,\pi_{-i}^t})$.
The optimization algorithm is evaluated by the following notions of Nash regret, 
\begin{equation*}
\begin{aligned}
    & \text{Nash-Regret}(T):=\frac{1}{T}\sum_{\tau=0}^{T-1}\text{Nash-gap}(t), \\ 
    & \text{Nash-Regret}^*(T):=\frac{1}{T}\sum_{\tau=0}^{T-1}\text{Nash-gap}(t)^2.\\ 
\end{aligned}
\end{equation*}
It is clear that the Nash gap is positive and policy $\pi^t$ is an $\epsilon$-Nash equilibrium if $\text{Nash-gap}(t)\leq\epsilon$. Moreover, if the Nash regret (or Nash regret$^*$) is less than $\epsilon$, the policy $\pi^{t^*}$ with $t^*\in \arg\min_{t} \text{Nash-gap}(t)$, is an $\epsilon$ (or $\sqrt{\epsilon}$)-Nash equilibrium. 



\section{POLICY GRADIENT ALGORITHM}

We now analyze the independent projected policy gradient algorithm for average reward Markov potential games. The algorithm adopts a direct policy parameterization. At each step, each agent $i$ updates its policy independently along the gradient direction and projects it back to the policy space via $\text{Proj}_{\Pi_i}(\pi) :=\arg\min_{p\in \Pi_i}\lVert p-\pi\rVert_2$. 

We first consider the oracle-based setting, where the algorithm has access to a gradient-oracle that can exactly calculate the policy gradient of a given policy. We study the convergence performance in this setting and its time complexity. Subsequently, we consider the setting where there is no access to any oracle. We propose a gradient estimator based on trajectory samples, and study its sample complexity.

The key step in the analysis relies on the smoothness of the average value function. Unlike the discounted setting  where the gradient has a closed form expression since $I-\gamma P^{\pi}$ is invertible and its power series can be bounded \citep{agarwal2021theory, leonardos2021global}, in the average case, a similar analysis can not be applied since $\gamma=1$. In the following \cref{lm:smooth}, with the help of perturbation theory of stochastic matrices we show that the average reward function is smooth in both single-agent and multi-agent situations. The proof is in \cref{B.2}.

\begin{lemma}[Smoothness of $\rho$ and $\Phi$]\label{lm:smooth}
Denote $A_{\max}:=\max_iA_i$, $L:=\kappa_0^2S^{3/2}A_{\max}+\kappa_0\sqrt{S}A_{\max}$, and $L_{\Phi}:=N(\kappa_0^2S^{3/2}A_{\max}+\kappa_0(SA_{\max}+2A_{\max})+A_{\max})$. 

(a) For any i, and $\pi_{-i}\in\Pi_{-i}$, the average value $\rho_i^{\pi}$ is $\kappa_0^2S^{3/2}A_i+\kappa_0\sqrt{S}A_i$-smooth with respect to policy $\pi_i$. Moreover, for any $i$, $\rho_i^{\pi}$ is $L$-smooth with respect to policy $\pi_i$, i.e. $\lVert\nabla_{\pi_i}\rho_i^{\pi_i,\pi_{-i}}-\nabla_{\pi_i}\rho_i^{\pi_i',\pi_{-i}}\rVert_2\leq L\lVert\pi_i-\pi_i'\rVert_2$ for $\forall\ i,\ and\ \pi_i,\pi_i'\in\Pi_i$.

(b) The potential function $\Phi(\pi)$ is $L_{\Phi}$-smooth with respect to joint policy $\pi$, i.e. $\lVert\nabla\Phi(\pi)-\nabla\Phi(\pi')\rVert_2\leq L_{\Phi}\lVert\pi-\pi'\rVert_2$ for $\forall\ \pi,\pi'\in\Pi$.
\end{lemma}

We note that compared to the smoothness factor in discounted reward settings ($\frac{2\gamma A_{\max}}{(1-\gamma)^3}$ for single-agent, $\frac{2N\gamma A_{\max}}{(1-\gamma)^3}$ for multi-agent), the smoothness factor for the average reward setting has an extra dependence on state size $S$. The reason is that the second order linear derivative of $\rho^{\pi}$ depends on the $\ell_1$ norm of the linear derivative of $\nu^{\pi}$, while \cref{def:4} can only guarantee an $\ell_2$ bound. The exchange between $\ell_1$ and $\ell_2$ norms introduces the factor $S$.



\subsection{Policy Gradient Algorithm with Gradient Oracle}

\begin{algorithm}[t!]
	\caption{Independent projected policy gradient ascent }	
	\label{alg:1}
	\begin{algorithmic}[1]
	\STATE \textbf{Input:} learning rate $\beta>0$
        \STATE \textbf{Initialization:} $\pi^{(0)}_i(a_i|s)=1/A_i$ for any $i$, $s$, $a_i$
		\FOR {$t=0$ to $T-1$ }
		\STATE $\pi_i^{(t+1)}=\text{Proj}_{\Pi_i}(\pi_i^t+\beta\nabla_{\pi_i}\rho_i^{\pi^t}),\ \forall i$
		\ENDFOR
	\end{algorithmic}
\end{algorithm} 


We first introduce the distribution mismatch coefficient, which also appears in the convergence behavior of the policy gradient algorithm in discounted single-agent MDPs \citep{agarwal2021theory} and discounted Markov potential games \citep{ding2022independent,leonardos2021global}.

\begin{definition}[Distribution mismatch coefficient]\label{def:6}
$D:=\max_{\pi,\pi'\in\Pi}\lVert\frac{\nu^{\pi}}{\nu^{\pi'}}\rVert_{\infty}$.
\end{definition}
In the average reward setting, the coefficient $D$ can be upper bounded by $\frac{1}{1-\Gamma}$. The independent projected policy gradient ascent algorithm is given in \cref{alg:1}, the regret of which is bounded as follows, with its proof given in \cref{B.2}.


\begin{theorem}\label{thm:1}
    Choose learning rate $\beta:=\frac{1}{L_{\Phi}}$. Then the Nash-regret* of \cref{alg:1} is bounded by
    \begin{equation*}
        \text{Nash-regret}^*(T)= O\left(\frac{D^2L_{\Phi}C_{\Phi}S}{T}\right).
    \end{equation*}
\end{theorem}

Therefore, by setting time $T:=O\left(\frac{NC_{\Phi}D^2S^{5/2}A_{\max}\kappa_0^2}{\epsilon^2}\right)$, it yields an $\epsilon$-Nash equilibrium. 

 The analysis has two parts. Based on the smoothness estimated in \cref{lm:smooth}, the algorithm will converge to a stationary point by the optimization theory of gradient ascent. We then show that the stationary policy is a Nash equilibrium in the average reward Markov potential game, which establishes the convergence of the algorithm.

\subsection{Sample-Based Policy Gradient Estimate}
In practice, we usually do not have access to a gradient oracle. To apply the policy gradient algorithm, we need to estimate the gradient from trajectory samples. 
We propose a gradient estimator (\cref{alg:2.1}) which only relies on a single trajectory and is thus more practical in real applications since it does not require resetting the Markov process or a generative model. The sample-based policy gradient ascent algorithm is described in \cref{alg:2}.



The estimator $\hat{g}_i$ in \cref{alg:2.1} is based on \cref{lm:3}, which relates the policy gradient with $\overline{Q}^\pi_{i}(s, a_i)$, and we approximate it by $R_i(t)=\sum_{k=0}^{N}(r_i(s^{t+k},\mathbf{a}^{t+k})-\hat{\rho_i})$. It is generally hard to estimate the policy gradient in the average reward case. Unlike the discounted reward criterion, the policy gradient might be unbounded under average reward setting. We resolve this issue by adapting the sample length $N$ based on the mixing rate in \cref{lm:4}.
However, $\hat{g}_i$ may have large variance when $\pi_i(a_i'|s')$ is small. To overcome this, we restrict the policy class to $\Pi_{\alpha}:=\Pi_{1,\alpha}\times\hdots\times\Pi_{N,\alpha}$, where $\Pi_{i,\alpha}:=\{(1-\alpha)\pi_i +\alpha u_i|\forall \pi_i\in\Pi_i\}$ with $u_i:=(\text{Unif}_{\mathcal{A}_i})^S\in\Pi_i$ being the uniform policy. Such restriction has been considered in \cite{leonardos2021global, ding2022independent} previously. We can balance the representation power of the policy class and the variance of the gradient estimator by adjusting $\alpha$.

\begin{algorithm}[t!]
    \caption{Policy gradient ascent with estimation}	
    \label{alg:2}
    \begin{algorithmic}[1]
    \STATE \textbf{Input:} learning rate $\beta>0$, $K$, $N_1$, $N_2$, $N_3$ 
        \STATE \textbf{Initialization:} $\pi^{(0)}_i(a_i|s)=1/A_i$ for any $i$, $s$, $a_i$
        \FOR {$t=0$ to $T-1$ }
            \STATE agents take action independently and synchronously for $N_1+KN_2$ time steps to collect trajectories $\{\mathcal{T}_i^t\}$
            \FOR{agent $i$}
                \STATE $\hat{g}_i^t\gets \text{gradient estimation}(\mathcal{T}_i^t,\pi_i^t,K,N_1,N_2,N_3)$
                \STATE $\pi_i^{(t+1)}=\text{Proj}_{\Pi_{i,\alpha}}(\pi_i^t+\beta\hat{g}_i^t)$
            \ENDFOR
        \ENDFOR
    \end{algorithmic}
\end{algorithm} 

\begin{algorithm}[t!]
    \caption{Gradient estimation}
    \label{alg:2.1}
    \begin{algorithmic}[1]
    \STATE \textbf{Input:} trajectory $\mathcal{T}=(s^0,a_i^0,r_i^0,...,s^t,a_i^t,r_i^t)$, policy $\pi_i$, $K$, $N_1$, $N_2$. ($t=N_1+KN_2-1$) 
    \STATE $\hat{\rho}_i\gets\frac{2}{N_1}\sum_{t=\frac{N_1}{2}}^{N_1-1}r_i^t$  
    \STATE $g\gets0$
    \FOR{$k=0$ to $K-1$}
        \STATE $t_k\gets N_1+kN_2$
        \STATE $R(k)\gets\sum_{\tau=t_k}^{t_k+N_2-1}(r_i^{\tau}-\hat{\rho}_i)$
        \STATE  $g\gets g+R(k)\nabla_{\pi_i} \log\pi_i(a_i^{t_k}|s^{t_k})$
    \ENDFOR
    \STATE $\hat{g}_i\gets\frac{g}{K}$
    \STATE \textbf{Output:}$\hat{g}_i$
    \end{algorithmic}
\end{algorithm}

\begin{lemma}\label{lm:6}
    For any agent $i$, consider the gradient estimate $\hat{g}_i$ defined in \cref{alg:2.1}. Given the $(s,a,r)$-trajectory of length $KN_2+N_1$ and the policy $\pi_i\in\Pi_{i,\alpha}$ that generated it, the estimated gradient has $\ell_2$ error bounded as
\begin{equation*}
\begin{aligned}
    \mathbb{E}\lVert \hat{g}_i-\frac{\partial \rho_i^{\pi}}{\partial \pi_i}\rVert_2^2\leq& \left(\frac{1}{\alpha} + 1\right)\frac{2A_{\max}N_2^2}{K}\\
    &+\frac{4C_pA_{\max}}{1-\varrho^{N_2}}\left(\sqrt{\frac{2}{\alpha}}+\sqrt{2}\right)\frac{N_2^2}{K}\varrho^{N_2}\\
    &+\frac{16A_{\max}C_p^2}{(1-\varrho)^2}\frac{N_2^2}{N_1^2}\varrho^{N_1}+\frac{2A_{\max}C_p^2}{(1-\varrho)^2}\varrho^{2N_2}.
\end{aligned}
\end{equation*}
\end{lemma}
We can guarantee an $\ell_2$ error of $O(\delta)$ by choosing $N_1=N_2=O(\log\frac{1}{\delta}) $, and $K=\tilde O(\frac{A_{\max}}{\alpha\delta})$. A detailed proof is in \cref{B.3}.


\begin{theorem}\label{thm:2}
    If all players independently and synchronously run \cref{alg:2} with learning rate $\beta=\frac{1}{L_{\Phi}}$, the Nash regret is bounded as:
    \begin{align*}
        \mathbb{E}[\textnormal{Nash-regret}^*(T)]=&O{\Big (}\frac{D^2SL_{\Phi}C_{\Phi}}{T}+\kappa^2\alpha^2 \\
        &  +\kappa^2DA_{\max}L_{\Phi}^2\delta {\Big )}.
    \end{align*}
\end{theorem}
We can therefore determinate an $\epsilon$-Nash equilibrium by 
choosing $T=O(\frac{NC_{\Phi}D^2S^{5/2}A_{max}\kappa_0^2}{\epsilon^2})$, $\alpha=O(\frac{\epsilon}{\kappa})$, $\delta=O(\frac{\epsilon^2}{\kappa^2DA_{\max}L_{\Phi}^2})$, $K=\tilde O(\frac{A_{\max}}{\alpha\delta})$, and $N_1=N_2=\tilde O(1)$. Substituting the bound for $\kappa$ and $L_{\Phi}$, the sample complexity is then $T(KN_2+N_1)=\tilde O(\frac{N^3D^3C_{\Phi}S^7A_{\max}^5\kappa_0^9}{\epsilon^5(1-\Gamma)^{3/2}})$. 

To analyze \cref{alg:2}, we introduce the shadow policy $\tilde\pi^{t+1}:=\text{Proj}_{\Pi_{\alpha}}(\pi^t+\beta\nabla_{\pi}\Phi(\pi^t))$ projected from the true deterministic policy gradient. The distance between $\tilde\pi^t$ and $\pi^t$ is bounded by the estimation error. Since the shadow policy $\tilde\pi^t$ can capture the optimality criterion in the update rule in \cref{alg:2}, a bound on the Nash gap with policy improvement w.r.t.~potential value can be provided for each time step. Together with the bounded distance from true policy $\pi^t$, the conclusion follows. 

\textit{Remark.} We note that there are some mistakes in the analysis of the sample-based policy gradient in \cite[Theorem 4.7]{leonardos2021global} about the minimizer of the Moreau envelope and the update analysis, which are elaborated in \cref{B.4}. We address the difficulties in analyzing the sample-based policy gradient algorithm, and as far as we are aware, \cref{thm:2} is the first sample-based projected policy gradient algorithm for Markov potential games with a rigorous performance guarantee.


\section{PROXIMAL-Q ALGORITHM}
\begin{algorithm}[t!]
    \caption{Independent proximal-Q}\label{alg:3}
    \begin{algorithmic}[1]
    \STATE \textbf{Input:} learning rate $\beta>0$
    \STATE \textbf{Initialization:} $\pi^{(0)}_i(a_i|s)=1/A_i$ for any $i$, $s$, $a_i$
    \FOR{$t=0$ to $T-1$}
        \STATE $\pi_i^{t+1}(\cdot|s)=\underset{p(\cdot|s)\in\triangle(\mathcal{A}_i)}{\arg\max} \{\beta\langle\overline{Q}^{\pi^t}_i(s,\cdot), p(\cdot|s)\rangle_{\mathcal{A}_i} - \frac{1}{2}\lVert p(\cdot|s)-\pi_i^t(\cdot|s)\rVert_2^2 \}\ \forall s,i$
    \ENDFOR
    \end{algorithmic}
\end{algorithm}

In this section, we analyze another algorithm, the proximal-Q algorithm, under the assumption of the availability of an oracle for the differential value function. The more general case where there is no oracle, and its sample complexity analysis are addressed in \cref{C.1}. Instead of the policy gradient $\frac{\partial \rho_i^{\pi}}{\partial \pi_i(a_i|s)} = \overline{Q}_{i}^{\pi}(s,a_i)\nu^{\pi}(s)$, \cref{alg:3} uses the differential Q function as the ascent direction, which is less sensitive for states $s$ with a small visiting rate $\nu^{\pi}(s)$.
The key part of the regret analysis is to connect the one-step policy update rule with the difference between the respective potential values. The analysis in the discounted setting is based on the performance difference lemma and second order performance difference lemma \cite[Lemma 21]{ding2022independent}, both relying on the backward induction enabled by the discount factor $\gamma<1$, which are unfortunately not applicable in the average reward setting. To bound the second order difference for the average reward, we can use its smoothness property established in \cref{lm:smooth}, or carefully analyze the sensitivity of the two parts $Q^{\pi}$ and $\nu^{\pi}$ in the performance difference \cref{lm:2}. We emphasize that the sensitivity bounds for differential value functions also play an important role in establishing the regret bound independent of the size of the action set, as shown in Section 5. 

Define $\lVert\pi\rVert_{1,\infty}:=\max_s \lVert\pi(\cdot|s)\rVert_1$ and $\lVert \mathbf{M}\rVert_{\infty}:=\max_{\lVert \mathbf{x}\rVert_{\infty}=1}\lVert \mathbf{Mx}\rVert_{\infty}$ for matrix $\mathbf{M}$. One may note that $\lVert \mathbf{M}_1+\mathbf{M}_2\rVert_{\infty}\leq\lVert \mathbf{M}_1\rVert_{\infty}+\lVert \mathbf{M}_2\rVert_{\infty}$, $\lVert \mathbf{M}_1\mathbf{M}_2\rVert_{\infty}\leq\lVert \mathbf{M}_1\rVert_{\infty}\lVert \mathbf{M}_2\rVert_{\infty}$, and $\lVert \mathbf{M}\rVert_{\infty}\leq \max_i\sum_j|\mathbf{M}_{ij}|$.

Let $P^{\pi,\infty}:=((\nu^{\pi})^T,...,(\nu^{\pi})^T)^T$, with $\nu^{\pi}\in(0,1)^{1\times S}$, denote the infinite-step state transition matrix. There exists a closed form expression $V^{\pi}=(I-P^{\pi}+P^{\pi,\infty})^{-1}(I-P^{\pi,\infty})r^{\pi}$ \citep{puter}.

\begin{definition}\label{asp:7}
    For any policy $\pi\in\Pi$, $ (I-P^{\pi}+P^{\pi,\infty})$ is invertible (\cite{puter}). Define $ \kappa_1 := \max_{\pi} \lVert(I-P^{\pi}+P^{\pi,\infty})^{-1}\rVert_{\infty}$.
\end{definition}
Note that $(I-P^{\pi}+P^{\pi,\infty})^{-1}=I+\sum_{t=1}^{\infty}(P^{\pi}-P^{\pi,\infty})^t $. By \cref{lm:4} and $P^{\pi}P^{\pi,\infty}=P^{\pi,\infty}P^{\pi}=P^{\pi,\infty} $, $P^{\pi,\infty}P^{\pi,\infty}=P^{\pi,\infty}$, we have $\kappa_1\leq \lVert I\rVert_{\infty}+\sum_{t=1}^{\infty}\lVert (P^{\pi})^t-P^{\pi,\infty}\rVert_{\infty}\leq 1+\sum_{t=1}^{\infty}C_p\varrho^t=1+C_p\frac{\varrho}{1-\varrho}$.

Consider a general average reward MDP with reward function $r$. $\rho_r^{\pi}$, $V_r^{\pi}$ and $Q_r^{\pi}$ can be defined similarly as in Section 2.
\begin{lemma}[Sensitivity bounds for average reward MDP] \label{lm:7}
For any reward function taking values in $[0,1]$, and any policies $\pi,\pi'\in\Pi$, the following bounds hold:
    \begin{align*}
        |\nu^{\pi}(s)-\nu^{\pi'}(s)|\leq&\kappa\lVert\pi-\pi'\rVert_{1,\infty},\ \forall s\in\mathcal{S}\\
        |\rho^{\pi}_r-\rho^{\pi'}_r|\leq&\kappa\lVert\pi-\pi'\rVert_{1,\infty},\\
        \lVert V_r^{\pi}-V_r^{\pi'}\rVert_{\infty} \leq& \kappa_1(2+S(\kappa+\kappa_1)+S\kappa\kappa_1)\lVert\pi-\pi'\rVert_{1,\infty},\\
        \lVert Q^{\pi}_r-Q_r^{\pi'}\rVert_{\infty} \leq& (\kappa+2\kappa_1+S\kappa_1(\kappa+\kappa_1)+S\kappa\kappa_1^2)\\
        &\times\lVert\pi-\pi'\rVert_{1,\infty}.
    \end{align*}
\end{lemma}

Proofs are provided in \cref{C}. \cref{lm:7} is a general result for both single-agent and multi-agent settings. Consider a joint policy
$\pi(\mathbf{a}|s):=\pi_1(a_1|s)\pi_2(a_2|s)...\pi_N(a_N|s)$. If $\pi=(\pi_{j},\pi_{-j})$, $\pi'=(\pi'_j,\pi_{-j})$, then $||\pi-\pi'||_{1,\infty}=||\pi_j-\pi'_j||_{1,\infty}$. The following proposition is directly derived from \cref{lm:7}.
\begin{proposition}\label{prop:2}
$\lVert Q_i^{\pi_j,\pi_{-j}}-Q_i^{\pi_j',\pi_{-j}}\rVert_{\infty}\leq \kappa_Q\lVert\pi_j-\pi_j'\rVert_{1,\infty} $, where $\kappa_Q:=\kappa+2\kappa_1+S\kappa_1(\kappa+\kappa_1)+S\kappa\kappa_1^2$.
\end{proposition}

\begin{theorem}\label{thm:3}
If the learning rate is chosen as $\beta\leq\max\{\frac{1-\Gamma}{(N-1)(\kappa_Q+S\kappa^2)A_{\max}},\frac{1-\Gamma}{2L_{\Phi}}\}$, then \cref{alg:3} has a bounded Nash regret:
    \begin{equation*}
        \text{Nash-regret}(T)\leq \sqrt{D}(\kappa\sqrt{A_{max}}+\frac{2}{\beta})\sqrt{2\beta C_{\Phi}}\frac{1}{\sqrt{T}}.
    \end{equation*}
\end{theorem}

If we set the learning rate to $\beta=\frac{1-\Gamma}{2L_{\Phi}}$, the time complexity for an $\epsilon$-Nash equilibrium is $T=O(\frac{NC_{\Phi}DS^{3/2}A_{max}\kappa_0^2}{(1-\Gamma)\epsilon^2})$. The proof is given in \cref{C}.

\section{NATURAL POLICY GRADIENT}
\begin{algorithm}[t!]
    \caption{Independent natural policy gradient ascent}
    \label{alg:5}
    \begin{algorithmic}[1]
    \STATE \textbf{Input:} learning rate $\beta>0$
    \STATE \textbf{Initialization:} $\pi^{(0)}_i(a_i|s)=1/A_i$ for any $i$, $s$, $a_i$
    \FOR{$t=0$ to $T-1$}
        \STATE $\pi_i^{t+1}(\cdot|s)=\underset{p(\cdot|s)\in\triangle(\mathcal{A}_i)}{\arg\max} \{\beta\langle\overline{Q}^{\pi^t}_i(s,\cdot), p(\cdot|s)\rangle_{\mathcal{A}_i} - D^p_{\pi_i^{t}}(s) \} \ \forall s,i$
    \ENDFOR
    \end{algorithmic}
\end{algorithm}

Finally, we analyze the natural policy gradient (NPG) algorithm under the average reward setting. NPG is a powerful technique to accelerate the convergence of policy update with Fisher information via preconditioning \citep{kakade2001natural}. We consider the independent NPG \cref{alg:5} under the availability of a differential value function oracle. 

Under the softmax parameterization, the joint policy $\pi^{\theta}=(\pi_1^{\theta_1},\hdots,\pi_N^{\theta_N})$ with $\theta=(\theta_1,\hdots,\theta_N)\in\mathbb{R}^{SA}$ is $\pi^{\theta_i}_i(s,a)=\frac{\exp(\theta_{s,a_i})}{\sum_{a_i'\in\mathcal{A}_i}\theta_{s,a_i'}}$. The gradient of $\rho$ (or $\Phi$) w.r.t.~$\theta$ is $\frac{\partial\rho_i^{\pi}}{\partial\theta}=\sum_{s,a}\nu^{\pi}(s)\pi(a|s)\frac{\partial\log\pi(s,a)}{\partial\theta}Q_i^{\pi}(s,a)$ \citep{sutton1999policy}. With the Fisher information matrix defined as $F_i(\theta):=\mathbb{E}_{s\sim\nu^{\pi^{\theta}},a_i\sim\pi^{\theta_i}(\cdot|s)}[(\frac{\partial \log\pi^{\theta_i}(a_i|s)}{\partial\theta_i})(\frac{\partial \log\pi^{\theta_i}(a_i|s)}{\partial\theta_i})^T]$ \citep{kakade2001natural}, the natural policy gradient update is 
$\theta_i^{t+1}=\theta_i^t+F_i(\theta_i)^{\dagger}\nabla_{\theta_i}\rho_i^{\pi^{\theta}}.$ 
It can be shown that the NPG update is equivalent to the update in \cref{alg:5} (the proof is provided in \cref{D}). In \cref{alg:5}, $D^p_q(s):=\sum_a p(a|s)\log\frac{p(a|s)}{q(a|s)}$ is the Kullback–Leibler (KL) Divergence between distributions $p(\cdot|s)$ and $q(\cdot|s)$. There is a closed-form expression for the update, $\pi_i^{t+1}(a_i|s)\varpropto \pi_i^t(a_i|s)\exp\left( \beta \overline{Q}_i^{\pi^t}(s,a_i)\right)\varpropto \pi_i^t(a_i|s)\exp\left( \beta \overline{A}_i^{\pi^t}(s,a_i)\right)$, which can be verified by the Karush–Kuhn–Tucker (KKT) condition. This does not require the calculation of the inverse of the Fisher information matrix and the gradient oracle, but employs a differential value function oracle instead. 



We first show the monotonic improvement property of the NPG one-step update.
\begin{lemma}\label{lm:8}
Let $Z_t^{i,s}:=\sum_{a_i}\pi_i^t(a_i|s)\exp\left(\beta \overline{A}_i^t(s,a_i)\right)$. When $\beta\leq\max\{\frac{1-\Gamma}{(N-1)(\kappa_Q+S\kappa^2)},\frac{1-\Gamma}{L_{\Phi}} \}$,
    \begin{equation*}
    \begin{aligned}
        \Phi(\pi^{t+1}) - \Phi(\pi^t) \geq \frac{1}{\beta}\sum_i\mathbb{E}_{s\sim\nu^{\pi_i^{t+1},\pi_i^t}}\log Z_i^{t,s} \geq 0.
    \end{aligned}
    \end{equation*}
\end{lemma}
In the discounted reward setting, the performance difference lemma provides a bound for the monotone improvement \cite[Lemma 20]{zhang2022global} which is applicable only when the potential function $\Phi(\pi)$ can be decomposed as the discounted summation of some state action reward function $\Phi(\pi)=\mathbb{E}_{a\sim\pi,s\sim \eta}\sum_t\gamma^t\phi(s,a)$. Here we do not need such an additional assumption, but utilize the smoothness (\cref{lm:smooth}) or the sensitivity bound for differential Q function (\cref{lm:7}) instead. 

Let $c(t):= \min_i\min_s\sum_{a_i^*\in argmax_{a_i\in\mathcal{A}_i}\overline{Q}_i^{\pi^t}(s,a_i)}\pi^{t}_i(a_i|s)$, which is also considered in \cite{zhang2022global}. It indicates the exploration power of the policy $\pi_t$, i.e., the probability mass covering the optimal actions. Define $c:=\inf_t c(t)$. The following lemma shows that $c$ is strictly positive, which is key to bounding the convergence rate. The proofs are given in \cref{D}.
\begin{lemma}\label{lm:9}
    If all stationary points of the potential function $\Phi(\theta)=\Phi(\pi(\theta))$ are isolated, $\beta\leq\min\{\max\{\frac{1-\Gamma}{(N-1)(\kappa_Q+S\kappa^2)},\frac{1-\Gamma}{L_{\Phi}} \}, \frac{1}{2\kappa}\}$, \cref{alg:5} asymptotically converges to a Nash equilibrium. Then $c>0$.
\end{lemma}


\begin{theorem}\label{thm:4}
    If all stationary points of the potential function $\Phi(\theta)=\Phi(\pi(\theta))$ are isolated, $\beta\leq\min\{\max\{\frac{1-\Gamma}{(N-1)(\kappa_Q+S\kappa^2)},\frac{1-\Gamma}{L_{\Phi}} \}, \frac{1}{2\kappa}\}$, the regret of \cref{alg:5} can be bounded as
    $$\text{Nash-regret}^*(T)\leq\frac{3C_{\Phi}}{c\beta(1-\Gamma) T}.$$
\end{theorem}
Substituting the bound for $\kappa$ and $\kappa_Q$, $\beta=\frac{1-\Gamma}{NS^{3/2}\kappa_0^2\min\{A_{\max},\frac{S^2\kappa_0}{(1-\Gamma)^{7/2}}\}}$, the time complexity is $T=O(\frac{NC_{\Phi}S^{3/2}\kappa_0^2\min\{\frac{S^2\kappa_0}{(1-\Gamma)^{7/2}}, A_{\max}\}}{c(1-\Gamma)^2\epsilon^2})$.
When the action set has a large cardinality $A_{\max}$, it gives an $A_{\max}$-independent bound for the chosen learning rate.

\section{EXPERIMENTS}

\begin{figure}[t!]
\vspace{.3in}
\subfigure[Large LVR, large RG]{\label{fig:a}\includegraphics[width=0.50\linewidth]{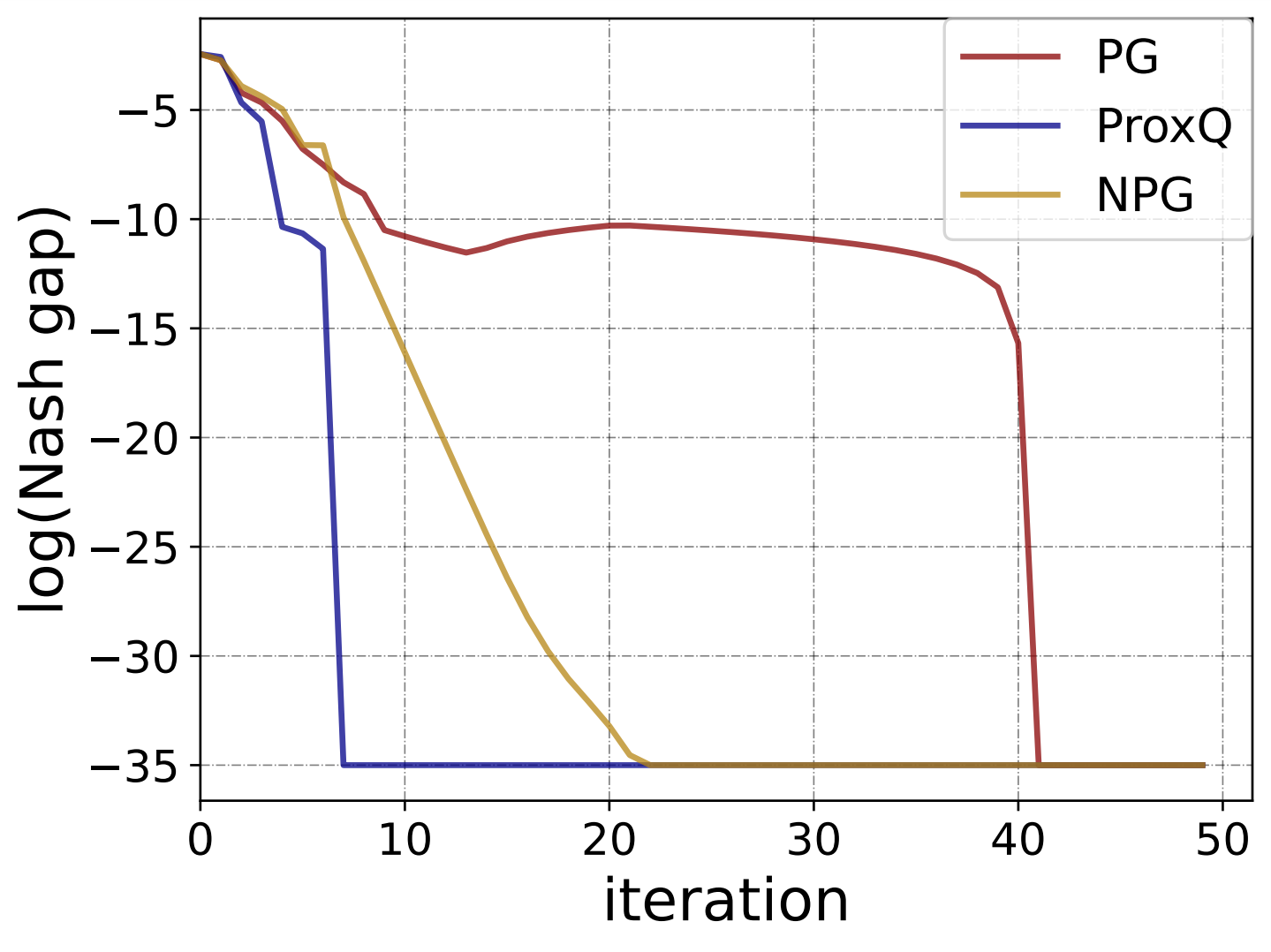}}
\subfigure[Small LVR, large RG]{\label{fig:b}\includegraphics[width=0.49\linewidth]{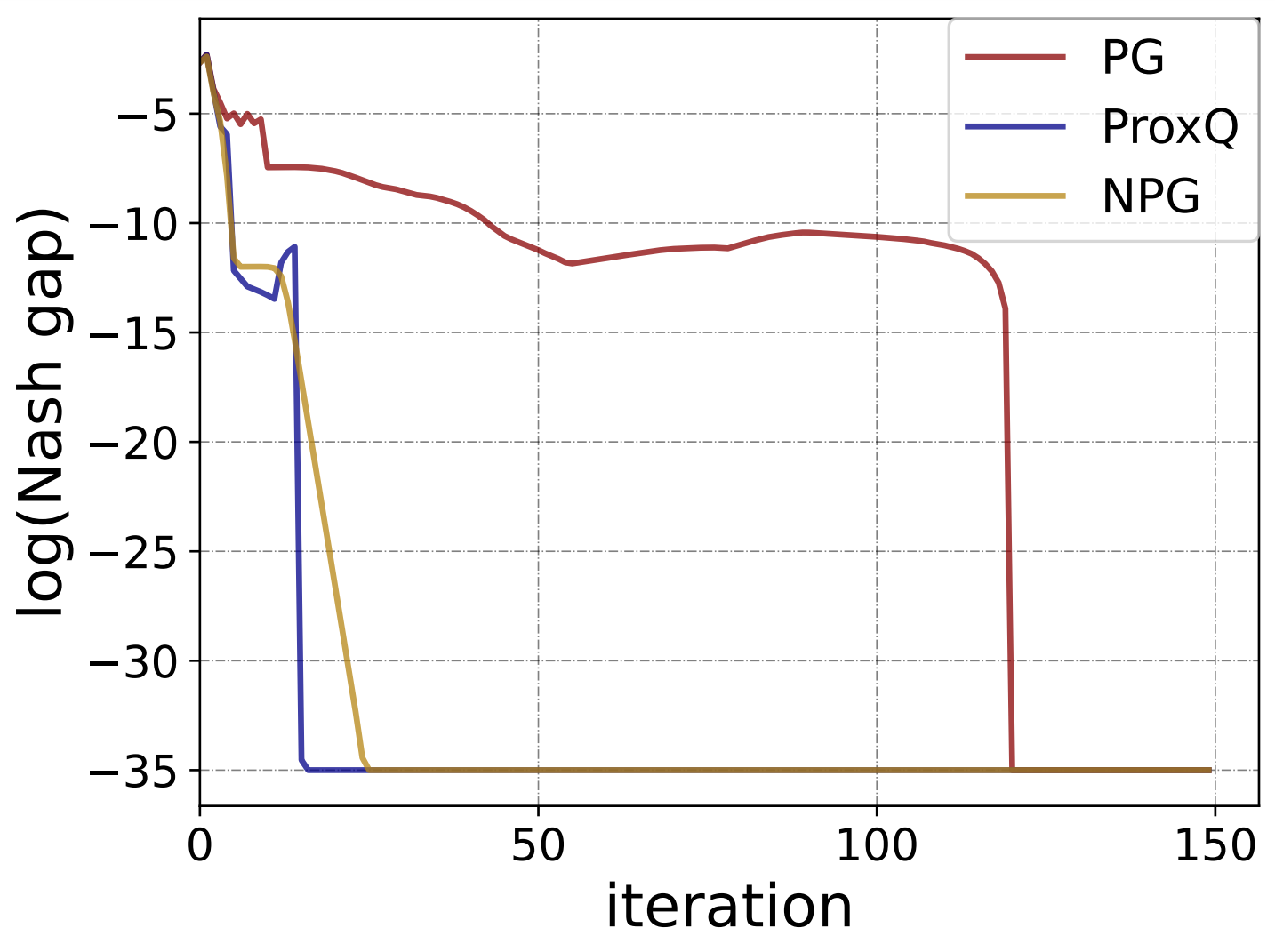}}
\subfigure[Large LVR, small RG]{\label{fig:c}\includegraphics[width=0.50\linewidth]{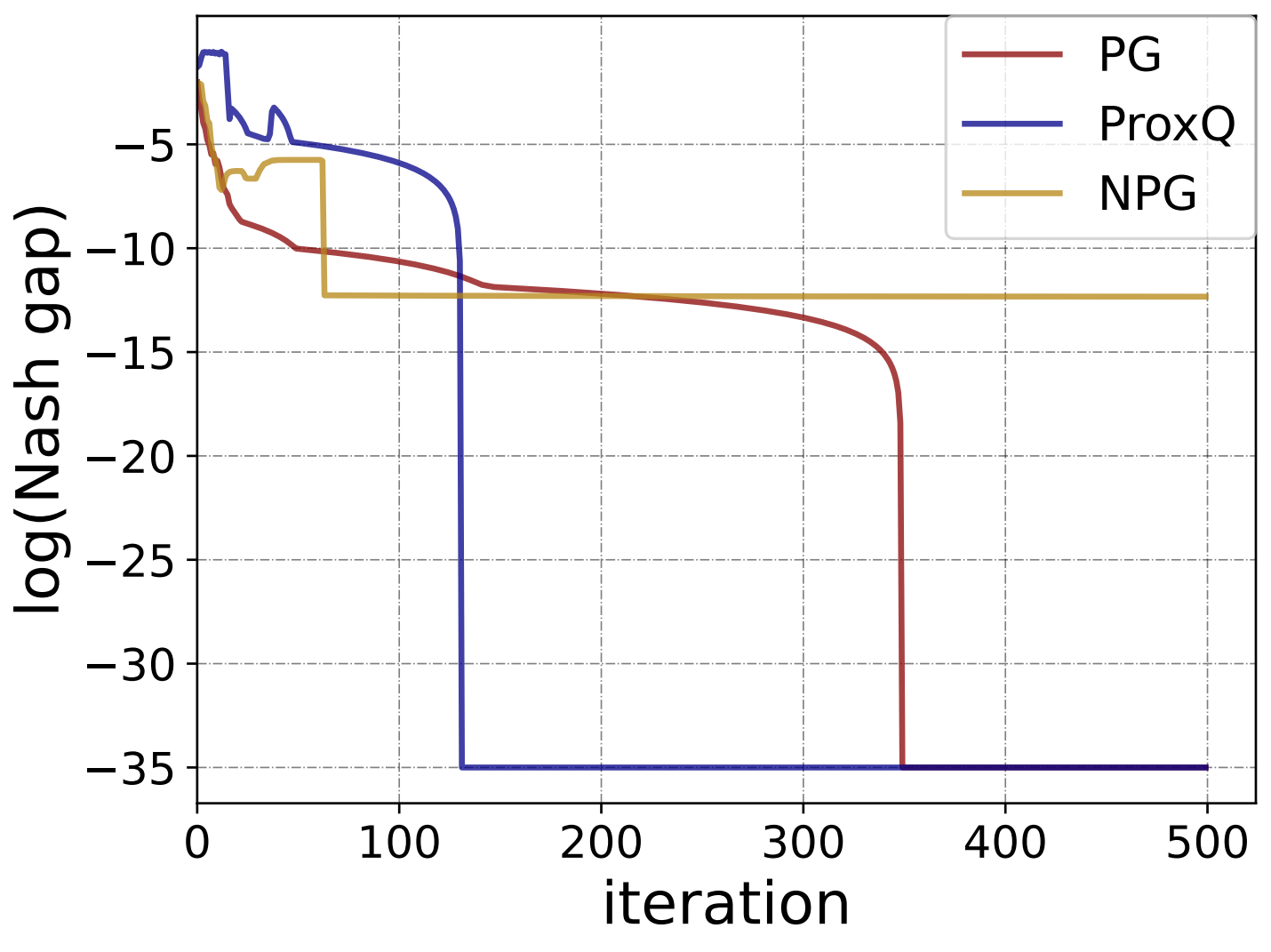}}
\subfigure[Large LVR, small RG]{\label{fig:d}\includegraphics[width=0.49\linewidth]{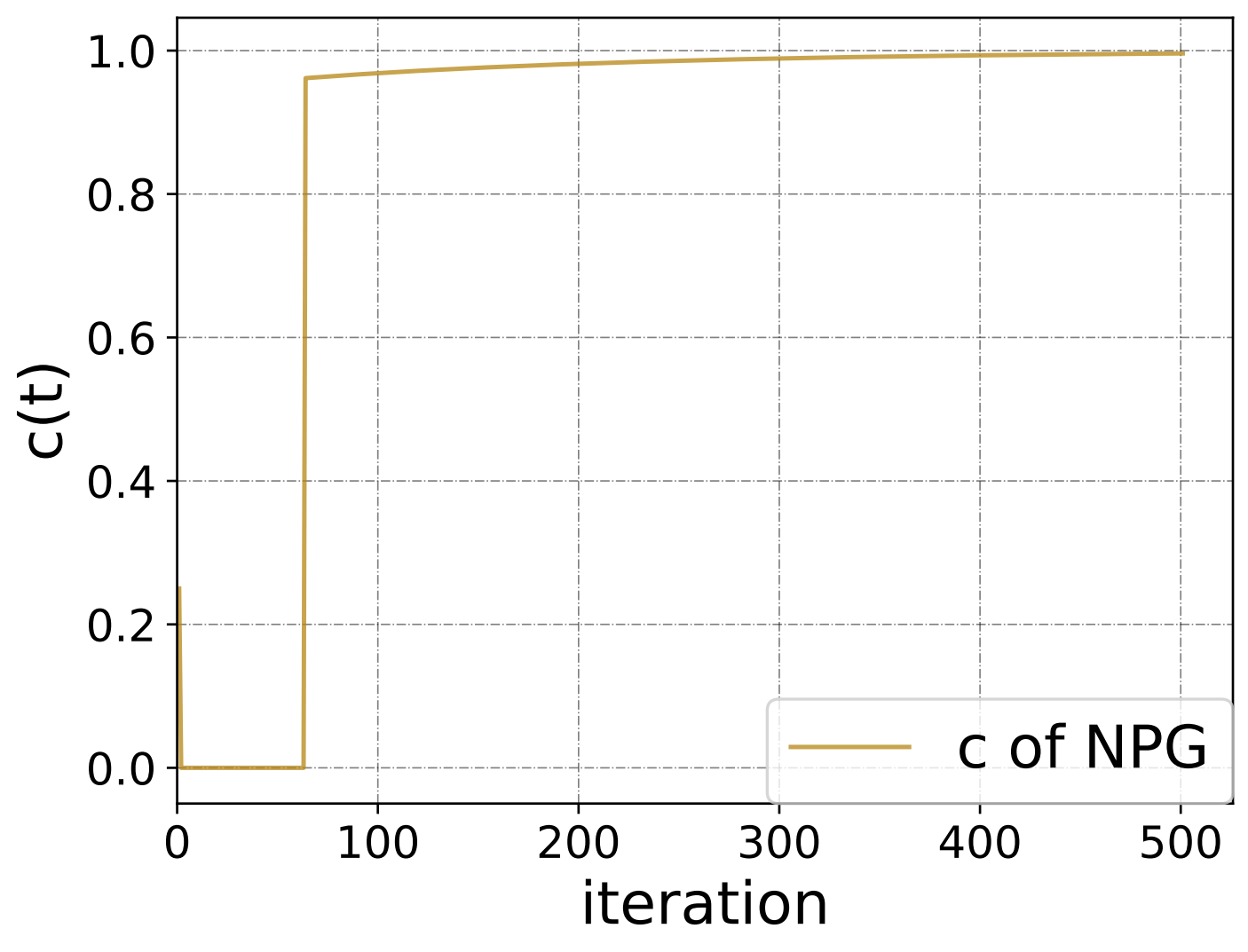}}
\subfigure[$\ell_1$ accuracy]{\label{fig:e}\includegraphics[width=0.49\linewidth]{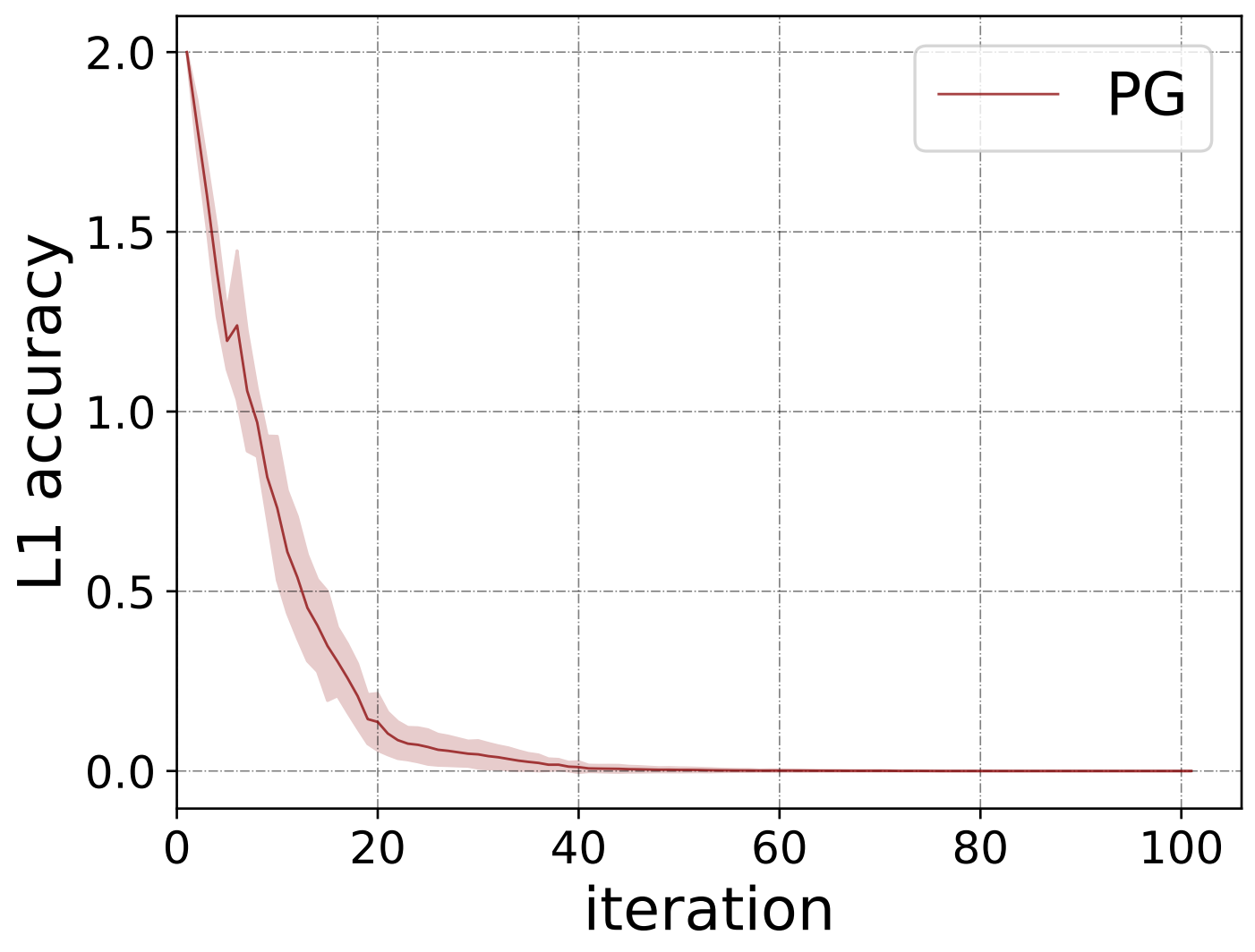}}
\subfigure[Nash gap]{\label{fig:f}\includegraphics[width=0.5\linewidth]{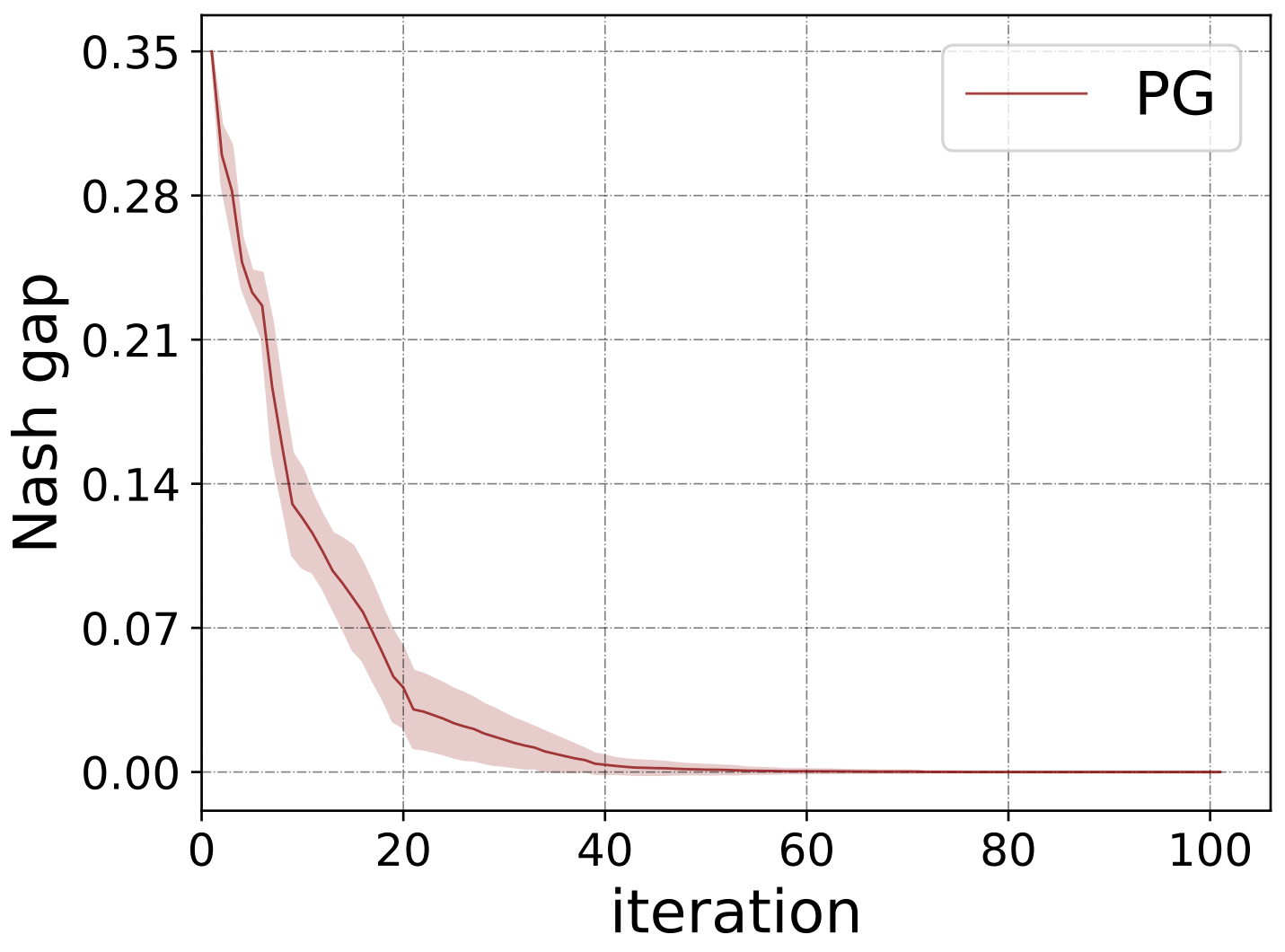}}
\caption{(a)(b)(c)(d) are results for the oracle setting. 
Since the Nash gap can be as low as $0.0$, we truncate the $\log(\text{Nash gap})$ at $-35$ from below. (d) depicts the change in $c(t)$ of the NPG algorithm for \cref{fig:c}. (e) and (f) present the results of \cref{alg:2}. The solid lines are the means of trajectories over seven random seeds and shaded regions are the standard deviations.}
\end{figure}

We first illustrate the convergence of the algorithms in the oracle setting. We randomly generate a Markov potential game model with $S=100$ states, and $A_1=4$, $A_2=3$, $A_3=2$ actions for the $N=3$ agents. We choose the largest learning rate. This yields the fastest convergence for each algorithm. The numerical experiments corroborate our theoretical findings in Theorems \ref{thm:1}, \ref{thm:3}, and \ref{thm:4}. 

Recall that in the theorems, a small least visited rate (LVR) $(1-\Gamma)$ has a negative impact on the convergence rate. As shown in Fig. \ref{fig:a} and \ref{fig:b},the small LVR impedes the convergence of all three algorithms. Comparing the theoretical findings of policy gradient and that of proximal-Q and NPG, we note that the complexity of the former is of order $S^{5/2}$ while the latter is of order $S^{3/2}$. This is also verified in Fig.~\ref{fig:b} and the effect is more significant when LVR is small. In addition, the convergence of NPG depends on the exploration factor $c(t)$.
We generate a reward function with a small reward gap (RG) between the optimal action and the second best optimal one, which increases the exploration difficulty and results in a small value of $c(t)$.
Despite $c(t)$ nearing 1, with very small RG, NPG updates can become minor, causing the algorithm to get stuck near a Nash equilibrium (\cref{fig:d}). Further discussion is in \cref{experiment}.

We further demonstrate that the proposed sample-based independent policy gradient \cref{alg:2} converges to the desired policy in Fig. \ref{fig:e} and \ref{fig:f}, under both the $\ell_1$ accuracy, i.e., $\frac{1}{N}\sum_{i=1}^N\lVert\pi_i^t-\pi_i^*\rVert_1$ and Nash gap. 


\begin{figure}[t!]
    \vspace{.3in}
    \centering
    \includegraphics[width=0.85\linewidth]{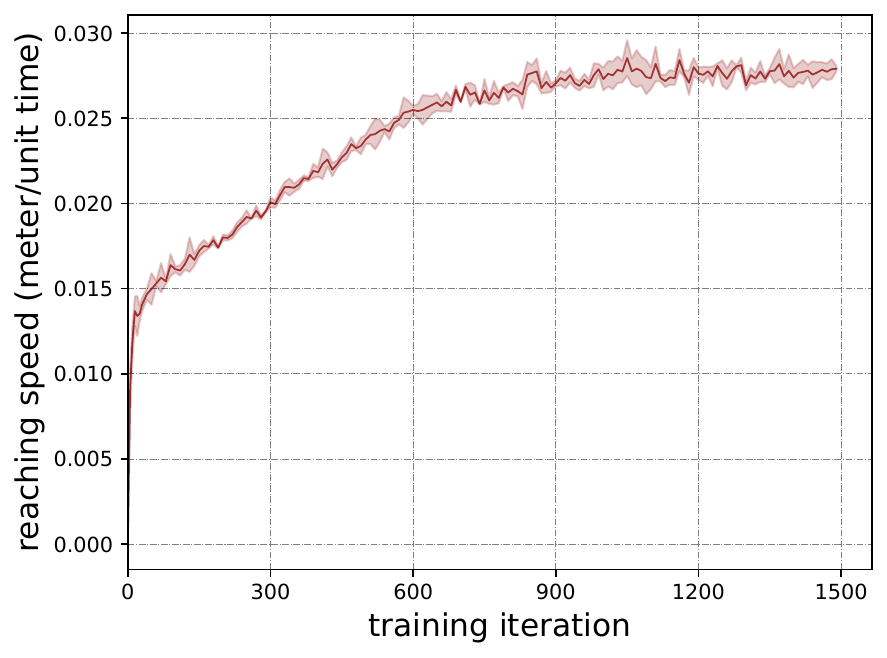}
    \caption{Training process of robot navigation task. The solid line is the mean of trajectories over three random seeds and shaded regions are the standard deviations.}
    \label{fig:2}
\end{figure}

To illustrate the potential of the independent learning scheme in averaged reward MARL, a more complex robot navigation task is conducted. There are two controllers (linear speed controller and angular speed controller) in a robot with each viewed as an agent. The agents gain rewards when the robot is moving toward the target. 
We implement a practical version of the independent average NPG algorithm with a neural network, which is inspired by Algorithm 1 of \cite{ma2021average}, and showcase its performance in \cref{fig:2}.

\section{CONCLUSION}
In this paper, we study Markov potential games under the average reward criterion and analyze three algorithms, policy gradient ascent, proximal-Q, and NPG under the access to an oracle. We establish time complexity that matches the results in discounted reward settings. We also propose a gradient estimator, which only relies on a single trajectory. The sample-based policy gradient ascent algorithm is shown to converge to a Nash equilibrium, and a sample complexity is provided. We also close several technical gaps in the analysis of policy gradient methods between the discounted and average reward settings.

\subsubsection*{Acknowledgements}
This material is based upon work partially supported by the US Army Contracting Command under W911NF-22-1-0151 and W911NF2120064, US National Science Foundation
under CMMI-2038625, and US Office of Naval Research under N00014-21-1-2385. 

\bibliography{references}

\newpage
 \clearpage
\onecolumn

\appendix
\section{EXPERIMENTAL DETAILS}\label{experiment}
\subsection{Oracle-based Algorithms}
We provide more details regarding the numerical experiments described in Section 6. In the oracle setting, 
when the state set size is greater than the size of the action set, according to \cref{thm:1}, \cref{thm:3} and \cref{thm:4}, the time complexities for projected policy gradient ascent, proximal-Q and NPG are $$O\left(\frac{NC_{\Phi}S^{5/2}A_{\max}\kappa_0^2}{(1-\Gamma)^2\epsilon^2}\right), O(\frac{NC_{\Phi}S^{3/2}A_{max}\kappa_0^2}{(1-\Gamma)^2\epsilon^2}), \mbox{ and } O(\frac{NC_{\Phi}S^{3/2}A_{\max}\kappa_0^2}{c(1-\Gamma)^2\epsilon^2}),$$ respectively. Notably, projected policy gradient ascent exhibits an additional $S$ dependency, while NPG has an additional dependency on $\frac{1}{c}$. Meanwhile, a small least visited rate (LVR) $1-\Gamma$ has a negative effect on all algorithms. To illustrate these effects, we conducted simulations on Markov potential games with large state spaces, varying LVR (large or small), and different exploration factors (large or small $c$), as described in the next paragraph.

We randomly generate a cooperative Markov potential game with the following parameters: $S=100$, $A_1=4$, $A_2=3$, and $A_3=2$. To control the LVR $(1-\Gamma)$, the elements of the transition probability matrix $P \in [0,1]^{SA_1A_2A_3\times S}$ are generated as follows. First, each entry is generated using a uniform distribution $\text{Unif}[0,1]$. Then we randomly select half of the states (denoting chosen states by $s'$) and generate the probabilities $P(s' | s, \mathbf{a})$ for every action profile $\mathbf{a}$ according to a uniform distribution $\text{Unif}[0,0.01]$ (for small LVR), or $\text{Unif}[0,0.1]$ or $\text{Unif}[0,1]$ (for large LVR). Subsequently, each row of the matrix is normalized. To reduce the exploration factor $c$, we generate a reward function with a small reward gap (RG) between the reward of the best action and that of the second best action, thereby increasing exploration difficulty and leading to a smaller $c$. The reward function is identical for all agents. It is randomly generated and denoted as $R \in [0,1]^{SA_1A_2A_3}$. For scenarios with a small RG, we use a uniform distribution $\text{Unif}[0,1]$ for all entries, or set reward $r(s,\mathbf{a}')$ to be $0.001$ smaller than $\max_{\mathbf{a}} r(s,\mathbf{a})$. This setup can indeed make $c(t)$ smaller as illustrated in \cref{fig:d}. In cases with a large RG, we select an action profile for each state at random and generate rewards according to $\text{Unif}[0.4,1]$, with the remaining entities generated according to $\text{Unif}[0,0.6]$. We observed that a larger learning rate leads to faster convergence, therefore, we chose the largest learning rate.

\subsection{Sample-based Algorithm}
In the sample-based setting, we run \cref{alg:2} for a manually designed Markov potential game with $S=2$, $A_1=A_2=2$, an action-independent transition probability matrix $P=\begin{pmatrix}
0.9 & 0.1\\
0.3 & 0.7
\end{pmatrix}$, and a reward function that is identical for each agent. The rewards for states 1 and 2 are $R_1=\begin{pmatrix}
1 & 0.2\\
0.8 & 0.2
\end{pmatrix}$ and $R_1=\begin{pmatrix}
0.2 & 1\\
0.1 & 0.6
\end{pmatrix}$, where columns indicate actions for agent 1, and rows indicate actions for agent 2. 
To achieve $\alpha=0.01$ and $\delta=0.01$, \cref{thm:2} suggests a learning rate $\beta\approx\frac{1}{N\kappa_0^2S^{3/2}A_{\max}}\approx0.01$ and trajectory length $KN_2+N_1\approx 10^8$. To reduce the number of samples needed, we choose a trajectory length $KN_2+N_1=51000$ with $N_1=1000, N_2=50, K=1000$, and reduce the step size every $20$ steps, from initially $0.5$ to eventually $0.0001$, to accommodate inaccurate gradient estimates.

\subsection{Robot Navigation Task}
We demonstrate the efficacy of average reward MARL by solving a complex robot navigation task for TurtleBot, a two-wheeled mobile robot
\citep{zhou2024natural, rengarajan2022enhanced, rengarajan2022reinforcement} in a simulation platform, Gazebo. The navigation task is to guide the robot to any designated target within a 2-meter radius. The state space is defined by a continuous 2-dimensional space representing the distance and relative orientation between the robot and the target. The robot is equipped with two controllers: an angular controller with control effect ranging from -1.5 rad/s to 1.5 rad/s, and a linear velocity controller with control effect ranging from 0 cm/s to 15 cm/s. 
To model the agents' control strategy, we implemented two independent agents to manage the angular and linear speeds, following an empirical adaptation of \cref{alg:5}. The reward function is formulated to be proportional to the product of the distance and the scaled orientation between the robot and the target. Hitting the boundary incurs a penalty of -200, while reaching the target yields a reward of 200. Each trajectory terminates upon collision, achievement of the goal, or after 500 simulation steps have elapsed.


All experiments were conducted on a CPU with an 11th Gen Intel(R) Core(TM) i7-11700 @ 2.50GHz.

\section{PROOFS OF SECTION 2}\label{A}
We first explore a sufficient condition for a Markov game to be a potential game. Previous works have studied the sufficient condition for the discounted reward setting \citep{leonardos2021global, zhang2021multi, 9992762}. These conditions encompass some important Markov games.

\begin{proposition}\label{prop:1}
    Consider a Markov game that is a static potential game $r_i(s,\mathbf{a})$ at every state $s\in\mathrm{S}$, i.e., there exists a common potential function $\phi(s,\mathbf{a})$ and an utility function $u_i(s,a_{-i})$ for each agent $i$, such that $r_i(s,\mathbf{a})=\phi(s,\textbf{a})+u_i(s,a_{-i})$. If one of the following conditions is satisfied, then the Markov game is an average reward Markov potential game:
    \begin{enumerate}
        \item The transition probabilities do not depend on the action $\mathbf{a}$ taken, i.e., $P(s'|s,\mathbf{a})=P(s'|s)$ for all $\mathbf{a}$.
        \item For every agent $i$, there exist a constant $c_i$ such that $\overline{u}^{\pi_{-i}}_i(s):=\sum_{a_{-i}}(\Pi_{j\neq i}\pi_j(a_j|s))u_i(s,a_{-j})$ satisfies $\nabla_{\pi_i(\cdot|s)}\langle \nu^{\pi}, \overline{u}^{\pi_{-i}}_i \rangle = c_i\mathbf{1}_{A_i}$ for every state $s$ and policy $\pi$.
    \end{enumerate}
\end{proposition}

\begin{proof}
    Let $\overline{\phi}^{\pi}(s):=\sum_{\mathbf{a}}\pi(\mathbf{a}|s)\phi(s,\mathbf{a})$. $\rho_i^{\pi}=\langle \nu^{\pi},\overline{\phi}^{\pi}+\overline{u}_i^{\pi_{-i}} \rangle$. If condition 1 is satisfied, $\nu^{\pi}\equiv\nu$, then $ \langle \nu,\overline{\phi}^{\pi} \rangle$ is the potential function.

    If condition 2 is satisfied, $\rho_i^{\pi}=\langle \nu^{\pi},\overline{\phi}^{\pi} \rangle+\langle \nu^{\pi}, \overline{u}_i^{\pi_{-i}}\rangle$. With the interpolation of differential function, for any policies $\pi_i, \pi_i'$ and $\pi_{-i}$, there exist a constant $a\in[0,1]$, $\pi_i^*=\pi_i+a(\pi_i'-\pi_i)$, such that $\rho^{\pi_i,\pi_{-i}}-\rho_i^{\pi_i',\pi_{-i}}=\langle \nu^{\pi},\overline{\phi}^{\pi} \rangle - \langle \nu^{\pi_i',\pi_{-i}},\overline{\phi}^{\pi_i',\pi_{-i}} \rangle + \langle\pi_i-\pi_i', \nabla_{\pi_i(\cdot|s)}\langle \nu^{\pi_i^*,\pi_{-i}}, \overline{u}^{\pi_{-i}}_i \rangle \rangle =\langle \nu^{\pi},\overline{\phi}^{\pi} \rangle - \langle \nu^{\pi_i',\pi_{-i}},\overline{\phi}^{\pi_i',\pi_{-i}} \rangle + \langle\pi_i-\pi_i', c_i\mathbf{1}_{A_i}\rangle=\langle \nu^{\pi},\overline{\phi}^{\pi} \rangle - \langle \nu^{\pi_i',\pi_{-i}},\overline{\phi}^{\pi_i',\pi_{-i}} \rangle  $. Therefore, $\langle \nu^{\pi},\overline{\phi}^{\pi} \rangle$ is the potential function.
\end{proof}

\paragraph{Remark} A fully cooperative game, where all agents have the same reward function $r_i\equiv r$, is an important special case of a Markov potential game. It satisfies Condition 2 in \cref{prop:1} with $u_i\equiv 0$, while the lake usage problem \citep{dechert2006stochastic} satisfies Condition 1.

\begin{lemma}[Restatement of \cref{lm:2}]
    \begin{equation}
    \begin{aligned}
        \rho_i^{\hat{\pi}_j,\pi_{-j}}-\rho_i^{\tilde{\pi}_j,\pi_{-j}} &= \mathbb{E}_{s\sim\nu^{\hat{\pi}_j, \pi_{-j}}}[\langle\overline{Q}_{j:i}^{\tilde{\pi}_j,\pi_{-j}}(s,\cdot), \hat{\pi}_j(\cdot|s)-\tilde{\pi}_j(\cdot|s)\rangle_{\mathcal{A}_j}] \\ 
        &= \mathbb{E}_{s\sim\nu^{\hat{\pi}_j, \pi_{-j}}}\sum_{a_j}\hat\pi_j(a_j|s)\overline{A}_{j:i}^{\tilde{\pi}_j,\pi_{-j}}(s,a_j).
    \end{aligned}
    \end{equation}
\end{lemma}
\begin{proof}
    \begin{equation*}
        \begin{aligned}
            &\rho_i^{\hat{\pi}_j,\pi_{-j}}-\rho_i^{\tilde{\pi}_j,\pi_{-j}}\\
            =& \rho_i^{\hat{\pi}_j,\pi_{-j}}-\rho_i^{\tilde{\pi}_j,\pi_{-j}} + \mathbb{E}_{s'\sim\nu^{\hat{\pi}_j,\pi_{-j}}}[V_i^{\tilde{\pi}_j,\pi_{-j}}(s')] - \mathbb{E}_{s\sim\nu^{\hat{\pi}_j,\pi_{-j}}}[V_i^{\tilde{\pi}_j,\pi_{-j}}(s)]\\
            =&\mathbb{E}_{s\sim\nu^{\hat{\pi}_j,\pi_{-j}}, \textbf{a}\sim(\hat{\pi}_j,\pi_{-j})}[r_i(s,a)-\rho_i^{\tilde{\pi}_j,\pi_{-j}}+\mathbb{E}_{s'\sim P(\cdot|s,a)}[V_i^{\tilde{\pi}_j,\pi_{-j}}(s')] -V_i^{\tilde{\pi}_j,\pi_{-j}}(s)]\\
            =&\mathbb{E}_{s\sim\nu^{\hat{\pi}_j,\pi_{-j}}, \textbf{a}\sim(\hat{\pi}_j,\pi_{-j})}[Q_i^{\tilde{\pi}_j,\pi_{-j}}(s,\textbf{a}) - \langle\overline{Q}_{j;i}^{\tilde{\pi}_j,\pi_{-j}}(\cdot,s),\tilde{\pi}_j(\cdot|s)\rangle_{\mathcal{A}_j}]\\
            =&\mathbb{E}_{s\sim\nu^{\hat{\pi}_j,\pi_{-j}}}[\mathbb{E}_{a_j\sim\hat{\pi}_i}\overline{Q}_{j;i}^{\tilde{\pi}_j,\pi_{-j}}(s,a_j) - \langle\overline{Q}_{j;i}^{\tilde{\pi}_j,\pi_{-j}}(\cdot,s),\tilde{\pi}_j(\cdot|s)\rangle_{\mathcal{A}_j}]\\
            =&\mathbb{E}_{s\sim\nu^{\hat{\pi}_j, \pi_{-j}}}[\langle\overline{Q}_{j;i}^{\tilde{\pi}_j,\pi_{-j}}(s,\cdot), \hat{\pi}_j(\cdot|s)-\tilde{\pi}_j(\cdot|s)\rangle_{\mathcal{A}_j}]\\
            =&\mathbb{E}_{s\sim\nu^{\hat{\pi}_j, \pi_{-j}}}[\sum_{a_j}\overline{Q}_{j;i}^{\tilde{\pi}_j,\pi_{-j}}(s,a_j) \hat{\pi}_j(a_j|s)-V_i^{\tilde{\pi}_j,\pi_{-j}}(s)]\\
            =&\mathbb{E}_{s\sim\nu^{\hat{\pi}_j, \pi_{-j}}}[\sum_{a_j}\left(\overline{Q}_{j;i}^{\tilde{\pi}_j,\pi_{-j}}(s,a_j) -V_i^{\tilde{\pi}_j,\pi_{-j}}(s)\right)\hat{\pi}_j(a_j|s)]\\
            =&\mathbb{E}_{s\sim\nu^{\hat{\pi}_j, \pi_{-j}}}[\sum_{a_j}\hat{\pi}_j(a_j|s)\overline{A}_{j;i}^{\tilde{\pi}_j,\pi_{-j}}(s,a_j)].\\
        \end{aligned}
    \end{equation*}
\end{proof}

\begin{lemma}[Restatement of \cref{lm:3}]
    \begin{equation*}
        \frac{\partial \rho_i^{\pi}}{\partial \pi_j(a_j|s)} = \overline{Q}_{j;i}^{\pi}(s,a_j)\nu^{\pi}(s),\         \frac{\partial \Phi(\pi)}{\partial \pi_j(a_j|s)} = \overline{Q}_{j}^{\pi}(s,a_j)\nu^{\pi}(s)
    \end{equation*}
\end{lemma}
\begin{proof}
    Recall that $\overline{Q}_{j:i}^{\pi}(s,a_j)=\underset{a_{-j}\in\mathcal{A}_{-j}}{\sum} \pi_{-j}(a_{-j}|s)Q^{\pi}_i(s,a_j,a_{-j})=\underset{a_{-j}\in\mathcal{A}_{-j}}{\sum} \pi_{-j}(a_{-j}|s)(r(s,\textbf{a})-\rho^{\pi}_i+\mathbb{E}_{s'\sim P(\cdot|s,a)}V^{\pi}_i(s'))$. Therefore, $V_i^{\pi}(s)=\sum_{a_j}\pi_j(a_j|s)\overline{Q}_{j;i}^{\pi}(s,a_j)$. Differentiating with respect to $\pi_j$:
    \begin{equation*}
        \begin{aligned}
            \frac{\partial V_i^{\pi}(s)}{\partial\pi_j}&=\frac{\partial}{\partial\pi_j}\sum_{a_j}\pi_j(a_j|s)\overline{Q}_{j;i}^{\pi}(s,a_j)\\
            &=\sum_{a_j}\frac{\partial\pi_j(a_j|s)}{\partial\pi_j}\overline{Q}_{j;i}^{\pi}(s,a_j)+\sum_{a_j}\pi_j(a_j|s)\frac{\partial\overline{Q}_{j;i}^{\pi}(s,a_j)}{\partial\pi_j}\\
            &=\sum_{a_j}\frac{\partial\pi_j(a_j|s)}{\partial\pi_j}\overline{Q}_{j;i}^{\pi}(s,a_j)+\sum_{a_j}\pi_j(a_j|s)\frac{\partial}{\partial\pi_j}(r_i^{\pi_{-j}}(s,a_j)-\rho_i^{\pi}+\mathbb{E}_{s'\sim P(\cdot|s,a_j,a_{-j}),a_{-j}\sim\pi_{-j}(\cdot|s)}V^{\pi}_i(s'))\\
            &=\sum_{a_j}\frac{\partial\pi_j(a_j|s)}{\partial\pi_j}\overline{Q}_{j;i}^{\pi}(s,a_j)-\frac{\partial\rho_i^{\pi}}{\partial\pi_j}+\mathbb{E}_{s'\sim P(\cdot|s,a),\textbf{a}\sim\pi(\cdot|s)}\frac{\partial V^{\pi}_i(s')}{\partial \pi_j}.\\
        \end{aligned}
    \end{equation*}
    Multiplying each side with $\nu^{\pi}(s)$, taking the summation over $s$, with the definition of stationary distribution $ \mathbb{E}_{s\sim\nu^{\pi},s'\sim P(\cdot|s,a),a\sim\pi(\cdot|s)}=\mathbb{E}_{s'\sim\nu^{\pi}}$, we obtain:
    \begin{equation*}
        \begin{aligned}
            \sum_s\nu^{\pi}(s)\frac{\partial V_i^{\pi}(s)}{\partial\pi_j}=&\sum_s\nu^{\pi}(s)\sum_{a_j}\frac{\partial\pi_j(a_j|s)}{\partial\pi_j}\overline{Q}_{j;i}^{\pi}(s,a_j)-\frac{\partial\rho_i^{\pi}}{\partial\pi_j}+\mathbb{E}_{s'\sim\nu^{\pi}}\frac{\partial V^{\pi}_i(s')}{\partial \pi_j},\\
            0=&\sum_s\nu^{\pi}(s)\sum_{a_j}\frac{\partial\pi_j(a_j|s)}{\partial\pi_j}\overline{Q}_{j;i}^{\pi}(s,a_j)-\frac{\partial\rho_i^{\pi}}{\partial\pi_j},\\
            \frac{\partial\rho_i^{\pi}}{\partial\pi_j}=&\sum_s\nu^{\pi}(s)\sum_{a_j}\overrightarrow{\mathbf{e}}(s,a_j)\overline{Q}_{j;i}^{\pi}(s,a_j).
        \end{aligned}
    \end{equation*}

    By the definition of the potential function, it can be noted that $\frac{\partial\Phi(\pi)}{\partial\pi_j(a_j|s)}=\frac{\partial\rho_j^{\pi}}{\partial\pi_j(a_j|s)}=\overline{Q}_{j}^{\pi}(s,a_j)\nu^{\pi}(s)$.
\end{proof}

\begin{lemma}[\text{\citealp[Theorem 3.1, equation (4)]{kale2013eigenvalues}}]\label{lm:kale}
    Given a Markov chain with transition probability matrix $P\in\mathbb{R}^{S\times S}$, let $\nu$ be its stationary distribution. For any $s,s'\in\mathcal{S}$, we have:
    \begin{equation*}
        |\frac{P^t(s'|s)}{\nu(s')}-1|\leq\frac{\lambda_2(P)^t}{\sqrt{\nu(s)\nu(s')}}.
    \end{equation*}
\end{lemma}

\begin{lemma}[Restatement of \cref{lm:4}]
    There exist constants $C_p:=\min\{\sqrt{\frac{S}{1-\Gamma}},\frac{1}{1-\Gamma}\}$ and $\varrho:=1-\frac{1}{\kappa_0}$ such that for any policy $\pi$: 
    \begin{equation*}
        sup_s\lVert (P^{\pi})^t(\cdot|s_0=s)-\nu^{\pi}\rVert_1\leq C_p\varrho^t, \forall\ t>0.
    \end{equation*}
\end{lemma}
\begin{proof}
    For any policy $\pi\in\Pi$, $|\frac{(P^{\pi})^t(s'|s)}{\nu^{\pi}(s')}-1|\leq\frac{\lambda_2(P^{\pi})^t}{\sqrt{\nu^{\pi}(s)\nu^{\pi}(s')}}$ by \cref{lm:kale}. Then $\sum_{s'} |(P^{\pi})^t(s'|s)-\nu^{\pi}(s')|\leq \lambda_2(P^{\pi})^t\sum_{s'}\sqrt{\frac{\nu^{\pi}(s')}{\nu^{\pi}(s)}}\leq\left(1-(\frac{1}{1-\lambda_2(P^{\pi})})^{-1} \right)^t\sum_{s'}\sqrt{\nu^{\pi}(s')}\frac{1}{\sqrt{1-\Gamma}}\leq(1-\frac{1}{\kappa_0})^t\sqrt{\frac{S}{1-\Gamma}}$.

    Similarly, recall $|\frac{(P^{\pi})^t(s'|s)}{\nu^{\pi}(s')}-1|\leq\frac{\lambda_2(P^{\pi})^t}{\sqrt{\nu^{\pi}(s)\nu^{\pi}(s')}}\leq\frac{\lambda_2(P^{\pi})^t}{1-\Gamma}$, then $\sum_{s'} |(P^{\pi})^t(s'|s)-\nu^{\pi}(s')|\leq \frac{\lambda_2(P^{\pi})^t}{1-\Gamma}=(1-\frac{1}{\kappa_0})^2\frac{1}{1-\Gamma}$.
\end{proof}

\begin{proposition}\label{prop:3} For any policy $\pi$ and any agent $i$, the differential Q function has a bounded $\ell_{\infty}$ norm:
    \begin{equation*}
        \lVert Q^{\pi}_i\rVert_{\infty}\leq C_p\kappa_0.
    \end{equation*}
\end{proposition}
\begin{proof}
    For any $s,\mathbf{a}$ and joint policy $\pi$, $|Q^{\pi}_i(s,\mathbf{a})|=|\mathbb{E}_{\pi}[\sum_{t=0}^{\infty}(r_i(s^t,\mathbf{a}^t)-\rho_i^{\pi})|s^0=s,\mathbf{a}^0=\mathbf{a}]|\leq 1+|\mathbb{E}_{s'\sim P(\cdot|s,\mathbf{a})}\sum_{t=1}^{\infty}\langle(P^{\pi})^t(\cdot|s')-\nu^{\pi},r^{\pi}\rangle|\leq1+\sum_{t=1}^{\infty}C_p\varrho^t= 1+\frac{C_p\varrho}{1-\varrho}=1+C_p(\kappa_0-1)<C_p\kappa_0$. 
\end{proof}

\section{PROOFS OF SECTION 3 }
\subsection{Auxiliary Lemmas} 
\begin{lemma}[\text{\citealp[Lemma 4.1]{leonardos2021global}}] \label{lm:14}
    Let $\pi=(\pi_1,\pi_2,...,\pi_N)$ be the policy profile for all agents and let $\pi'=\pi+\lambda\nabla_{\pi}\Phi(\pi)$ be the result from a gradient step on the potential function with learning rate $\lambda>0$. Then
    \begin{equation*}
        \textnormal{Proj}_{\Pi_1\times...\times\Pi_N}(\pi')=(\textnormal{Proj}_{\Pi_1}(\pi_1'),...,\textnormal{Proj}_{\Pi_N}(\pi_N')).
    \end{equation*}
\end{lemma}
\begin{lemma}[\text{\citealp[Proposition B.1]{agarwal2021theory}}]\label{lm:15}
    Let $f(\pi)$ be an $l$-smooth function. Define the gradient mapping $G(\pi):=\frac{1}{\beta}(\textnormal{Proj}_{\Pi}(\pi+\beta\nabla_{\pi}f(\pi))-\pi)$. Then the update rule for the projected gradient is $\pi^+=\pi+\beta G(\pi)$. If $\lVert G(\pi)\rVert_2\leq\epsilon$, then
    \begin{equation*}
        \underset{\pi^++\delta\in\Pi,\lVert\delta\rVert_2\leq1}{\max}\delta^T\nabla_{\pi}f(\pi^+)\leq \epsilon(\beta l+1).
    \end{equation*}
\end{lemma}
There is a typographical mistake in proposition B.1 of \cite{agarwal2021theory}; the underlying max should be taken on $ \pi^++\delta\in\Pi$ instead of $\pi+\delta\in\Pi$.

\begin{lemma}[\text{\citealp[Lemma 3.6]{bubeck2015convex}}]\label{lm:16}
    Let $f$ be an $l$-smooth function over a convex domain $C$. Let $x\in C$, $x^+=\textnormal{Proj}_C(x-\beta\nabla f(x))$. Then :
    \begin{equation*}
        f(x^+)-f(x)\leq(\frac{l}{2}-\frac{1}{\beta})\lVert x-x^+\rVert_2^2 .
    \end{equation*}
\end{lemma}

\begin{lemma}\label{lm:17}
    Assume function $f(\cdot)$ is $l$-smooth over a convex set $C$. Let $x^+:=P_C(x-\beta g)$ with $x\in C$. Then 
    \begin{equation*}
        f(x^+)-f(x)\leq (\frac{3l}{4}-\frac{1}{\beta})\lVert x-x^+\rVert_2^2+\frac{1}{l}\lVert \nabla_xf(x)-g\rVert_2^2.
    \end{equation*}
\end{lemma}
\begin{proof}
    \begin{align*}
        f(x^+)-f(x)\leq&\langle\nabla f(x),x^+-x\rangle+\frac{l}{2}\lVert x^+-x\rVert_2^2\\
        =&\langle g,x^+-x\rangle+\frac{l}{2}\lVert x^+-x\rVert_2^2+\langle\nabla f(x)-g,x^+-x\rangle\\
        \overset{(a)}{\leq}&-\frac{1}{\beta}\lVert x-x^+\rVert_2^2+\frac{l}{2}\lVert x^+-x\rVert_2^2+\frac{1}{2}(\frac{2}{l}\lVert\nabla_xf(x)-\hat\nabla_xf(x)\rVert_2^2+\frac{l}{2}\lVert x^+-x\rVert_2^2)\\
        =&(\frac{3l}{4}-\frac{1}{\beta})\lVert x-x^+\rVert_2^2+\frac{1}{l}\lVert \nabla_xf(x)-g\rVert_2^2.
    \end{align*}
    We use the property of projection in (a) $\langle x^+-(x-\beta g),x^+-x \rangle\leq 0$ and $\langle x,y\rangle\leq\frac{\frac{1}{a}\lVert x\rVert_2^2+a\lVert y\rVert_2^2}{2}$ for any positive constant $a$.
\end{proof}

\subsection{Proofs of Lemma 4 and Theorem 1}\label{B.2}
\begin{lemma}[Restatement of \cref{lm:smooth}]
Denote $A_{\max}:=\max_iA_i$, $L:=\kappa_0^2S^{3/2}A_{\max}+\kappa_0\sqrt{S}A_{\max}$, and $L_{\Phi}:=N(\kappa_0^2S^{3/2}A_{\max}+\kappa_0(SA_{\max}+2A_{\max})+A_{\max})$. 

(a) For any i, and $\pi_{-i}\in\Pi_{-i}$, the average value $\rho_i^{\pi}$ is $\kappa_0^2S^{3/2}A_i+\kappa_0\sqrt{S}A_i$-smooth with respect to policy $\pi_i$. Moreover, for any $i$, $\rho_i^{\pi}$ is $L$-smooth with respect to policy $\pi_i$, i.e. $\lVert\nabla_{\pi_i}\rho_i^{\pi_i,\pi_{-i}}-\nabla_{\pi_i}\rho_i^{\pi_i',\pi_{-i}}\rVert_2\leq L\lVert\pi_i-\pi_i'\rVert_2$ for $\forall\ i,\ and\ \pi_i,\pi_i'\in\Pi_i$.

(b) The potential function $\Phi(\pi)$ is $L_{\Phi}$-smooth with respect to policy profile $\pi$, i.e. $\lVert\nabla\Phi(\pi)-\nabla\Phi(\pi')\rVert_2\leq L_{\Phi}\lVert\pi-\pi'\rVert_2$ for $\forall\ \pi,\pi'\in\Pi$.
\end{lemma}

\begin{proof}
    By the potential function property $\Phi(\pi_i,\pi_{-i})-\Phi(\pi_i',\pi_{-i})=\rho_i^{\pi_i,\pi_{-i}}-\rho_i^{\pi_i',\pi_{-i}}$, $\frac{\partial}{\partial \pi_i}\Phi(\pi_i,\pi_{-i})=\frac{\partial}{\partial \pi_i}\rho_i^{\pi_i,\pi_{-i}}$, and $\nabla_{\pi}\Phi=(\frac{\partial\rho^{\pi}_1}{\partial\pi_1},...,\frac{\partial\rho^{\pi}_N}{\partial\pi_N})^T$. To show the smoothness,
    \begin{align*}
        &\lVert\nabla_{\pi}\Phi(\pi)-\nabla_{\pi}\Phi(\pi')\rVert_2^2\\
        =&\sum_{i=1}^N\lVert\nabla_{\pi_i}\Phi(\pi)-\nabla_{\pi_i}\Phi(\pi')\rVert_2^2\\
        =&\sum_{i=1}^N\lVert\nabla_{\pi_i}\Phi(\pi)-\nabla_{\pi_i}\Phi(\pi_1',\pi_{2\sim N})+\nabla_{\pi_i}\Phi(\pi_1',\pi_{2\sim N})-\nabla_{\pi_i}\Phi(\pi_{1,2}',\pi_{3\sim N})+\hdots \\
        &+\nabla_{\pi_i}\Phi(\pi_{1\sim(N-1)},\pi'_N)-\nabla_{\pi_i}\Phi(\pi')\rVert_2^2\\
        \leq&\sum_{i=1}^N\sum_{j=1}^NN\lVert\nabla_{\pi_i}\Phi(\pi_{1\sim j-1},\pi'_{j\sim N})-\nabla_{\pi_i}\Phi(\pi_{1\sim j},\pi'_{j+1\sim N})\rVert_2^2\\
        =&\sum_{i=1}^N\sum_{j=1}^NN\lVert\frac{\partial\rho_i^{\pi_{1\sim j-1},\pi'_{j\sim N}}}{\partial\pi_i}-\frac{\partial\rho_i^{\pi_{1\sim j},\pi'_{j+1\sim N}}}{\partial\pi_i}\rVert_2^2.\\
    \end{align*}

    If we can show $\frac{\partial\rho_i^{\pi}}{\partial\pi_i}$ is $\frac{L_{\Phi}}{N}$-lipschitz for any $\pi_j$ within $\Pi_j$, the above inequality implies that $\lVert\nabla_{\pi}\Phi(\pi)-\nabla_{\pi}\Phi(\pi')\rVert_2^2\leq\sum_{i=1}^N\sum_{j=1}^N\frac{L_{\Phi}^2}{N}\lVert\pi_j-\pi_j'\rVert_2^2=L_{\Phi}^2\lVert\pi-\pi'\rVert_2^2$. In the following proof, we will fix $i$ and denote $\rho=\rho_i$.
    
    Define $\rho(\epsilon)=\rho_i^{\pi_i+\epsilon u_i,\pi_{-i}}$ and $\rho(\tau,\epsilon)=\rho_i^{\pi_i+\epsilon u_i,\pi_j+\tau u_j,\pi_{-ij}}$. We wish to show that $|\frac{d^2\rho}{d\epsilon^2}|\leq L$, $|\frac{d^2\rho}{d\epsilon d\tau}|\leq \frac{L_{\Phi}}{N}$ for any $u_i, u_j$, and any $\epsilon, \tau\in\mathbb{R} $ such that $\pi_i+\epsilon u_i\in\Pi_i$, $\lVert u_i\rVert_2\leq 1$, $\pi_j+\tau u_j\in\Pi_j$, $\lVert u_j\rVert_2\leq 1$ and any $\pi\in\Pi$. Note that $\sum_{a_i}u_i(a_i|s)=0$, $\sum_{a_j}u_j(a_j|s)=0$ for any $s$.

    We first note that the unit eigenvector $X(\epsilon)$ of the perturbed square matrix $P^{\pi}+\epsilon P^{u_i}$ corresponding to the simple eigenvalue $1$ is an analytic function \cite{kazdan1995matrices}. Recall that for any $\epsilon$, $X(\epsilon)=c\nu^{\pi_i+\epsilon u_i,\pi_{-i}}$ for some constant $c\neq0$, thus $\sum_{s\in\mathcal{S}} X_s(\epsilon)\neq0$.  $\nu^{\pi_i+\epsilon u_i,\pi_{-i}}=(\sum_sX_s(\epsilon))^{-1} X(\epsilon)$ is therefore analytic, which guarantees the existence of directional derivatives of both $\nu^{\pi_i+\epsilon u_i,\pi_{-i}}$ and $\rho^{\epsilon}$. Then by \cite{garrett2005hartogs}, $\nu^{\epsilon,\tau}=\nu^{\pi_i+\epsilon u_i,\pi_j+\tau u_j,\pi_{-ij}}$ is jointly analytic, and thus $\frac{d^2\nu^{\pi_i+\epsilon u_i,\pi_j+\tau u_j,\pi_{-ij} }}{d\epsilon d\tau}$ and $\frac{d^2\rho}{d\epsilon d\tau}$ exist.


    We first show $|\frac{d^2\rho}{d\epsilon^2}|\leq L$. Since $\rho(\epsilon)=\sum_{s,\textbf{a}}\nu^{\pi}(s)\pi(\textbf{a}|s)r(s,\textbf{a})=\sum_{s,a_i}\nu^{\pi_i+\epsilon u_i,\pi_{-i}}(s)(\pi_i(a_i|s)+\epsilon u_i(a_i|s))r^{\pi_{-i}}(s,a_i)$, we have
    \begin{equation}\label{eq:9}
        \begin{aligned}
            \frac{d\rho(\epsilon) }{d\epsilon} &= \sum_{s,a_i}\nu^{\pi_i+\epsilon u_i,\pi_{-i}}(s)u_i(a_i|s)r^{\pi_{-i}}(s,a_i)+\sum_{s,a_i}\frac{d\nu^{\pi_i+\epsilon u_i,\pi_{-i}}(s)}{d\epsilon}(\pi_i(a_i|s)+\epsilon u_i(a_i|s))r^{\pi_{-i}}(s,a_i), \\ 
            \frac{d^2\rho(\epsilon)}{d\epsilon^2} &= 2\sum_{s,a_i}\frac{d\nu^{\pi_i+\epsilon u_i,\pi_{-i}}(s)}{d \epsilon}u_i(a_i|s)r^{\pi_{-i}}(s,a_i)+\sum_{s,a_i}\frac{d^2\nu^{\pi_i+\epsilon u_i,\pi_{-i}}(s)}{d\epsilon^2}(\pi_i(a_i|s)+\epsilon u_i(a_i|s))r^{\pi_{-i}}(s,a_i) .
        \end{aligned}
    \end{equation}

    Since the stationary distribution satisfies $\nu^{\pi_i+\epsilon u_i,\pi_{-i}}\overrightarrow{\mathbf{1}}\equiv1$, we have $\frac{d\nu^{\pi_i+\epsilon u_i,\pi_{-i}}}{d\epsilon}\overrightarrow{\mathbf{1}}=0$ and $\frac{d^2\nu^{\pi_i+\epsilon u_i,\pi_{-i}}}{d\epsilon^2}\overrightarrow{\mathbf{1}}=0$. In other words, $\frac{d\nu^{\pi_i+\epsilon u_i,\pi_{-i}}}{d\epsilon}$ and $\frac{d^2\nu^{\pi_i+\epsilon u_i,\pi_{-i}}}{d\epsilon^2}$ are orthogonal to $\overrightarrow{\mathbf{1}}$. Taking derivatives on both sides of $\nu^{\pi_i+\epsilon u_i,\pi_{-i}} = \nu^{\pi_i+\epsilon u_i,\pi_{-i}}P^{\pi_i+\epsilon u_i,\pi_{-i}}$ gives
    \begin{equation}\label{eq:10}
        \begin{aligned}
            &\frac{d\nu^{\pi_i+\epsilon u_i,\pi_{-i}}}{d\epsilon} = \nu^{\pi_i+\epsilon u_i,\pi_{-i}}\frac{dP^{\pi_i+\epsilon u_i,\pi_{-i}}}{d\epsilon}+\frac{d\nu^{\pi_i+\epsilon u_i,\pi_{-i}}}{d\epsilon}P^{\pi_i+\epsilon u_i,\pi_{-i}}, \\ 
            &\frac{d^2\nu^{\pi_i+\epsilon u_i,\pi_{-i}}}{d\epsilon^2} = \nu^{\pi_i+\epsilon u_i,\pi_{-i}}\frac{d^2P^{\pi_i+\epsilon u_i,\pi_{-i}}}{d\epsilon^2}+2\frac{d\nu^{\pi_i+\epsilon u_i,\pi_{-i}}}{d\epsilon}\frac{dP^{\pi_i+\epsilon u_i,\pi_{-i}}}{d\epsilon}+\frac{d^2\nu^{\pi_i+\epsilon u_i,\pi_{-i}}}{d\epsilon^2}P^{\pi_i+\epsilon u_i,\pi_{-i}}. \\ 
        \end{aligned}
    \end{equation}
    
    By \cref{def:4} and the fact that $\frac{d\nu^{\pi_i+\epsilon u_i,\pi_{-i}}}{d\epsilon}$ is orthogonal to the all-1 vector $\mathbf{1}$,
    \begin{equation*}
        \begin{aligned}
             &\lVert\frac{d\nu^{\pi_i+\epsilon u_i,\pi_{-i}}}{d\epsilon}\rVert_2\leq \lVert \nu^{\pi_i+\epsilon u_i,\pi_{-i}}\frac{dP^{\pi_i+\epsilon u_i,\pi_{-i}}}{d\epsilon}\rVert_2+ |\lambda_2(P^{\pi_i+\epsilon u_i,\pi_{-i}})|\lVert\frac{d\nu^{\pi_i+\epsilon u_i,\pi_{-i}}}{d\epsilon}\rVert_2,
        \end{aligned}      
    \end{equation*}
    which implies $\lVert\frac{d\nu^{\pi_i+\epsilon u_i,\pi_{-i}}}{d\epsilon}\rVert_2\leq \kappa_0\lVert \nu^{\pi_i+\epsilon u_i,\pi_{-i}}\frac{dP^{\pi_i+\epsilon u_i,\pi_{-i}}}{d\epsilon}\rVert_2$. 

    Denote $\overline{P}^{\pi_{-i}}(s'|s,a_i):=\sum_{a_{-i}}\pi_{-i}(a_{-i}|s)P(s'|s,a_i,a_{-i})$. We have $\frac{dP^{\pi_i+\epsilon u_i,\pi_{-i}}}{d \epsilon}(s'|s)=\sum_{a_i}u_i(a_i|s)\overline{P}^{\pi_{-i}}(s'|s,a_i)$, since $P^{\pi_i+\epsilon u_i,\pi_{-i}}(s'|s)=\sum_{a_i}\sum_{a_{-i}}\pi_{-i}(a_{-i}|s)(\pi_i(a_i|s)+\epsilon u_i(a_i|s))P(s'|s,a)$. For any $\nu\in\Delta(S)$,
    \begin{equation*}
    \begin{aligned}
        \lVert \nu \frac{dP^{\pi_i+\epsilon u_i,\pi_{-i}}}{d \epsilon}\rVert_2^2=&\sum_{s'}(\sum_s\nu(s)\sum_{a_i}\overline{P}^{\pi_{-i}}(s'|s,a_i)u(a_i|s))^2\\
        \leq&\sum_{s'}(max_s|\sum_{a_i}\overline{P}^{\pi_{-i}}(s'|s,a_i)u(a_i|s)|)^2\overset{(a)}{\leq}
        \sum_{s'}(\frac{1}{2}max_s\lVert u(\cdot|s)\rVert_1)^2\overset{(b)}{\leq}\frac{SA_i}{4}. 
    \end{aligned}
    \end{equation*}
    $(a)$ follows from $\sum_{a_i}u(a_i|s)=0$ for any $s$ and the fact that $|\sum_{a_i}\overline{P}(s'|s,a_i)u(a_i|s)|$ is dominated by either the positive part or the negative part of $u(\cdot|s)$. $(b)$ is due to $\lVert u(\cdot|s)\rVert_1\leq \sqrt{A_i}\lVert u(\cdot|s)\rVert_2$ and $\lVert u(\cdot|s) \rVert_2\leq 1$. Thus, 
    \begin{align*}
        \lVert\frac{d\nu^{\pi_i+\epsilon u_i,\pi_{-i}}}{d\epsilon}\rVert_2\leq&\kappa_0\lVert \nu^{\pi_i+\epsilon u_i,\pi_{-i}}\frac{dP^{\pi_i+\epsilon u_i,\pi_{-i}}}{d \epsilon}\rVert_2\leq\frac{\kappa_0\sqrt{SA_i}}{2}.
    \end{align*}

    Since $\frac{d^2P^{\pi_i+\epsilon u_i,\pi_{-i}}}{d \epsilon^2}=0$, we have $\lVert\frac{d^2\nu^{\pi_i+\epsilon u_i,\pi_{-i}}}{d\epsilon^2}\rVert_2\leq 2\kappa_0\lVert \frac{d\nu^{\pi_i+\epsilon u_i,\pi_{-i}}}{d\epsilon}\frac{dP^{\pi_i+\epsilon u_i,\pi_{-i}}}{d \epsilon}\rVert_2$ by \cref{eq:10}.
    \begin{equation}\label{eq:11}
        \begin{aligned}
        \lVert \frac{d\nu^{\pi_i+\epsilon u_i,\pi_{-i}}}{d\epsilon}\frac{dP^{\pi_i+\epsilon u_i,\pi_{-i}}}{d \epsilon}\rVert_2
        &\overset{(a)}{\leq}\sqrt{\sum_{s'}\lVert\frac{d\nu^{\pi_i+\epsilon u_i,\pi_{-i}}}{d\epsilon}\rVert_2^2\sum_s(\sum_{a_i}\overline{P}^{\pi_{-i}}(s'|s,a_i)u(a_i|s))^2}\\
        &\leq\sqrt{\sum_{s'}\lVert\frac{d\nu^{\pi_i+\epsilon u_i,\pi_{-i}}}{d\epsilon }\rVert_2^2\sum_sA_i\sum_{a_i}u(a_i|s)^2}  \\
        &\overset{(b)}{\leq}\sqrt{\sum_{s'}\lVert\frac{d\nu^{\pi_i+\epsilon u_i,\pi_{-i}}}{d\epsilon}\rVert_2^2A_i}\leq\frac{\kappa_0SA_i}{2},
        \end{aligned}
    \end{equation}
    where $(a)$ follows from $\langle x,y\rangle\leq\lVert x\rVert_2\lVert y\rVert_2$ and $(b)$ is due to $\lVert u\rVert_2\leq 1$. Thus $\lVert\frac{d^2\nu^{\pi_i+\epsilon u_i,\pi_{-i}}}{d\epsilon^2}\rVert_2\leq\kappa_0^2SA_i$. 
    
    Substituting the above inequalities back to $\frac{d^2\rho(\epsilon)}{d\epsilon^2}$ in \cref{eq:9},
    \begin{equation*}
        |\frac{d^2\rho(\epsilon)}{d\epsilon^2}| \overset{(a)}{\leq} 2\lVert\frac{d\nu^{\pi_i+\epsilon u_i,\pi_{-i}}}{d \epsilon}\rVert_2\sqrt{A_i}\lVert u_i\rVert_2+\lVert\frac{d^2\nu^{\pi_i+\epsilon u_i,\pi_{-i}}}{d\epsilon^2}\rVert_1 \leq \kappa_0\sqrt{S}A_i+\kappa_0^2S^{3/2}A_i\leq L.
    \end{equation*}
    $(a)$ is due to $\sum_{s,a_i}\frac{d\nu^{\pi_i+\epsilon u_i,\pi_{-i}}(s)}{d \epsilon}u_i(a_i|s)r^{\pi_{-i}}(s,a_i)\leq\lVert\frac{d\nu^{\pi_i+\epsilon u_i,\pi_{-i}}}{d \epsilon}\rVert_2\sqrt{\sum_{s}(\sum_{a_i}u_i(a_i|s)r^{\pi_{-i}}(s,a_i))^2}\leq\lVert\frac{d\nu^{\pi_i+\epsilon u_i,\pi_{-i}}}{d \epsilon}\rVert_2\sqrt{\sum_{s}\lVert u_i(\cdot|s)\rVert_1^2}$. 
    
    We will next show $|\frac{d^2\rho}{d\epsilon d\tau}|\leq \frac{L_{\Phi}}{N}$. Use $\nu^{\epsilon,\tau}:=\nu^{\pi_i+\epsilon u_i,\pi_j+\tau u_j,\pi_{-ij}}$, $P^{\epsilon,\tau}:=P^{\pi_i+\epsilon u_i,\pi_j+\tau u_j,\pi_{-ij}} $. Differentiating twice gives
    \begin{equation*}
    \begin{aligned}
        \frac{d^2\rho(\epsilon,\tau)}{d\epsilon d\tau}=&\sum_{s,\mathbf{a}}\frac{d^2\nu^{\epsilon,\tau}(s)}{d\epsilon d\tau}(\pi_i(a_i|s)+\epsilon u_i(a_i|s))(\pi_j(a_j|s)+\tau u_j(a_j|s))\pi_{-ij}(a_{-ij}|s)r(s,\mathbf{a})\\
        &+\sum_{s,\mathbf{a}}\frac{d\nu^{\epsilon,\tau}(s)}{d\epsilon}(\pi_i(a_i|s)+\epsilon u_i(a_i|s))u_j(a_j|s)\pi_{-ij}(a_{-ij}|s)r(s,\mathbf{a})\\
        &+\sum_{s,\mathbf{a}}\frac{d\nu^{\epsilon,\tau}(s)}{d\tau}u_i(a_i|s)(\pi_j(a_j|s)+\tau u_j(a_j|s))\pi_{-ij}(a_{-ij}|s)r(s,\mathbf{a})\\
        &+\sum_{s,\mathbf{a}}\nu^{\epsilon,\tau}(s)u_i(a_i|s)u_j(a_j|s)\pi_{-ij}(a_{-ij}|s)r(s,\mathbf{a}),\\
        |\frac{d^2\rho(\epsilon,\tau)}{d\epsilon d\tau}|\leq&\lVert\frac{d^2\nu^{\epsilon,\tau}}{d\epsilon d\tau}\rVert_1+ 2\sqrt{A_j}\lVert\frac{d\nu^{\epsilon,\tau}}{d\epsilon}\rVert_2\lVert u_j\rVert_2+2\sqrt{A_i}\lVert\frac{d\nu^{\epsilon,\tau}}{d\tau}\rVert_2 \lVert u_i\rVert_2+\lVert\nu^{\epsilon,\tau}\rVert_1\max_s\lVert u_i(\cdot|s)\rVert_1\max_s\lVert u_j(\cdot|s)\rVert_1\\
        \leq&\sqrt{S}\lVert\frac{d^2\nu^{\epsilon,\tau}}{d\epsilon d\tau}\rVert_2+ 2\sqrt{A_j}\lVert\frac{d\nu^{\epsilon,\tau}}{d\epsilon}\rVert_2+2\sqrt{A_i}\lVert\frac{d\nu^{\epsilon,\tau}}{d\tau}\rVert_2+\sqrt{A_iA_j}.\\
    \end{aligned}
    \end{equation*}

    Similarly, as before, $\frac{d^2\nu^{\epsilon,\tau}}{d\epsilon d\tau}=\nu^{\epsilon,\tau}\frac{d^2P^{\epsilon,\tau}}{d\epsilon d\tau}+(\frac{d\nu^{\epsilon,\tau}}{d\tau})(\frac{dP^{\epsilon,\tau}}{d\epsilon})+(\frac{d\nu^{\epsilon,\tau}}{d\epsilon})(\frac{dP^{\epsilon,\tau}}{d\tau})+(\frac{d^2\nu^{\epsilon,\tau}}{d\epsilon d\tau})P^{\epsilon,\tau}$, $\lVert\frac{d\nu^{\epsilon,\tau}}{d\epsilon}\rVert_2\leq\frac{\kappa_0\sqrt{SA_i}}{2}$, $\lVert\frac{d\nu^{\epsilon,\tau}}{d\tau}\rVert_2\leq\frac{\kappa_0\sqrt{SA_j}}{2}$, $\frac{d^2P^{\epsilon,\tau}(s'|s)}{d\epsilon d\tau}=\sum_{a_i,a_j} \overline P^{\pi_{-ij}}(s'|s,a_i,a_j)u_i(a_i|s)u_j(a_j|s)$. For any $\nu\in\Delta(S)$:

    \begin{align*}
        \lVert \nu\frac{d^2P^{\epsilon,\tau}}{d\epsilon d\tau}\rVert_2^2&=\sum_{s'}(\sum_s\nu(s)\sum_{a_i,a_j}P(s'|s,a_i,a_j)u_i(a_i|s)u_j(a_j|s))^2\\
        &\leq \sum_{s'}(\max_s|\sum_{a_i,a_j}P(s'|s,a_i,a_j)u_i(a_i|s)u_j(a_j|s)|)^2\\
        &\leq\sum_{s'}(\max_s\sum_{a_i,a_j}|u_i(a_i|s)u_j(a_j|s)|)^2\\
        &\leq S (\max_s\lVert u_i(\cdot|s)\rVert_1\max_s\lVert u_j(\cdot|s)\rVert_1)^2\leq SA_iA_j.
    \end{align*}
    For the last inequality, we use $\lVert x\rVert_1\leq \sqrt{n}\lVert x\rVert_2$ and $\lVert u_i\rVert_2\leq1$, $\lVert u_j\rVert_2\leq 1$.
    \begin{equation*}
    \begin{aligned}
        \lVert\frac{d^2\nu^{\epsilon,\tau}}{d\epsilon d\tau}\rVert_2&\leq\kappa_0\left(\lVert \nu^{\epsilon,\tau}\frac{d^2P^{\epsilon,\tau}}{d\epsilon d\tau}\rVert_2+\lVert\frac{d\nu^{\epsilon,\tau}}{d\tau}\frac{dP^{\epsilon,\tau}}{d\epsilon}\rVert_2+\lVert\frac{d\nu^{\epsilon,\tau}}{d\epsilon}\frac{dP^{\epsilon,\tau}}{d\tau}\rVert_2\right)\\
        &\leq\kappa_0\left(\lVert \nu^{\epsilon,\tau}\frac{d^2P^{\epsilon,\tau}}{d\epsilon d\tau}\rVert_2+ \frac{\kappa_0SA_i}{2}+\frac{\kappa_0 SA_j}{2}\right)\ (apply\ \cref{eq:11})\\ 
        &\leq\kappa_0(\sqrt{SA_iA_j}+\frac{\kappa_0S(A_i+A_j)}{2})\\
        |\frac{d^2\rho(\epsilon,\tau)}{d\epsilon d\tau}|&\leq\kappa_0(S\sqrt{A_iA_j}+\frac{\kappa_0S^{3/2}(A_i+A_j)}{2}+2\sqrt{SA_iA_j})+\sqrt{A_iA_j}\leq \frac{L_{\Phi}}{N}.\\
    \end{aligned}
    \end{equation*}

    Therefore, the $L_{\Phi}$-smoothness follows.
\end{proof}

\begin{theorem}[Restatement of \cref{thm:1}]
    Choose learning rate $\beta=\frac{1}{L_{\Phi}}$. Then Nash-regret* of \cref{alg:1} is bounded by
    \begin{align*}
        \textnormal{Nash-regret}^*(T)\leq&\frac{32D^2SC_{\Phi}N(\kappa_0^2S^{5/2}A_{\max}+\kappa_0(SA_{\max}+2A_{\max})+A_{\max})}{T}\\
        =&O(\frac{D^2\kappa_0^2S^{5/2}A_{\max}NC_{\Phi}}{T}).
    \end{align*}
\end{theorem}
\begin{proof}
    First, we bound the Nash-gap for time $t$:
    \begin{align*}
        \text{Nash-gap}(t)&\overset{(a)}{=}\rho^{\pi_i^{t,*},\pi_{-i}^t}_i-\rho_i^{\pi^t}\\
        &=E_{s\sim\nu^{\pi_i^{t,*},\pi_{-i}^{t}}}[\langle\overline{Q}_i^{\pi_t}(s,\cdot),\pi_{i}^{t,*}(\cdot|s)-\pi_i^t(\cdot|s)\rangle] \\
        &\leq \sum_s \nu^{\pi_i^{t,*},\pi_{-i}^t}(s) \left(\max_{\pi'(\cdot|s)\in\Delta(A_i)}\langle\overline{Q}_i^{\pi^t}(s,\cdot),\pi'(\cdot|s)-\pi_i^t(\cdot|s)\rangle\right) \\
        &\overset{(b)}{\leq} \max_s\frac{\nu^{\pi_i^{t,*},\pi_{-i}^t}(s)}{\nu^{\pi^t}(s)}\langle\nabla_{\pi_i}\Phi(\pi^t),\pi'-\pi_i^t\rangle \\
        &\overset{(c)}{\leq} 2D\sqrt{S}\underset{\pi_i^t+\delta\in\Pi_i,\lVert\delta\rVert_2\leq1}{\max}\delta^T\nabla_{\pi_i}\Phi(\pi^t) \\
        &\overset{(d)}{\leq} 2D\sqrt{S}\lVert\pi_i^t-\pi_i^{t-1}\rVert_2(L+L_{\Phi}) \\
        &\leq 2(1+\frac{1}{N})D\sqrt{S}\lVert\pi^t-\pi^{t-1}\rVert_2L_{\Phi}. \\
    \end{align*}
    In (a) we use $i$ to denote the agent that achieves the maximum of Nash-gap, i.e., $i\in \arg\max_i\ \max_{p\in\Pi_i}(\rho^{p,\pi_{-i}^t}-\rho_i^{\pi_i^t,\pi_{-i}^t})$, and $\pi_i^{t,*}\in \arg\max_{p\in\Pi_i}(\rho^{p,\pi_{-i}^t}-\rho_i^{\pi_i^t,\pi_{-i}^t})$. (b) is true since $\max_{\pi'(\cdot|s)\in\Delta(A_i)}\langle\overline{Q}_i^{\pi_t}(s,\cdot),\pi'(\cdot|s)-\pi_t^i(\cdot|s)\rangle\geq0$ and $\frac{\partial\Phi(\pi^t)}{\pi_i(a_i|s)}=\nu^{\pi^t}(s)\overline{Q}^{\pi^t}_i(s,a_i)$. We also use $\pi'$ to represent the policy that achieves the maximum. (c) is due to $\lVert\pi'-\pi_i^t\rVert_2^2\leq\sum_s\lVert\pi'(\cdot|s)-\pi_i^t(\cdot|s)\rVert_1^2\leq 4S$. (d) is obtained from \cref{lm:15} and that $\Phi(\pi_i,\pi_{-i})$ is $L$-smooth w.r.t. $\pi_i$ for any $\pi_{-i}\in\Pi_{-i}$.

    Since $1+\frac{1}{N}\leq 2$, we can bound the Nash-regret* as follows
    \begin{align*}
        \sum_{t=1}^T\text{Nash-gap}(t)^2&\leq 16D^2L_{\Phi}^2S\sum_{t=1}^T\lVert\pi_t-\pi_{t-1}\rVert_2^2 \\
        &\overset{(a)}{\leq}16D^2L_{\Phi}^2S\sum_{t=1}^T\frac{2}{L_{\Phi}}(\Phi(\pi_{t-1})-\Phi(\pi_{t})) \\
        &=32D^2L_{\Phi}S(\Phi(\pi_0)-\Phi(\pi_T)) \\
        &\leq 32D^2L_{\Phi}C_{\Phi}S, \\
        \text{Nash-regret}^*(T)&\leq\frac{32D^2L_{\Phi}C_{\Phi}S}{T}.
    \end{align*}

    (a) is due to \cref{lm:16} by applying $f=-\Phi$.
\end{proof}

\subsection{Proof of Lemma 5 and Theorem 2}\label{B.3}
\begin{lemma}[Restatement of \cref{lm:6}]
    For any agent $i$, consider the gradient estimate $\hat{g}_i$ defined in \cref{alg:2.1}. Given the $(s, a, r)$-trajectory of length $KN_2+N_1$ and the policy $\pi_i\in\Pi_{i,\alpha}$ that generated it, under the assumption that all the other policies $\pi_{-i}$ are fixed during the generation of the trajectory, the estimated gradient has $\ell_2$ error bounded as
\begin{equation*}
\begin{aligned}
    &|\mathbb{E}[\hat{\rho}_i]-\rho^{\pi}_i|\leq\frac{2C_p}{(1-\varrho)N_1}\varrho^{N_1/2},\\
    &\mathbb{E}(\hat{\rho}_i-\rho^{\pi}_i)^2\leq (2+4C_p\frac{\varrho}{1-\varrho})\frac{1}{N_1},\\
    &\lVert \mathbb{E}\hat{g}_i-\frac{\partial \rho_i^{\pi}}{\partial \pi_i}\rVert_2^2\leq C_p^2A_{\max}\left(\frac{N_2^2}{1-\varrho^{2N_2}}\varrho^{2N_1}+\frac{1}{(1-\varrho)^2}\frac{8N_2^2}{N_1^2}\varrho^{N_1}+\frac{2}{(1-\varrho)^2}\varrho^{2N_2}\right),\\
    &\mathbb{E}\lVert \hat{g}_i-\frac{\partial \rho_i^{\pi}}{\partial \pi_i}\rVert_2^2\leq (\frac{1}{\alpha} + 1)\frac{2A_{\max}N_2^2}{K}+\frac{4C_pA_{\max}}{1-\varrho^{N_2}}(\sqrt{\frac{2}{\alpha}}+\sqrt{2})\frac{N_2^2}{K}\varrho^{N_2}+\frac{16A_{\max}C_p^2}{(1-\varrho)^2}\frac{N_2^2}{N_1^2}\varrho^{N_1}+\frac{2A_{\max}C_p^2}{(1-\varrho)^2}\varrho^{2N_2}.
\end{aligned}
\end{equation*}
\end{lemma}
\begin{proof}
Note that $ \mathbb{E}[\hat{\rho}_i]=\mathbb{E}[\mathbb{E}[\hat{\rho}_i|s_0]]$, $\mathbb{E}[\hat{\rho}_i]-\rho^{\pi}_i=\mathbb{E}[\mathbb{E}[\hat{\rho}_i|s_0]-\rho^{\pi}_i]$,
    \begin{equation*}
        \begin{aligned}
            |\mathbb{E}[\hat{\rho}_i|s_0]-\rho^{\pi}_i|&=|\sum_{s}(\frac{2}{N_1}\sum_{t=\frac{N_1}{2}}^{N_1-1}(P^{\pi})^t(s|s_0)-\nu^{\pi}(s))\sum_a\pi(a|s)r_i(s,a)| \\ 
            &=|\frac{2}{N_1}\sum_{t=\frac{N_1}{2}}^{N_1-1}(\sum_{s}((P^{\pi})^t(s|s_0)-\nu^{\pi}(s))\sum_a\pi(a|s)r_i(s,a)| \\ 
            &\leq \frac{2}{N_1}\sum_{t=\frac{N_1}{2}}^{N_1-1}\sum_{s}|(P^{\pi})^t(s|s_0)-\nu^{\pi}(s)|\\ 
            &\leq \frac{2}{N_1}\sum_{t=\frac{N_1}{2}}^{N_1-1}C_p\varrho^t\\ 
            &\leq\frac{2}{N_1}C_p\varrho^{N_1/2}\frac{1}{1-\varrho}.
        \end{aligned}
    \end{equation*}
    Therefore, $|\mathbb{E}[\hat{\rho}_i]-\rho^{\pi}_i|\leq\frac{2}{N_1}C_p\varrho^{N_1/2}\frac{1}{1-\varrho}$. Use $\mathcal{F}_k$ to denote all the $(s,a,r)$ pairs until episode $k$. 
    \begin{equation*}
        \begin{aligned}
            \mathbb{E}(\hat{\rho}_i-\rho^{\pi}_i)^2&=\mathbb{E}(\frac{2}{N_1}\sum_{t=\frac{N_1}{2}}^{N_1-1}r_i^t-\rho^{\pi}_i)^2\\
            &=\frac{4}{N_1^2}(\sum_{t=\frac{N_1}{2}}^{N_1-1}\mathbb{E}(r_i^t-\rho^{\pi}_i)^2+2\sum_{t<\tau}\mathbb{E}[(r_i^t-\rho^{\pi}_i)(r_i^{\tau}-\rho^{\pi}_i)])\\
            &\leq \frac{2}{N_1}+\frac{8}{N_1^2}\sum_{t<\tau}\mathbb{E}[(r_i^t-\rho_i^{\pi})\mathbb{E}[r_i^{\tau}-\rho_i^{\pi}|F_t]]\\
            &\leq \frac{2}{N_1}+\frac{8}{N_1^2}\sum_{t<\tau}\mathbb{E}[|r_i^t-\rho_i^{\pi}||\mathbb{E}[r_i^{\tau}-\rho_i^{\pi}|F_t]|]\\
            &\leq \frac{2}{N_1}+\frac{8}{N_1^2}\sum_{t<\tau}\mathbb{E}|\mathbb{E}[r_i^{\tau}-\rho_i^{\pi}|F_t]|\\
            &\leq \frac{2}{N_1}+\frac{8}{N_1^2}\sum_{t<\tau}C_p\varrho^{\tau-t}
            =\frac{2}{N_1}+\frac{8}{N_1^2}\sum_{\tau=\frac{N_1}{2}}^{N_1-1}\sum_{t=\frac{N_1}{2}}^{\tau-1}C_p\varrho^{\tau-t}\\
            &\leq \frac{2}{N_1}+C_p\frac{8}{N_1^2}\frac{\varrho}{1-\varrho}\frac{N_1}{2}=(\frac{1}{2}+C_p\frac{\varrho}{1-\varrho})\frac{4}{N_1}.
        \end{aligned}
    \end{equation*}
    Note that $t_k=N_1+kN_2$ is the starting time step for the $k$-th episode, $R(k)=\sum_{\tau=t_k}^{t_k+N_2-1}(r_i^{\tau}-\hat\rho_i)$ is the accumulated bias for the $N_2$-length interval. Then $\hat g_i=\frac{1}{K}\sum_{k=0}^{K-1}R(k)\nabla_{\pi_i}\log\pi_i(a_i^{t_k}|s^{t_k})\in\mathbb{R}^{S\times A_i}$, where $ \nabla_{\pi_i}\log\pi_i(a_i|s)=\frac{1}{\pi_i(a_i|s)}\overrightarrow{\textbf{e}}_{(s,a_i)}\in\mathbb{R}^{S\times A_i}$ is a unit vector with the only non-zero element corresponding to $(s,a_i)$.
    \begin{align*}
        &\lVert \mathbb{E}\hat{g}_i-\frac{\partial \rho_i^{\pi}}{\partial \pi_i}\rVert_2^2\\
        =&\lVert \frac{1}{K}\sum_{k=0}^{K-1}\mathbb{E}[R(k)\nabla_{\pi_i}\log\pi_i(a_i^{t_k}|s^{t_k})]-\mathbb{E}_{s\sim\nu^{\pi},a_i\sim\pi_i(\cdot|s)}[\overline Q_{i}^{\pi}(s,a_i)\nabla_{\pi_i}\log\pi_i(a_i|s)]\rVert_2^2\\
        \leq&\lVert \frac{1}{K}\sum_{k=0}^{K-1}(\mathbb{E}[R(k)\nabla_{\pi_i}\log\pi_i(a_i^{t_k}|s^{t_k})]-\mathbb{E}_{s^{t_k}\sim\nu^{\pi},a_i^{t_k}\sim\pi(\cdot|s)}[R(k)\nabla_{\pi_i}\log\pi_i(a_i^{t_k}|s^{t_k})])\rVert_2^2\\
        &+\lVert\frac{1}{K}\sum_{k=0}^{K-1}(\mathbb{E}_{s^{t_k}\sim\nu^{\pi},a_i^{t_k}\sim\pi(\cdot|s)}[R(k)\nabla_{\pi_i}\log\pi_i(a_i^{t_k}|s^{t_k})]-\mathbb{E}_{s\sim\nu^{\pi},a_i\sim\pi(\cdot|s)}[\overline Q_{i}^{\pi}(s,a_i)\nabla_{\pi_i}\log\pi_i(a_i|s)])\rVert_2^2\\
        \leq&\lVert \frac{1}{K}\sum_{k=0}^{K-1}\sum_{\tau=t_k}^{t_k+N_2-1} \sum_{s^{t_k},a_i^{t_k},s^{\tau},a_i^{\tau}}\nabla_{\pi_i}\log\pi_i(a_i^{t_k}|s^{t_k})(\mathbb{P}(s^{t_k},a_i^{t_k})-\nu^{\pi}(s^{t_k},a_i^{t_k}))\mathbb{P}(s^{\tau},a_i^{\tau}|s^{t_k},a_i^{t_k})(r(s^{\tau},a_i^{\tau})-\hat{\rho}_i)\rVert_2^2\\
        &+\frac{1}{K}\sum_{k=0}^{K-1}\lVert\mathbb{E}_{s^{t_k}\sim\nu^{\pi},a_i^{t_k}\sim\pi(\cdot|s)}[(R(k)-\overline Q_{i}^{\pi}(s^{t_k},a_i^{t_k}))\nabla_{\pi_i}\log\pi_i(a^{t_k}_i|s^{t_k})]\rVert_2^2\\
        \leq&\lVert \frac{1}{K}\sum_{k=0}^{K-1}\sum_{\tau=t_k}^{t_k+N_2-1} \sum_{s^{t_k},a_i^{t_k}}\overrightarrow{\textbf{e}}_{s^{t_k},a_i^{t_k}}|(P^{\pi})^{t_k}(s^{t_k}|s_0)-\nu^{\pi}(s^{t_k})|\rVert_2^2\\
        &+\frac{1}{K}\sum_{k=0}^{K-1}\lVert\sum_{a_i^{t_k}}\sum_{s^{t_k}}\nu^{\pi}(s^{t_k})\overrightarrow{\textbf{e}}_{s^{t_k},a_i^{t_k}}\mathbb{E}[(R(k)-\overline Q_{i}^{\pi}(s^{t_k},a_i^{t_k}))|s^{t_k},a_i^{t_k}]\rVert_2^2\\
        \leq&\frac{1}{K}\sum_{k=0}^{K-1}N_2^2\sum_{a_i^{t_k}}\sum_{s^{t_k}}|(P^{\pi})^{t_k}(s^{t_k}|s_0)-\nu(s^{t_k})|^2+\frac{1}{K}\sum_{k=0}^{K-1}\sum_{a_i^{t_k}}\sum_{s^{t_k}}\nu^{\pi}(s^{t_k})^2|\mathbb{E}[(R(k)-\overline Q_{i}^{\pi}(s^{t_k},a_i^{t_k}))|s^{t_k},a_i^{t_k}]|^2.\\
    \end{align*}
    Starting from any $s^0=s$ and $a_i^0=a_i$, the difference between $R=\sum_{t=0}^{N_2-1}(r_i^t-\hat{\rho})$ and $\overline Q^{\pi}_i(s,a_i)$ can be bounded as below:
    \begin{align*}
        &|\mathbb{E}[\sum_{t=0}^{N_2-1}(r_i(s^t,a^t)-\hat{\rho})-\sum_{t=0}^{\infty}(r_i(s^t,a^t)-\rho_i^{\pi})]|\\
        =&|\mathbb{E}[\sum_{t=0}^{N_2-1}(\rho_i^{\pi}-\hat{\rho}_i)]-\mathbb{E}[\sum_{t=N_2}^{\infty}(r_i(s^t,a^t)-\rho_i^{\pi})]|\\
        \leq& N_2\frac{2C_p}{(1-\varrho)N_1}\varrho^{N_1/2}+\sum_{t=N_2}^{\infty}C_p\varrho^t\\
        =&N_2\frac{2C_p}{(1-\varrho)N_1}\varrho^{N_1/2}+\frac{C_p}{1-\varrho}\varrho^{N_2}.
    \end{align*}
    Therefore, we can bound $\lVert \mathbb{E}\hat{g}_i-\frac{\partial \rho_i^{\pi}}{\partial \pi_i}\rVert_2^2$ as:
    \begin{align*}
        \lVert \mathbb{E}\hat{g}_i-\frac{\partial \rho_i^{\pi}}{\partial \pi_i}\rVert_2^2\leq&\frac{1}{K}\sum_{k=0}^{K-1}N_2^2A_{\max}(C_p\varrho^{t_k})^2+A_{\max}(N_2\frac{2}{N_1}C_p\varrho^{N_1/2}\frac{1}{1-\varrho}+C_p\varrho^{N_2}\frac{1}{1-\varrho})^2\\
        =&\frac{1}{K}\sum_{k=0}^{K-1}N_2^2A_{\max}(C_p\varrho^{N_1+kN_2})^2+A_{\max}(N_2\frac{2}{N_1}C_p\varrho^{N_1/2}\frac{1}{1-\varrho}+C_p\varrho^{N_2}\frac{1}{1-\varrho})^2\\
        \leq&C_p^2A_{\max}\left(\frac{N_2^2}{1-\varrho^{2N_2}}\varrho^{2N_1}+\frac{1}{(1-\varrho)^2}\frac{8N_2^2}{N_1^2}\varrho^{N_1}+\frac{2}{(1-\varrho)^2}\varrho^{2N_2}\right).\\
    \end{align*}
    Denote $\overline{R}(k):=\mathbb{E}_{s^{t_k}\sim\nu^{\pi},a^{t_k}\sim\pi(\cdot|s)}[R(k)\nabla_{\pi_i}\log\pi_i(a_i^{t_k}|s^{t_k})]$. As above, we can bound the variance as:
    \begin{align*}
        \mathbb{E}\lVert \hat{g}_i-\frac{\partial \rho_i^{\pi}}{\partial \pi_i}\rVert_2^2
        \leq& 2\mathbb{E}\lVert\frac{1}{K}\sum_{k=0}^{K-1}R(k)\nabla_{\pi_i}\log\pi_i(a_i^{t_k}|s^{t_k})-\mathbb{E}_{s^{t_k}\sim\nu^{\pi},a^{t_k}\sim\pi(\cdot|s)}[R(k)\nabla_{\pi_i}\log\pi_i(a_i^{t_k}|s^{t_k})]\rVert_2^2\\
        &+2\lVert\mathbb{E}_{s^{t_k}\sim\nu^{\pi},a^{t_k}\sim\pi(\cdot|s)}[R(k)\nabla_{\pi_i}\log\pi_i(a_i^{t_k}|s^{t_k})]-\mathbb{E}_{s\sim\nu^{\pi},a\sim\pi(\cdot|s)}[Q_{i}^{\pi}(s,a)\nabla_{\pi_i}\log\pi_i(a_i|s)]\rVert_2^2\\
        \leq&\frac{2}{K^2}\sum_{k=0}^{K-1}\mathbb{E}\lVert R(k)\nabla_{\pi_i}\log\pi_i(a_i^{t_k}|s^{t_k})-\overline{R}(k)\rVert_2^2\\
        &+\frac{4}{K^2}\sum_{k<\tau}\mathbb{E}\langle R(k)\nabla_{\pi_i}\log\pi_i(a_i^{t_k}|s^{t_k})-\overline{R}(k), R(\tau)\nabla_{\pi_i}\log\pi_i(a_i^{t_{\tau}}|s^{t_{\tau}})-\overline{R}(\tau)\rangle\\
        &+2A_{\max}(N_2\frac{2}{N_1}C_p\varrho^{N_1/2}\frac{1}{1-\varrho}+C_p\varrho^{N_2}\frac{1}{1-\varrho})^2.\\
    \end{align*}
    We bound the first two terms separately.
    \begin{align*}
        &\mathbb{E}\lVert R(k)\nabla_{\pi_i}\log\pi_i(a_i^{t_k}|s^{t_k})-\overline{R}(k)\rVert_2^2\\
        \leq&2\mathbb{E}\lVert R(k)\nabla_{\pi_i}\log\pi_i(a_i^{t_k}|s^{t_k})\rVert_2^2+2\lVert \overline{R}(k)\rVert_2^2\\
        =&2\sum_{s^{t_k},a_i^{t_k}}\mathbb{P}(s^{t_k})\pi_i(a_i^{t_k}|s^{t_k})\mathbb{E}[\lVert \frac{1}{\pi_i(a^{t_k}|s^{t_k})}R(k)\overrightarrow{\textbf{e}}_{s^{t_k},a_i^{t_k}}\rVert_2^2|s^{t_k},a^{t_k}_i]\\
        &+ 2\lVert\sum_{s^{t_k}}\nu^{\pi}(s^{t_k})\sum_{a_i^{t_k}}\mathbb{E}[R(k)\overrightarrow{\textbf{e}}_{s^{t_k},a^{t_k}_i}|s^{t_k},a_i^{t_k}]\rVert_2^2\\
        =&2\sum_{s^{t_k},a_i^{t_k}}\mathbb{P}(s^{t_k})\frac{1}{\pi_i(a_i^{t_k}|s^{t_k})}\mathbb{E}[R(k)^2|s^{t_k},a^{t_k}_i] +  2\sum_{s^{t_k}}\nu^{\pi}(s^{t_k})^2\sum_{a_i^{t_k}}\mathbb{E}[R(k)|s^{t_k},a_i^{t_k}]^2\\
        \leq&\frac{2A_{\max}}{\alpha}N_2^2 + 2A_{\max}N_2^2,\ \forall k,
    \end{align*}
    and 
    \begin{align*}
        &\mathbb{E}[\langle R(k)\nabla_{\pi_i}\log\pi_i(a_i^{t_k}|s^{t_k})-\overline{R}(k), R(\tau)\nabla_{\pi_i}\log\pi_i(a_i^{t_{\tau}}|s^{t_{\tau}})-\overline{R}(\tau)\rangle]\\
        =&\mathbb{E}\langle R(k)\nabla_{\pi_i}\log\pi_i(a_i^{t_k}|s^{t_k})-\overline{R}(k), \mathbb{E}[R(\tau)\nabla_{\pi_i}\log\pi_i(a_i^{t_{\tau}}|s^{t_{\tau}})-\overline{R}(\tau)|F_k]\rangle\\
        \leq&\mathbb{E}\Big[\lVert R(k)\nabla_{\pi_i}\log\pi_i(a_i^{t_k}|s^{t_k})-\overline{R}(k)\rVert_2\lVert \mathbb{E}[R(\tau)\nabla_{\pi_i}\log\pi_i(a_i^{t_{\tau}}|s^{t_{\tau}})-\overline{R}(\tau)\rangle|F_k]\rVert_2\Big]\\
        \leq&\mathbb{E}\Big[\lVert R(k)\nabla_{\pi_i}\log\pi_i(a_i^{t_k}|s^{t_k})-\overline{R}(k)\rVert_2\lVert \sum_{s^{t_{\tau}}}((P^{\pi})^{t_{\tau}-t_k}(s^{t_{\tau}}|s^{t_k})-\nu^{\pi}(s^{t_{\tau}}))\sum_{a_i^{t_{\tau}}}\mathbb{E}[R(\tau)\overrightarrow{\textbf{e}}_{s^{t_{\tau}},a_i^{t_{\tau}}}|s^{t_{\tau}},a_i^{t_{\tau}}]\rVert_2 \Big]\\
        \overset{(a)}{\leq}&\mathbb{E}\Big[\lVert R(k)\nabla_{\pi_i}\log\pi_i(a_i^{t_k}|s^{t_k})-\overline{R}(k)\rVert_2 \sum_s|(P^{\pi})^{t_{\tau}-t_k}(s|s^{t_k})-\nu^{\pi}(s)| \sqrt{N_2^2A_{\max}} \Big] \\
        \leq&\mathbb{E}\lVert R(k)\nabla_{\pi_i}\log\pi_i(a_i^{t_k}|s^{t_k})-\overline{R}(k)\rVert_2C_p\varrho^{(\tau-t)N_2}N_2\sqrt{A_{\max}}\\
        \overset{(b)}{\leq}&\sqrt{\mathbb{E}\lVert R(k)\nabla_{\pi_i}\log\pi_i(a_i^{t_k}|s^{t_k})-\overline{R}(t)\rVert_2^2}C_p\varrho^{(\tau-t)N_2}N_2\sqrt{A_{\max}} \\ 
        \overset{(c)}{\leq}&\left(\sqrt{\frac{2A_{\max}}{\alpha}N_2^2}+\sqrt{2A_{\max}N_2^2}\right)C_p\varrho^{(\tau-t)N_2}N_2\sqrt{A_{\max}}\\
        = &\left(\sqrt{\frac{2}{\alpha}}+\sqrt{2}\right)C_p\varrho^{(\tau-t)N_2}A_{\max}N_2^2.\\
    \end{align*}
    In (a) we used the fact $\lVert x\rVert_2\leq \lVert x\rVert_1$ in (a). (b) is due to Jensen's inequality, (c) is due to $\sqrt{a^2+b^2}\leq a+b$ for any positive $a,b$.
    Therefore, we can bound the variance as
    \begin{align*}
        \mathbb{E}\lVert \hat{g}_i-\frac{\partial \rho_i^{\pi}}{\partial \pi_i}\rVert_2^2&\leq2(\frac{1}{\alpha} + 1)A_{\max}N_2^2\frac{1}{K}+\frac{4C_p}{1-\varrho^{N_2}}(\sqrt{\frac{2}{\alpha}}+\sqrt{2})A_{\max}N_2^2\varrho^{N_2}\frac{1}{K}\\
        &+2A_{\max}(N_2\frac{2}{N_1}C_p\varrho^{N_1/2}\frac{1}{1-\varrho}+C_p\varrho^{N_2}\frac{1}{1-\varrho})^2.
    \end{align*}
\end{proof}

\begin{theorem}[Restatement of \cref{thm:2}]
    If all agents independently and synchronously run \cref{alg:2} with learning rate $\beta\leq\frac{1}{L_{\Phi}}$, then Nash regret is bounded by:
    \begin{equation*}
        \text{Nash-regret}^*(T)\leq 48D^2S(L+\frac{1}{\beta})^2\beta\frac{N}{T}+12\kappa^2\alpha^2+\left(\frac{48D^2S(L+\frac{1}{\beta})^2\beta}{L_{\Phi}}+6\kappa^2A_{\max}\beta^2+24D^2S(2L\beta+1)^2\right)\delta.
    \end{equation*}
\end{theorem}
\begin{proof}
    Let $\tilde\pi_i^{t+1}=P_{\Pi_{i,\alpha}}(\pi_i^t+\beta\nabla_{\pi_i}\rho^{\pi^t}_i)$ be the projection after a exact gradient step. Since $\pi_i^{t+1}=P_{\Pi_{i,\alpha}}(\pi_i^t+\beta\hat{g}_i^t)$, we have $\lVert \tilde\pi_i^{t+1}-\pi_i^{t+1}\rVert_2\leq\beta\lVert\nabla_{\pi_i}\rho_i^{\pi^t}-\hat{g}_i^t\rVert_2$ by the non-expansion of projection. Then
    \begin{align*}
        &\text{Nash-gap}(t)\\
        \overset{(a)}{=}&\rho^{\pi_i^{t,*},\pi_{-i}^t}_i-\rho_i^{\pi^t}\\
        =&E_{s\sim\nu^{\pi_i^{t,*},\pi_{-i}^t}}[\langle\overline{Q}_i^{\pi_t}(s,\cdot),\pi_i^{t,*}(\cdot|s)-\pi_i^t(\cdot|s)\rangle] \\
        =&E_{s\sim\nu^{\pi_i^{t,*},\pi_{-i}^t}}[\langle\overline{Q}_i^{\pi_t}(s,\cdot),\pi_i^{t,*}(\cdot|s)-\tilde\pi_i^t(\cdot|s)\rangle] + E_{s\sim\nu^{\pi_i^{t,*},\pi_{-i}^t}}[\langle\overline{Q}_i^{\pi_t}(s,\cdot),\tilde\pi_i^{t}(\cdot|s)-\pi_i^t(\cdot|s)\rangle]\\
        \leq& E_{s\sim\nu^{\pi_i^{t,*},\pi_{-i}^t}}[\langle\overline{Q}_i^{\pi_t}(s,\cdot),\pi_i^{t,*}(\cdot|s)-\tilde\pi_i^t(\cdot|s)\rangle] + \kappa \max_s\lVert\tilde\pi_i^{t}(\cdot|s)-\pi_i^t(\cdot|s)\rVert_1\\
        \leq& E_{s\sim\nu^{\pi_i^{t,*},\pi_{-i}^t}} [\langle\overline{Q}_i^{\pi_t}(s,\cdot),(1-\alpha)\pi^{t,*}_i(\cdot|s)+\alpha u_i(\cdot|s)-\tilde\pi_i^t(\cdot|s)\rangle]+ \alpha E_{s\sim\nu^{\pi_i^{t,*},\pi_{-i}^t}}[\langle\overline{Q}_i^{\pi_t}(s,\cdot),\pi^{t,*}_i(\cdot|s)- u_i(\cdot|s)\rangle]\\
        &+\kappa\sqrt{A_{\max}}\lVert\tilde\pi_i^{t}-\pi_i^t\rVert_2 \\
        \overset{(b)}{\leq}& E_{s\sim\nu^{\pi_i^{t,*},\pi_{-i}^t}}[\max_{\pi'_i\in\Pi_i}\langle\overline{Q}_i^{\pi_t}(s,\cdot),(1-\alpha)\pi'_i(\cdot|s)+\alpha u_i(\cdot|s)-\tilde\pi_i^t(\cdot|s)\rangle ]+ 2\alpha\lVert\overline{Q}_i^{\pi_t}\rVert_{\infty}+\kappa\sqrt{A_{\max}}\lVert\tilde\pi_i^{t}-\pi_i^t\rVert_2\\
        \overset{(c)}{\leq}& \max_s\frac{\nu^{\pi_i^{t,*},\pi_t^{-i}}(s)}{\nu^{\pi_t}(s)}\langle\nabla_{\pi_i}\Phi(\pi^t),(1-\alpha)\pi'_i+\alpha u_i-\tilde\pi_i^t\rangle+2\alpha\kappa+\kappa\sqrt{A_{\max}}\lVert\tilde\pi_i^{t}-\pi_i^t\rVert_2 \\
        \leq& 2D\sqrt{S}\underset{\tilde\pi_i^t+\delta\in\Pi_{i,\alpha},\lVert\delta\rVert_2\leq1}{\max}\delta^T\nabla_{\pi_i}\Phi(\pi^t) +2\alpha\kappa+\kappa\sqrt{A_{\max}}\lVert\tilde\pi_i^{t}-\pi_i^t\rVert_2\\
        \leq& 2D\sqrt{S}\underset{\tilde\pi_i^t+\delta\in\Pi_{\alpha},\lVert\delta\rVert_2\leq1}{\max}\delta^T\nabla_{\pi_i}\Phi(\tilde\pi_i^{t},\pi_{-i}^t)+2D\sqrt{S}\underset{\tilde\pi_i^t+\delta\in\Pi_{i,\alpha},\lVert\delta\rVert_2\leq1}{\max}\delta^T(\nabla_{\pi_i}\Phi(\pi^{t})-\nabla_{\pi_i}\Phi(\tilde\pi_i^{t},\pi_{-i}^t))\\
        &+2\alpha\kappa+\kappa\sqrt{A_{\max}}\lVert\tilde\pi_i^{t}-\pi_i^t\rVert_2\\
        \overset{(d)}{\leq}& 2D\sqrt{S}\lVert \tilde\pi_i^{t}-\pi_i^{t-1}\rVert_2(L+\frac{1}{\beta})+2D\sqrt{S}L\lVert\pi_i^t-\tilde\pi_i^{t}\rVert_2+2\alpha\kappa+\kappa\sqrt{A_{\max}}\lVert\tilde\pi_i^{t}-\pi_i^t\rVert_2 \\
        =& 2D\sqrt{S}(L+\frac{1}{\beta})\lVert \tilde\pi_i^{t}-\pi_i^{t-1}\rVert_2 +2\alpha\kappa+(\kappa\sqrt{A_{\max}}+2D\sqrt{S}L)\lVert\tilde\pi_i^{t}-\pi_i^t\rVert_2\\ 
        \leq& 2D\sqrt{S}(L+\frac{1}{\beta})(\lVert \tilde\pi_i^{t}-\pi_i^t\rVert_2+\lVert\pi_i^t-\pi_i^{t-1}\rVert_2 ) +2\alpha\kappa+(\kappa\sqrt{A_{\max}}+2D\sqrt{S}L)\lVert\tilde\pi_i^{t}-\pi_i^t\rVert_2\\
        \leq& 2D\sqrt{S}(L+\frac{1}{\beta})\lVert\pi_i^t-\pi_i^{t-1}\rVert_2 +2\alpha\kappa+(\kappa\sqrt{A_{\max}}+2D\sqrt{S}(2L+\frac{1}{\beta}))\beta\lVert\nabla_{\pi_i}\rho_i^{\pi^{t-1}}-\hat{g}_i^{t-1}\rVert_2.
    \end{align*}
    In (a) we have used $i$ to denote the agent that achieves the maximum of Nash gap, i.e., $i\in \arg\max_i\ \max_{p\in\Pi_i}(\rho^{p,\pi_{-i}^t}-\rho_i^{\pi_i^t,\pi_{-i}^t})$, and $\pi_i^{t,*}\in \arg\max_{p\in\Pi_i}(\rho^{p,\pi_{-i}^t}-\rho_i^{\pi_i^t,\pi_{-i}^t})$. We use $\langle x,y\rangle\leq\lVert x\rVert_{\infty}\lVert y\rVert_1$ and $\lVert \pi_i(\cdot|s)\rVert_1=1$ in (b). (c) is true since $\max_{\pi'(\cdot|s)\in\Delta(A_i)}\langle\overline{Q}_i^{\pi_t}(s,\cdot),\pi'(\cdot|s)-\pi_t^i(\cdot|s)\rangle\geq0$. We also use $\pi'_i$ to represent the policy that achieves the maximum. (d) comes from \cref{lm:15} (note that $\Pi_{i,\alpha}$ is convex) and that $\Phi(\pi)$ is $L$-smooth w.r.t. $\pi_i$ for any $\pi_{-i}\in\Pi_{-i}$.

    Applying \cref{lm:17} with $f=-\Phi$, $E[\Phi(\pi^{t+1})-\Phi(\pi^t)]\geq (\frac{1}{\beta}-\frac{3L_{\Phi}}{4})E\lVert \pi^t-\pi^{t+1}\rVert_2^2-\frac{1}{NL}\sum_i\delta$. Choosing a learning rate $\beta\leq \frac{1}{L_{\Phi}}$, $4\beta E[\Phi(\pi^{t+1})-\Phi(\pi^t)]+\frac{4\beta}{L_{\Phi}}\sum_i\delta\geq E\lVert \pi^t-\pi^{t+1}\rVert_2^2$, we arrive at 
    
    \begin{align*}
        \mathbb{E}[\text{Nash-gap}(t)^2]\leq & 12D^2S(L+\frac{1}{\beta})^2\mathbb{E}\lVert \pi^{t}-\pi^{t-1}\rVert_2^2+12\alpha^2\kappa^2+3(\kappa\sqrt{A_{\max}}+2D\sqrt{S}(2L+\frac{1}{\beta}))^2\beta^2\delta,\\
        \mathbb{E}[\frac{1}{T}\sum_{t=1}^{T}\text{Nash-gap}(t)^2]\leq & 12D^2S(L+\frac{1}{\beta})^2\left(\mathbb{E}[\frac{1}{T}\sum_{t=1}^{T}4\beta(\Phi(\pi^{t+1})-\Phi(\pi^t)]+\frac{4\beta}{L_{\Phi}T}\sum_{i,t}\delta\right)+12\alpha^2\kappa^2\\
        &+(6\kappa^2A_{\max}\beta^2+24D^2S(2L\beta+1)^2)\delta\\
        \leq & 48D^2S(L+\frac{1}{\beta})^2\beta\mathbb{E}[\Phi(\pi^{T})-\Phi(\pi^0)]\frac{1}{T}+12\alpha^2\kappa^2\\
        &+\left(\frac{48D^2S(L+\frac{1}{\beta})^2\beta C_{\Phi}}{L_{\Phi}}+6\kappa^2A_{\max}\beta^2+24D^2S(2L\beta+1)^2\right)\delta\\
        \leq& 48D^2S(L+\frac{1}{\beta})^2\beta\frac{C_{\Phi}}{T}+12\alpha^2\kappa^2\\
        &+\left(\frac{48D^2S(L+\frac{1}{\beta})^2\beta C_{\Phi}}{L_{\Phi}}+6\kappa^2A_{\max}\beta^2+24D^2S(2L\beta+1)^2\right)\delta.\\
    \end{align*}    
\end{proof}

\subsection{Notes of Previous Work}\label{B.4}
The proof of Theorem 4.7 in \cite{leonardos2021global}, where they established a sample complexity result, appears to have two mistakes. In their latest version, the first equation on p. 16 was derived based on:
\begin{equation}\label{eq:err}
    -\Phi_{\mu}(y^{t+1})+\frac{1}{\lambda}\lVert\pi^t-y^{t+1}\rVert_2^2\leq \phi_{\lambda}(\pi^t),
\end{equation} 
where $y^{t+1}=\text{Proj}_{\Pi}(\pi^t+\eta\nabla_{\pi}\Phi_{\mu}(\pi^t))$, $\phi_{\lambda}(\pi^t)=\min_{y\in\Pi}(-\Phi_{\mu}(y)+\frac{1}{\lambda}\lVert y-\pi^t\rVert_2^2)$. Note that \cref{eq:err} is equivalent to $y^{t+1}$ reaching the minimizer of $\phi_{\lambda}(\pi^t)$. However, it should be noted that the minimizer of Moreau envelope $\arg\min_{y\in\Pi}(-\Phi_{\mu}(y)+\frac{1}{\lambda}\lVert y-\pi^t\rVert_2^2)$ may not be the one step updated policy of the projected gradient ascent algorithm. A counterexample can be found in \citealp[chapter 6.2.3]{beck2017first}. In \citealp[Theorem 2.1]{davis2018stochastic}, which the proof in \citep{leonardos2021global} mostly follow,  $y^{t+1}$ was defined as the minimizer of the Moreau envelope $\arg\min_{y\in\Pi}(-\Phi_{\mu}(y)+\frac{1}{\lambda}\lVert y-\pi^t\rVert_2^2)$, and it was only established that $\lVert\nabla_{\pi}\phi_{\lambda}(\pi^t) \rVert_2$ is bounded, not that $\lVert\nabla_{\pi}\Phi_{\mu}(\pi^t) \rVert_2$ is bounded. However, the latter appears critical in establishing the bounded Nash gap in \citealp[Theorem 4.7]{leonardos2021global}.


Furthermore, Eq. (14) in \cite{leonardos2021global} shows that there exists a time $t^*$ s.t. $\lVert y^{t^*+1}-\pi^{t^*}\rVert_2$ can be bounded. Subsequently, the authors used their Lemma D.3 and Lemma 4.2 to show a bounded Nash gap. However, Lemma D.3 only guarantees that $y^{t^*+1}$, instead of $\pi^{t^*}$, is a $\epsilon$-stationary point, accordingly an $\epsilon$-Nash equilibrium. It should further be noted that $y$ is not tractable with a finite number of samples.

\section{PROOFS OF SECTION 4}\label{C}
\begin{lemma}[Restatement of \cref{lm:7}]
For any reward function $r\in[0,1]$, any policies $\pi,\pi'\in\Pi$, the following bounds hold, where we replace $V_i$ with $V_r$ to indicate a general reward function:
    \begin{align*}
        &|\nu^{\pi}(s)-\nu^{\pi'}(s)|\leq\kappa\lVert\pi-\pi'\rVert_{1,\infty}, \forall s\in\mathcal{S}\\
        &|\rho^{\pi}_r-\rho^{\pi'}_r|\leq\kappa\lVert\pi-\pi'\rVert_{1,\infty},\\
        &\lVert V_r^{\pi}-V_r^{\pi'}\rVert_{\infty} \leq \kappa_1(2+S(\kappa+\kappa_1)+S\kappa\kappa_1)\lVert\pi-\pi'\rVert_{1,\infty},\\
        &\lVert Q^{\pi}_r-Q_r^{\pi'}\rVert_{\infty} \leq(\kappa+2\kappa_1+S\kappa_1(\kappa+\kappa_1)+S\kappa\kappa_1^2)\times\lVert\pi-\pi'\rVert_{1,\infty}.
    \end{align*}
\end{lemma}
\begin{proof}
First, we provide the sensitivity analysis of the policy-induced reward and the state transition probability matrix:
    \begin{align*}
        |r^{\pi}(s)-r^{\pi'}(s)|&=|\sum_a(\pi(a|s)-\pi'(a|s))r(s,a)|\leq\lVert\pi(\cdot|s)-\pi'(\cdot|s)\rVert_1,\ \forall s\in S,\\
        |P^{\pi}(s'|s)-P^{\pi'}(s'|s)|&=|\sum_a(\pi(a|s)-\pi'(a|s))P(s'|s,a)|\leq\lVert\pi(\cdot|s)-\pi'(\cdot|s)\rVert_1,\ \forall s,s'\in S.
    \end{align*}
    From \cref{def:5} and \cref{lm:2}, taking $\mathbb{I}(\cdot=s)$ as the reward function, it follows that $\lVert\nu^{\pi}-\nu^{\pi'}\rVert_{\infty}\leq\kappa\lVert\pi-\pi'\rVert_{1,\infty}$. Similarly by the performance difference lemma, $|\rho^{\pi}-\rho^{\pi'}|\leq\kappa\lVert\pi-\pi'\rVert_{1,\infty}$. 
    \begin{align*}
        H^{\pi}-H^{\pi'}=&(I-P^{\pi}+P^{\pi,\infty})^{-1}(I-P^{\pi,\infty})-(I-P^{\pi'}+P^{\pi',\infty})^{-1}(I-P^{\pi',\infty})\\
        = &\left((I-P^{\pi}+P^{\pi,\infty})^{-1}-(I-P^{\pi'}+P^{\pi',\infty})^{-1}\right)(I-P^{\pi,\infty})+(I-P^{\pi'}+P^{\pi',\infty})^{-1}(P^{\pi',\infty}-P^{\pi,\infty})\\
        =&\left((I-P^{\pi}+P^{\pi,\infty})^{-1}(P^{\pi}-P^{\pi,\infty}-P^{\pi'}+P^{\pi',\infty})(I-P^{\pi'}+P^{\pi',\infty})^{-1}\right)(I-P^{\pi,\infty})\\
        &+(I-P^{\pi'}+P^{\pi',\infty})^{-1}(P^{\pi',\infty}-P^{\pi,\infty}),\\
        \lVert V^{\pi}-V^{\pi'}\rVert_{\infty}=&\lVert H^{\pi}r^{\pi}-H^{\pi'}r^{\pi'}\rVert_{\infty}\\
        =&\lVert H^{\pi}(r^{\pi}-r^{\pi'})+(H^{\pi}-H^{\pi'})r^{\pi'}\rVert_{\infty}\\
        \leq & \lVert H^{\pi}(r^{\pi}-r^{\pi'})\rVert_{\infty}+\lVert(I-P^{\pi'}+P^{\pi',\infty})^{-1}(P^{\pi',\infty}-P^{\pi,\infty})r^{\pi'}\rVert_{\infty}\\
        &+\lVert\left((I-P^{\pi}+P^{\pi,\infty})^{-1}(P^{\pi}-P^{\pi,\infty}-P^{\pi'}+P^{\pi',\infty})(I-P^{\pi'}+P^{\pi',\infty})^{-1}\right)(I-P^{\pi,\infty})r^{\pi'}\rVert_{\infty}\\
        \leq & \lVert H^{\pi}\rVert_{\infty}\lVert r^{\pi}-r^{\pi'}\rVert_{\infty}+\lVert(I-P^{\pi'}+P^{\pi',\infty})^{-1}\rVert_{\infty}\lVert P^{\pi',\infty}-P^{\pi,\infty}\rVert_{\infty}\\
        &+\lVert(I-P^{\pi}+P^{\pi,\infty})^{-1}(P^{\pi}-P^{\pi,\infty}-P^{\pi'}+P^{\pi',\infty})(I-P^{\pi'}+P^{\pi',\infty})^{-1}\rVert_{\infty}\\
        \overset{(a)}{\leq}&2\kappa_1\lVert r^{\pi}-r^{\pi'}\rVert_{\infty}+ \kappa_1\lVert\nu^{\pi}-\nu^{\pi'}\rVert_{1}+\kappa_1^2\lVert P^{\pi}-P^{\pi'}\rVert_{\infty}+\kappa_1^2\lVert\nu^{\pi}-\nu^{\pi'}\rVert_{1}\\
        \leq& \kappa_1(2+S\kappa+\kappa_1S+\kappa_1S\kappa)\lVert\pi-\pi'\rVert_{1,\infty}.
    \end{align*}
    Above (a) is due to $\lVert P^{\pi',\infty}-P^{\pi,\infty}\rVert_{\infty}\leq\lVert\nu^{\pi}-\nu^{\pi'} \rVert_1$, $ \lVert P^{\pi}-P^{\pi'}\rVert_{\infty}\leq max_{s}\sum_{s'}|P^{\pi}(s'|s)-P^{\pi'}(s'|s) |$ and $\lVert(I-P^{\pi,\infty})r^{\pi'}\rVert_{\infty}\leq 1$. Now
    \begin{align*}
        &|Q^{\pi}(s,a)-Q^{\pi'}(s,a)|=|\rho^{\pi}-\rho^{\pi'}+\langle P(\cdot|s,a),V^{\pi}-V^{\pi'}\rangle|,\\
        & \lVert Q^{\pi}-Q^{\pi'}\rVert_{\infty}\leq |\rho^{\pi}-\rho^{\pi'}|+\lVert V^{\pi}-V^{\pi'}\rVert_{\infty}\leq (\kappa+2\kappa_1+S\kappa_1(\kappa+\kappa_1)+S\kappa\kappa_1^2)\lVert\pi-\pi'\rVert_{1,\infty}.
    \end{align*}
\end{proof}

\textbf{Remark} In the case of large state set $\mathcal{S}$ but with finite cardinality $S<\infty$, the general parameterized policy classes \cite[Assumption 1]{xu2020improving} is commonly utilized. With the \cite[Assumption 1(3)]{xu2020improving}, $\lVert Q^{\pi_w}-Q^{\pi_{w'}}\rVert_{\infty}\leq \kappa_Q\lVert\pi-\pi'\rVert_{1,\infty}\leq \kappa_QC_{\pi}\lVert w-w'\rVert_2 $, thus their derivatives in Appendix B still hold. The smoothness of average reward with respect to the general parameterized policy classes can be shown.

\begin{lemma}[policy improvement (a)]\label{lm:21} Let $\pi^t$ to be the policy at time $t$ of \cref{alg:3}, 
    \begin{equation*}
        \Phi(\pi^{t+1}) - \Phi(\pi^t) \geq \frac{1}{\beta}\left(1-\beta\frac{(N-1)(\kappa_Q+S\kappa^2)A_{\max}}{2(1-\Gamma)}\right)\sum_{i=1}^N \sum_s\nu^{\pi_i^{t+1},\pi_{-i}^t}(s)\lVert\pi_i^{t+1}(\cdot|s)-\pi_i^{t}(\cdot|s)\rVert_2^2.
    \end{equation*}
\end{lemma}
\begin{proof}
    As in \cite{ding2022independent}, we can derive a bound of $\Phi^{t+1}-\Phi^t$ with the decomposition:
    \begin{equation}\label{eq:12}
    \begin{aligned}
        \Phi^{t+1}-\Phi^t=&\underbrace{\sum_{i=1}^N (\Phi(\pi^{t+1}_i,\pi^t_{-i})-\Phi(\pi^{t}_i,\pi^t_{-i}))}_{\text{Diff}_1}\\
        &+ \underbrace{\sum_{i=1}^N\sum_{j=i+1}^N (\Phi(\pi_{-(i,j)}^{t,t+1},\pi_i^{t+1},\pi_j^{t+1})-\Phi(\pi_{-(i,j)}^{t,t+1},\pi_i^t,\pi_j^{t+1}) - \Phi(\pi_{-(i,j)}^{t,t+1},\pi_i^{t+1},\pi_j^t)+\Phi(\pi_{-(i,j)}^{t,t+1},\pi_i^t,\pi_j^t))}_{\text{Diff}_2},
    \end{aligned}
    \end{equation}
    (where $\pi_{-(i,j)}^{t,t+1}:=(\pi_{1\sim i-1,i+1\sim j-1}^t, \pi_{j+1\sim N}^{t+1}$)\ ).\\
    First we bound each term in $\text{Diff}_2$:
    \begin{align*}
        &\Phi(\pi_{-(i,j)}^{t,t+1},\pi_i^{t+1},\pi_j^{t+1})-\Phi(\pi_{-(i,j)}^{t,t+1},\pi_i^t,\pi_j^{t+1}) - \Phi(\pi_{-(i,j)}^{t,t+1},\pi_i^{t+1},\pi_j^t)+\Phi(\pi_{-(i,j)}^{t,t+1},\pi_i^t,\pi_j^t) \\
        =& \rho_i^{\pi_{-(i,j)}^{t,t+1},\pi_i^{t+1},\pi_j^{t+1}}-\rho_i^{\pi_{-(i,j)}^{t,t+1},\pi_i^t,\pi_j^{t+1}} - \rho_i^{\pi_{-(i,j)}^{t,t+1},\pi_i^{t+1},\pi_j^t}+\rho_i^{\pi_{-(i,j)}^{t,t+1},\pi_i^t,\pi_j^t}  \\
        =& \mathbb{E}_{s\sim \nu^{\pi_{-(i,j)}^{t,t+1},\pi_i^{t+1},\pi_j^{t+1}}} [\langle\overline{Q}_i^{\pi_{-(i,j)}^{t,t+1},\pi_i^t,\pi_j^{t+1}}(\cdot,s),\pi_i^{t+1}(\cdot|s) - \pi_i^t(\cdot|s)\rangle_{\mathcal{A}_i}]\\
        &- \mathbb{E}_{s\sim \nu^{\pi_{-(i,j)}^{t,t+1},\pi_i^{t+1},\pi_j^t}} [\langle\overline{Q}_i^{\pi_{-(i,j)}^{t,t+1},\pi_i^t,\pi_j^t}(\cdot,s),\pi_i^{t+1}(\cdot|s) - \pi_i^t(\cdot|s)\rangle_{\mathcal{A}_i}] \\
        =&\mathbb{E}_{s\sim \nu^{\pi_{-(i,j)}^{t,t+1},\pi_i^{t+1},\pi_j^{t+1}}} [\langle\overline{Q}_i^{\pi_{-(i,j)}^{t,t+1},\pi_i^t,\pi_j^{t+1}}(\cdot,s)-\overline{Q}_i^{\pi_{-(i,j)}^{t,t+1},\pi_i^t,\pi_j^t}(\cdot,s),\pi_i^{t+1}(\cdot|s) - \pi_i^t(\cdot|s)\rangle_{\mathcal{A}_i}]\\
        &+ \sum_s(\nu^{\pi_{-(i,j)}^{t,t+1},\pi_i^{t+1},\pi_j^{t+1}}(s) - \nu^{\pi_{-(i,j)}^{t,t+1},\pi_i^{t+1},\pi_j^{t}}(s)) \langle\overline{Q}_i^{\pi_{-(i,j)}^{t,t+1},\pi_i^t,\pi_j^t}(\cdot,s),\pi_i^{t+1}(\cdot|s) - \pi_i^t(\cdot|s)\rangle_{\mathcal{A}_i}\\
        \geq& -\max_s\lVert\overline{Q}_i^{\pi_{-(i,j)}^{t,t+1},\pi_i^t,\pi_j^{t+1}}(\cdot,s)-\overline{Q}_i^{\pi_{-(i,j)}^{t,t+1},\pi_i^t,\pi_j^t}(\cdot,s) \rVert_{\infty}\max_s\lVert\pi_i^{t+1}(\cdot|s) - \pi_i^t(\cdot|s)\rVert_1\\
        &-\kappa S \lVert\nu^{\pi_{-(i,j)}^{t,t+1},\pi_i^{t+1},\pi_j^t} - \nu^{\pi_{-(i,j)}^{t,t+1},\pi_i^{t+1},\pi_j^{t+1}} \rVert_{\infty}\max_s\lVert \pi_i^{t+1}(\cdot|s) - \pi_i^t(\cdot|s)\rVert_1\\
        \geq & -\kappa_Q\lVert\pi_j^{t+1}-\pi_j^t\rVert_{1,\infty}\lVert\pi_i^{t+1}-\pi_i^t\rVert_{1,\infty} - \kappa^2S\lVert\pi_j^{t+1}-\pi_j^t\rVert_{1,\infty}\lVert\pi_i^{t+1}-\pi_i^t\rVert_{1,\infty}\\
        = &-(\kappa_Q+S\kappa^2)\lVert\pi_j^{t+1}-\pi_j^t\rVert_{1,\infty}\lVert\pi_i^{t+1}-\pi_i^t\rVert_{1,\infty}\\
        \overset{(a)}{\geq} & -\frac{\kappa_Q+S\kappa^2}{2}(\lVert\pi_j^{t+1}-\pi_j^t\rVert_{1,\infty}^2+\lVert\pi_i^{t+1}-\pi_i^t\rVert_{1,\infty}^2), 
    \end{align*}
    where (a) is due to $ab\leq\frac{a^2+b^2}{2}$ for any $a,b$.  
    Therefore, we have
    \begin{align*}
    \text{Diff}_2
    &\geq -\frac{(N-1)(\kappa_Q+S\kappa^2)}{2}\sum_{i=1}^N\lVert\pi_i^{t+1}-\pi_i^t\rVert_{1,\infty}^2\\
    &\geq -\frac{(N-1)(\kappa_Q+S\kappa^2)A_{\max}}{2(1-\Gamma)}\sum_{i=1}^N\mathbb{E}_{s\sim\nu^{\pi_i^{t+1}, \pi^t_{-i}}} \lVert\pi_i^{t+1}(\cdot|s)-\pi_i^{t}(\cdot|s)\rVert_2^2.
    \end{align*}
    
    To bound $\text{Diff}_1$, we write
    \begin{equation}\label{eq:13}
    \begin{aligned}
        \text{Diff}_1 &= \sum_{i=1}^N\rho_i^{\pi_i^{t+1},\pi_{-i}^t} - \rho_i^{\pi_i^{t},\pi_{-i}^t} = \sum_{i=1}^N\mathbb{E}_{s\sim\nu^{\pi_i^{t+1}, \pi^t_{-i}}}[\langle\overline{Q}_i^{\pi^t}(s,\cdot), \pi^{t+1}_i(\cdot|s)-\pi^t_i(\cdot|s)\rangle_{\mathcal{A}_i}] \\
        &\geq \frac{1}{\beta}\sum_{i=1}^N\mathbb{E}_{s\sim\nu^{\pi_i^{t+1}, \pi^t_{-i}}} \lVert\pi_i^{t+1}(\cdot|s)-\pi_i^{t}(\cdot|s)\rVert_2^2.
    \end{aligned}
    \end{equation}
    The last inequality comes from the optimality criterion of the update rule in \cref{alg:3}. The update $\pi_i^{t+1}(\cdot|s)\in \arg\max_{p(\cdot|s)\in\Delta(\mathcal{A}_i)}\{\beta\langle\overline{Q}_i^{\pi_t}(s,\cdot),p(\cdot|s)\rangle_{A_i} - \frac{1}{2}\lVert p(\cdot|s)-\pi_i^t(\cdot|s)\rVert_2^2\} $ is a concave maximization problem. Therefore, $\beta\overline{Q}^{\pi_t}_i(s,\cdot)-\pi_i^{t+1}(\cdot|s)+\pi_i^t(\cdot|s)$ is not an increasing direction:
    \begin{equation}\label{eq:opt}
        \langle\beta\overline{Q}^{\pi_t}_i(s,\cdot)-\pi_i^{t+1}(\cdot|s)+\pi_i^t(\cdot|s),p(\cdot|s)-\pi_i^{t+1}(\cdot|s)\rangle_{\mathcal{A}_i}\leq 0,\ \forall p(\cdot|s)\in\Delta(\mathcal{A}_i).
    \end{equation}
    The last inequality of \cref{eq:13} is derived by substituting $p=\pi^t_i$ in the above inequality.
\end{proof}

\begin{lemma}[policy improvement (b)]\label{lm:22} Let $\pi^t$ to be the policy at time $t$ of \cref{alg:3}, 
    \begin{equation*}
        \Phi(\pi^{t+1}) - \Phi(\pi^t)\geq\frac{1}{\beta}(1-\beta\frac{L_{\Phi}}{1-\Gamma})\sum_{i=1}^N\mathbb{E}_{s\sim\nu^{\pi_i^{t+1}, \pi^t_{-i}}} \lVert\pi_i^{t+1}(\cdot|s)-\pi_i^{t}(\cdot|s)\rVert_2^2.
    \end{equation*}
\end{lemma}
\begin{proof}
    Bound each term in $\text{Diff}_2$ of \cref{eq:12}:
    \begin{equation}
    \begin{aligned}
        &\Phi(\pi_{-(i,j)}^{t,t+1},\pi_i^{t+1},\pi_j^{t+1})-\Phi(\pi_{-(i,j)}^{t,t+1},\pi_i^t,\pi_j^{t+1}) - \Phi(\pi_{-(i,j)}^{t,t+1},\pi_i^{t+1},\pi_j^t)+\Phi(\pi_{-(i,j)}^{t,t+1},\pi_i^t,\pi_j^t) \\ 
        = & \underbrace{\rho_i^{\pi_{-(i,j)}^{t,t+1},\pi_i^{t+1},\pi_j^{t+1}}-\rho_i^{\pi_{-(i,j)}^{t,t+1},\pi_i^t,\pi_j^{t+1}}}_{I_1} -\underbrace{(\rho_i^{\pi_{-(i,j)}^{t,t+1},\pi_i^{t+1},\pi_j^t}-\rho_i^{\pi_{-(i,j)}^{t,t+1},\pi_i^{t},\pi_j^t})}_{I_2}.\\
    \end{aligned}
    \end{equation}
    From the derivative of \cref{lm:smooth} $\frac{\partial^2\rho_i^{\pi}}{\partial\pi_i}$ is $\frac{L_{\Phi}}{N}$-Lipschitz w.r.t. $\pi_i$ or $\pi_j$, for any $\pi_i$, $\pi_j$. We aim to bound $I_1-I_2$ with l2 norms of $\lVert \pi_i^t-\pi_i^{t+1}\rVert_2$ and $\lVert\pi_j^t-\pi_j^{t+1}\rVert_2$. Using the interpolation for a differentiable function, there exists $a,b\in[0,1]$, such that
    \begin{equation*}
        \begin{aligned}
            I_1 =&\rho_i^{\pi_{-(i,j)}^{t,t+1},\pi_i^{t+1},\pi_j^{t+1}}-\rho_i^{\pi_{-(i,j)}^{t,t+1},\pi_i^t,\pi_j^{t+1}} = \langle \frac{\partial \rho_i^{\pi}}{\partial \pi_i}\left(\pi_{-(i,j)}^{t,t+1},\pi^{t+1}_i+a(\pi_i^{t}-\pi_i^{t+1}),\pi_j^{t+1}\right),\pi_i^{t+1}-\pi_i^t\rangle, \\
            I_2 =&\rho_i^{\pi_{-(i,j)}^{t,t+1},\pi_i^{t+1},\pi_j^{t}}-\rho_i^{\pi_{-(i,j)}^{t,t+1},\pi_i^t,\pi_j^{t}} = \langle \frac{\partial \rho_i^{\pi}}{\partial \pi_i}\left(\pi_{-(i,j)}^{t,t+1},\pi^{t+1}_i+b(\pi_i^{t}-\pi_i^{t+1}),\pi_j^{t}\right),\pi_i^{t+1}-\pi_i^t\rangle ,\\
            I_1-I_2 =& \langle \frac{\partial \rho_i^{\pi}}{\partial \pi_i}\left(\pi_{-(i,j)}^{t,t+1},\pi^{t+1}_i+a(\pi_i^{t}-\pi_i^{t+1}),\pi_j^{t+1}\right)-\frac{\partial \rho_i^{\pi}}{\partial \pi_i}\left(\pi_{-(i,j)}^{t,t+1},\pi^{t+1}_i+a(\pi_i^{t}-\pi_i^{t+1}),\pi_j^{t}\right),\pi_i^{t+1}-\pi_i^t\rangle \\
            & + \langle \frac{\partial \rho_i^{\pi}}{\partial \pi_i}\left(\pi_{-(i,j)}^{t,t+1},\pi^{t+1}_i+a(\pi_i^{t}-\pi_i^{t+1}),\pi_j^{t}\right)-\frac{\partial \rho_i^{\pi}}{\partial \pi_i}\left(\pi_{-(i,j)}^{t,t+1},\pi^{t+1}_i+b(\pi_i^{t}-\pi_i^{t+1}),\pi_j^{t}\right),\pi_i^{t+1}-\pi_i^t\rangle \\
            \geq& -\lVert\frac{\partial \rho_i^{\pi}}{\partial \pi_i}(\pi_{-(i,j)}^{t,t+1},\pi^{t+1}_i+a(\pi_i^{t}-\pi_i^{t+1}),\pi_j^{t+1})-\frac{\partial \rho_i^{\pi}}{\partial \pi_i}(\pi_{-(i,j)}^{t,t+1},\pi^{t+1}_i+a(\pi_i^{t}-\pi_i^{t+1}),\pi_j^{t})\rVert_2\lVert\pi_i^{t+1}-\pi_i^t\rVert_2 \\
            & - \lVert\frac{\partial \rho_i^{\pi}}{\partial \pi_i}(\pi_{-(i,j)}^{t,t+1},\pi^{t+1}_i+a(\pi_i^{t}-\pi_i^{t+1}),\pi_j^{t})-\frac{\partial \rho_i^{\pi}}{\partial \pi_i}(\pi_{-(i,j)}^{t,t+1},\pi^{t+1}_i+b(\pi_i^{t}-\pi_i^{t+1}),\pi_j^{t})\rVert_2\lVert\pi_i^{t+1}-\pi_i^t\rVert_2 \\
            \geq& -\frac{L_{\Phi}}{N}\lVert\pi_j^{t+1}-\pi_j^t\rVert_2\lVert\pi_i^{t+1}-\pi_i^t\rVert_2 - \frac{L_{\Phi}}{N}|a-b|\lVert\pi_i^t-\pi_i^{t+1}\rVert_2\lVert\pi_i^{t+1}-\pi_i^t\rVert_2 \\
            \geq& -\frac{L_{\Phi}}{2N}(\lVert\pi_j^{t+1}-\pi_j^t\rVert_2^2+\lVert\pi_i^{t+1}-\pi_i^t\rVert_2^2) - \frac{L_{\Phi}}{N}\lVert\pi_i^t-\pi_i^{t+1}\rVert_2^2\\
            =& -\frac{L_{\Phi}}{2N}\lVert\pi_j^{t+1}-\pi_j^t\rVert_2^2 - \frac{3L_{\Phi}}{2N}\lVert\pi_i^{t+1}-\pi_i^t\rVert_2^2.
        \end{aligned}
    \end{equation*}
    Since the permutation of agents' indexes does not change the above result, we have:
    \begin{align*}
        \text{Diff}_2 &\geq -(N-1)\frac{L_{\Phi}}{N}\sum_i \lVert\pi_i^{t+1}-\pi_i^t\rVert_2^2\\
        &\geq -L_{\Phi}\sum_i \sum_s\lVert\pi_i^{t+1}(\cdot|s)-\pi_i^t(\cdot|s)\rVert_2^2\\
        &\geq-\frac{L_{\Phi}}{1-\Gamma}\sum_{i=1}^N\mathbb{E}_{s\sim\nu^{\pi_i^{t+1}, \pi^t_{-i}}} \lVert\pi_i^{t+1}(\cdot|s)-\pi_i^{t}(\cdot|s)\rVert_2^2.
    \end{align*}
    Therefore, by the same bound for $\text{Diff}_1$ in \cref{lm:21} we can lower bound the policy improvement as
    \begin{equation*}
        \Phi(\pi^{t+1}) - \Phi(\pi^t)\geq(\frac{1}{\beta}-\frac{L_{\Phi}}{1-\Gamma})\sum_{i=1}^N\mathbb{E}_{s\sim\nu^{\pi_i^{t+1}, \pi^t_{-i}}} \lVert\pi_i^{t+1}(\cdot|s)-\pi_i^{t}(\cdot|s)\rVert_2^2.
    \end{equation*}
\end{proof}

\begin{theorem}[Restatement of \cref{thm:3}]
If $\beta\leq\max\{\frac{1-\Gamma}{(N-1)(\kappa_Q+S\kappa^2)A_{\max}},\frac{1-\Gamma}{2L_{\Phi}}\}$, \cref{alg:3} has a bounded Nash regret:
    \begin{equation}
        \textnormal{Nash-regret}(T)\leq \sqrt{D}(\kappa\sqrt{A_{\max}}+\frac{2}{\beta})\sqrt{2\beta C_{\Phi}}\frac{1}{\sqrt{T}}.
    \end{equation}
\end{theorem}
\begin{proof}
    \begin{equation*}
    \begin{aligned}
        \text{Nash-gap}(t) &= \max_i(\max_{\pi_i'}\rho_i^{\pi_i',\pi_{-i}^t} - \rho_i^{\pi_i^t,\pi_{-i}^t}) \\
        &\overset{(1)}{=}\mathbb{E}_{s\sim\nu^{\pi_i, \pi^t_{-i}}}[\langle\overline{Q}_i^{\pi^t}(s,\cdot), \pi_i(\cdot|s)-\pi^t_i(\cdot|s)\rangle_{\mathcal{A}_i}] \\
        &=\mathbb{E}_{s\sim\nu^{\pi_i, \pi^t_{-i}}}[\underbrace{\langle\overline{Q}^{\pi^t}(s,\cdot), \pi_i^{t+1}(\cdot|s)-\pi^t_i(\cdot|s)\rangle_{\mathcal{A}_i}}_{I_1}+\underbrace{\langle\overline{Q}^{\pi^t}(s,\cdot), \pi_i(\cdot|s)-\pi^{t+1}_i(\cdot|s)\rangle_{\mathcal{A}_i}}_{I_2}].
    \end{aligned}   
    \end{equation*}
    In $(1)$, we use $\pi_i$ to represent the policy that achieves $\arg\max_{\pi'_i}$ in the previous expression and assume that $i$ attains the maximum in $\max_i$. We will bound $I_1$ and $I_2$ separately. 
    
    Recall \cref{eq:opt} for any $p$ in the feasible policy set,
        $\langle\beta\overline{Q}^{\pi_t}_i(s,\cdot)-\pi_i^{t+1}(\cdot|s)+\pi_i^t(\cdot|s),p(\cdot|s)-\pi_i^{t+1}(\cdot|s)\rangle_{\mathcal{A}_i}\leq 0$. We can bound $I_1$ and $I_2$ as
    \begin{equation}
    \begin{split}
        I_2 &= \langle\overline{Q}_i^{\pi^t}(s,\cdot), \pi_i(\cdot|s)-\pi^{t+1}_i(\cdot|s)\rangle_{\mathcal{A}_i}\\
        &\leq\ \frac{1}{\beta}\langle\pi_i^{t+1}(\cdot|s)-\pi_i^t(\cdot|s),\pi_i(\cdot|s)-\pi_i^{t+1}(\cdot|s)\rangle_{\mathcal{A}_i}\\
        &\overset{(a)}{\leq}\frac{1}{\beta}\lVert\pi_i^{t+1}(\cdot|s)-\pi_i^t(\cdot|s)\rVert_{\infty}\lVert\pi_i(\cdot|s)-\pi_i^{t+1}(\cdot|s) \rVert_1\\
        &\leq \frac{2}{\beta}\lVert\pi_i^{t+1}(\cdot|s)-\pi_i^t(\cdot|s) \rVert_{\infty}\\
        &\leq \frac{2}{\beta}\lVert\pi_i^{t+1}(\cdot|s)-\pi_i^t(\cdot|s) \rVert_2,\\
        I_1 &= \langle\overline{Q}_i^{\pi^t}(s,\cdot), \pi_i^{t+1}(\cdot|s)-\pi^t_i(\cdot|s)\rangle_{\mathcal{A}_i} \\
        &\overset{(b)}{\leq} \kappa\lVert\pi_i^{t+1}(\cdot|s)-\pi^t_i(\cdot|s) \rVert_1\\
        &\leq\kappa\sqrt{A_i}\lVert\pi_i^{t+1}(\cdot|s)-\pi^t_i(\cdot|s) \rVert_2.
    \end{split}
    \end{equation}
    (a) and (b) result from $\langle x,y\rangle\leq\lVert x\rVert_1\lVert y\rVert_{\infty}$. 

    Therefore we can bound the Nash-gap as:
    \begin{equation*}
    \begin{split}
        \text{Nash-gap}(t) &\leq \sum_{s}\nu^{\pi_i,\pi_{-i}^t}(s)(\kappa\sqrt{A_i}+\frac{2}{\beta})\lVert\pi_i^{t+1}(\cdot|s)-\pi^t_i(\cdot|s) \rVert_2 \\
        &\leq \sum_{s}\nu^{\pi_i,\pi_{-i}^t}(s)(\kappa\sqrt{A_{\max}}+\frac{2}{\beta})\lVert\pi_i^{t+1}(\cdot|s)-\pi^t_i(\cdot|s) \rVert_2\\
        &\leq\sqrt{D}(\kappa\sqrt{A_{\max}}+\frac{2}{\beta})\sum_{s}\sqrt{\nu^{\pi_i,\pi_{-i}^t}(s)}\sqrt{\nu^{\pi^{t+1}_i,\pi_{-i}^t}(s)\lVert\pi_i^{t+1}(\cdot|s)-\pi^t_i(\cdot|s) \rVert_2^2},\\
        \sum_{t=0}^{T-1}\text{Nash-gap}(t)&\overset{(a)}{\leq} \sqrt{D}(\kappa\sqrt{A_{\max}}+\frac{2}{\beta})\sqrt{\sum_{t=0}^{T-1}\sum_s \nu^{\pi^{t+1}_i,\pi_{-i}^t}(s)}\sqrt{\sum_{t=0}^{T-1}\sum_{s}\sum_{i}\nu^{\pi^{t+1}_i,\pi_{-i}^t}(s)\lVert\pi_i^{t+1}(\cdot|s)-\pi^t_i(\cdot|s) \rVert_2^2}\\
        &=\sqrt{D}(\kappa\sqrt{A_{\max}}+\frac{2}{\beta})\sqrt{T}\sqrt{\sum_{t=0}^{T-1}\sum_{s}\sum_{i}\nu^{\pi^{t+1}_i,\pi_{-i}^t}(s)\lVert\pi_i^{t+1}(\cdot|s)-\pi^t_i(\cdot|s) \rVert_2^2}.\\
    \end{split}
    \end{equation*}

    (a) is due to the Cauchy-Schwarz inequality.

    From \cref{lm:21} and \cref{lm:22}, we have:
    \begin{equation*}
        \Phi(\pi^{t+1}) - \Phi(\pi^t) \geq \frac{1}{\beta}\left(1-\beta\min\{\frac{(N-1)(\kappa_Q+S\kappa^2)A_{\max}}{2(1-\Gamma)},\frac{L_{\Phi}}{1-\Gamma}\}\right)\sum_{i=1}^N \sum_s\nu^{\pi_i^{t+1},\pi_{-i}^t}(s)\lVert\pi_i^{t+1}(\cdot|s)-\pi_i^{t}(\cdot|s)\rVert_2^2.
    \end{equation*}
    
    Therefore, by substituting learning rate $\beta\leq\frac{1}{2}\max\{\frac{2(1-\Gamma)}{(N-1)(\kappa_Q+S\kappa^2)A_{\max}},\frac{1-\Gamma}{L_{\Phi}}\}$,
    \begin{equation*}
    \begin{aligned}
        \sum_{t=0}^{T-1}\text{Nash-gap}(t)&\leq \sqrt{D}(\kappa\sqrt{A_{max}}+\frac{2}{\beta})\sqrt{T}\sqrt{\sum_{t=0}^{T-1}2\beta(\Phi(\pi^{t+1})-\Phi(\pi^t))}\\
        &\leq \sqrt{D}(\kappa\sqrt{A_{max}}+\frac{2}{\beta})\sqrt{T}\sqrt{2\beta C_{\Phi}}.
    \end{aligned}
    \end{equation*}    
\end{proof}

\subsection{Sample Complexity of proximal-Q}\label{C.1}
\begin{algorithm}[t!]
    \caption{proximal-Q algorithm with sample estimates}
    \label{alg:6}
    \begin{algorithmic}[1]
    \STATE \textbf{Input:} learning rate $\beta$, gradient estimation parameters $B$, $N_1$
    \STATE \textbf{Initialization:} $\pi^{(0)}_i(a_i|s)=1/A_i$ for any $s\in\mathcal{S}$, $a_i\in\mathcal{A}_i$
    \FOR {$t=0$ to $T-1$ }
        \STATE all agents take action independently and synchronously for $B$ time steps to collect trajectories $\{\mathcal{T}_i^t\}$ 
            \FOR{agent $i$}
            \FOR{$s\in\mathcal{S}$}
            \STATE $\hat{q}_i^t(s,\cdot)\gets \text{Q estimation}(\mathcal{T}_i^t, s, \pi_i^t, B, N_1)$
            \ENDFOR
            \STATE $\pi_i^{(t+1)}(\cdot|s)=\underset{p(\cdot|s)\in\triangle_{\alpha}(\mathcal{A}_i)}{\arg\max} \{\beta\langle\hat{q}^{\pi^t}_i(s,\cdot), p(\cdot|s)\rangle_{\mathcal{A}_i} - \frac{1}{2}\lVert p(\cdot|s)-\pi_i^t(\cdot|s)\rVert_2^2 \},\ \forall s\in\mathcal{S}$
            \ENDFOR
        \ENDFOR
    \end{algorithmic}
\end{algorithm}

\begin{algorithm}[t!]
    \caption{Q estimation \citealp[Lemma 6]{wei2020model}}
    \label{alg:7}
    \begin{algorithmic}[1]
    \STATE \textbf{Input:} trajectory $\mathcal{T}=(s^0,a_i^0,r_i^0,...,s^B,a_i^B,r_i^B)$, state $s$, policy $\pi_i$, parameters $B$ and $N_1$ 
    \STATE $\tau\gets 0$
    \STATE $k\gets 0 $
    \WHILE{$\tau\leq B-N_1$}
        \IF{$s^{\tau}=s$}
            \STATE $k\gets k+1$
            \STATE $R\gets\sum_{t=\tau}^{\tau+N_1-1}r_i^{t}$
            \STATE $y_k\gets\frac{R}{\pi_i(a_i^{\tau}|s)}\mathbf{1}[a=a_i^{\tau}]$  ($y_k\in\mathbb{R}^{A_i}$)
            \STATE $\tau\gets\tau+2N_1$
        \ELSE
            \STATE $\tau\gets\tau+1$
        \ENDIF
    \ENDWHILE
    \IF{$k\neq0$}
        \STATE \textbf{return} $\frac{1}{k}\sum_{j=1}^ky_j$
    \ELSE
        \STATE \textbf{return} $\mathbf{0}$
    \ENDIF
    \end{algorithmic}
\end{algorithm}

\begin{lemma}[\text{\citealp[Lemma 6]{wei2020model}}]\label{lm:wei}
    Let $\mathbb{E}_t[x]$ denote the expectation of a random variable x conditioned on all history before episode $t$ (note that $\pi^t$ is updated at the end of episode $t-1$). If $B$ is large enough, such that there exists $N_2<B$ with $C_p\varrho^{N_2}\leq\frac{1}{2}(1-\Gamma)$, then for any $t$, $s$, $a_i$, the estimated $\hat{q}$ from \cref{alg:7} satisfies:
    
    \begin{equation}
    \begin{aligned}
        \mathbb{E}_t[(\hat{q}_i^t(s,a_i)-(\overline{Q}_i^{\pi^t}(s,a_i)+N_1\rho^{\pi^t}_i))^2]\leq&6(1+\frac{2C_p\varrho^{2N_1}}{1-\Gamma}\frac{B}{2N_1})\left(\frac{N_1^2}{\alpha}+\frac{C_p^2}{(1-\varrho)^2}+\frac{C_p^2\varrho^{2N_1}}{(1-\varrho)^2}\right)\frac{(1-\frac{1-\Gamma}{2})^{n'}+\frac{2N_1+n'N_2}{B}}{1-(1-\frac{1-\Gamma}{2})^{\lfloor\frac{B-N_1}{N_2}\rfloor}}\\
        &+(1+\frac{2C_p\varrho^{2N_1}}{1-\Gamma}\frac{B}{2N_1})(\frac{2C_p^2\varrho^{2N_1}}{(1-\varrho)^2}+(1-\frac{1-\Gamma}{2})^{\lfloor\frac{B-N_1}{N_2}\rfloor}\frac{C_p}{1-\varrho}).
    \end{aligned}
    \end{equation}
\end{lemma}
For completeness, we provide a brief proof below.
\begin{proof}
    We define some notation first. Let $\tau_j$ be the evoked time at line 5 ($s^{\tau_j}=s$), $w_j$ be the waiting time, where $w_j=\tau_j-(\tau_{j-1}+2N_1)$ for $j>1$, and $w_1=\tau_1$ for $j=1$. Furthermore, let $q_i^{\pi}(s,a_i)=\overline{Q}_i^{\pi}(s,a)+N_1\rho^{\pi}$, $\hat{q}_{i,j}^{\pi}(s,\cdot)=y_j(\cdot)$ where $y_j$ is defined in line 8 of \cref{alg:7}, and $\hat{q}_i^{\pi}(s,\cdot)=\frac{1}{k}\sum_{j=1}^k\hat{q}_{i,j}^{\pi}(s,\cdot)$ if $k>0$; otherwise, $\hat{q}_i^{\pi}=\mathbf{0}$. 

    The main difficulty of analyzing the bias and variance of $\hat{q}_i^{\pi}(s,a_i)$ lies in the random number $k$, which is the times state $s$ is visited (used in line 14 of \cref{alg:7}),  determined by $\{w_1,w_2,...\}$ only. In the proof of \cite{wei2020model}, they first calculate the bias and variance under an "imaginary" world, where the state distribution is reset to $\nu^{\pi}$ at any time $\tau_j+2N_1$. Then, they demonstrate that the event $\{\tau_1,\tau_2,...\}$ has a similar probability measure between the real world and the imaginary world. Since $\tau_{j+1}$ and $\tau_j$ are independent, it is easier to bound the bias and variance under an imaginary world. We use $\mathbb{E}'$ to denote the expectation in the imaginary world and $\mathbb{E}$ for the real world.

    \textbf{Step 1: Bound the bias and variance under imaginary world}
    \begin{align*}
        \mathbb{E}'[\hat{q}_i^{\pi}(s,a_i)]=&\mathbb{P}(w_1\leq B-N_1)\mathbb{E}'[\frac{1}{k}\sum_{j=1}^k\mathbb{E}'[\hat{q}_{i,j}^{\pi}(s,a_i)|{w_1}]|w_1\leq B-N]+\mathbb{P}(w_1> B-N_1)\times0\\
        \overset{(a)}{=}&\mathbb{P}(w_1\leq B-N_1)\mathbb{E}'[\frac{1}{k}(\sum_{j=1}^kq_i^{\pi}(s,a_i)-\delta(s,a_i))] \\ 
        \overset{(b)}{=}&q^{\pi}_i(s,a_i)-\delta'(s,a_i).
    \end{align*}
    
    In (a), tail is defined as $\delta(s,a_i):=\mathbb{E}_{P,\pi}[\sum_{t=N_1+1}^{\infty}(r(s,a)-\rho^{\pi})|s^0=s,a_i^0=a_i]$. It is easy to bound it by $|\delta(s,a_i)|\leq \sum_{t=N_1}^{\infty}C_p\varrho^t=C_p\varrho^{N_1}\frac{1}{1-\varrho}$. In (b), $\delta'(s,a_i)=(1-\mathbb{P}(w_1\leq B-N_1))(q_i^{\pi}(s,a_i)-\delta(s,a_i))+\delta(s,a_i)$.

    To give a bound for $\delta'(s,a_i)$, let's analyze $|q_i^{\pi}(s,a_i)|$ and $\mathbb{P}(w_1\leq B-N_1)$ separately. By \cref{prop:3} $|q_i^{\pi}(s,a_i)|\leq C_p\kappa_0+N_1$. $1-\mathbb{P}(w_1\leq B-N_1)$ is the probability of never visiting $s$ during time $0$ to time $B-N_1$. If $B$ is large enough, there exists $N_2<B$ such that $C_p\varrho^{N_2}\leq\frac{1}{2}(1-\Gamma)$, which means $|\mathbb{P}(s^{N_2}=s|s^0=s')-\nu^{\pi}(s)|\leq\frac{1}{2}(1-\Gamma)$ and $\mathbb{P}(s^{N_2}=s|s^0=s')\geq\nu^{\pi}(s)-\frac{1}{2}(1-\Gamma)\geq\frac{1}{2}(1-\Gamma)$ for any $s'$. Therefore, $\mathbb{P}(w_1>B-N_1)\leq(1-\frac{1-\Gamma}{2})^{\lfloor\frac{B-N_1}{N_2}\rfloor}$.
    \begin{align*}
         |\mathbb{E}'[\hat{q}_i^{\pi}(s,a_i)]-q_i^{\pi}(s,a_i)|=&|\delta'(s,a_i)| \\
         \leq & (1-\frac{1-\Gamma}{2})^{\lfloor\frac{B-N_1}{N_2}\rfloor}(C_p\kappa_0+N_1)+C_p\varrho^{N_1}\frac{1}{1-\varrho}.
    \end{align*}

    To bound the variance, denote $\Delta_j=\hat{q}_{i,j}^{\pi}(s,a_i)-q_i^{\pi}(s,a_i)+\delta(s,a_i)$. Then $\mathbb{E}'[\Delta_j|{w_j}]=0$.
    \begin{align*}        &\mathbb{E}'[(\hat{q}_i^{\pi}(s,a_i)-q_i^{\pi}(s,a_i))^2]\\
        =        &\mathbb{P}(w_1\leq B-N_1)\mathbb{E}'[(\hat{q}_i^{\pi}(s,a_i)-q_i^{\pi}(s,a_i))^2|w_1\leq B-N_1]+\mathbb{P}(w_1>B-N_1)|q_i^{\pi}(s,a_i)|^2\\
        \leq&\mathbb{E}'[(\frac{1}{k}\sum_{j=1}^k\Delta_j-\delta(s,a_i))^2|w_1\leq B-N_1]+(1-\frac{1-\Gamma}{2})^{\lfloor\frac{B-N_1}{N_2}\rfloor}(C_p\kappa_0+N_1)^2\\
        \leq&\mathbb{E}'[2(\frac{1}{k}\sum_{j=1}^k\Delta_j)^2+2\delta(s,a_i))^2|w_1\leq B-N_1]+(1-\frac{1-\Gamma}{2})^{\lfloor\frac{B-N_1}{N_2}\rfloor}(C_p\kappa_0+N_1)^2\\
        \leq&\mathbb{E}'[2(\frac{1}{k}\sum_{j=1}^k\Delta_j)^2|w_1\leq B-N_1]+\frac{2C_p^2\varrho^{2N_1}}{(1-\varrho)^2}+(1-\frac{1-\Gamma}{2})^{\lfloor\frac{B-N_1}{N_2}\rfloor}(C_p\kappa_0+N_1)^2\\
        \leq&\mathbb{E}'[\frac{2}{k^2}\sum_{j=1}^k\mathbb{E}'[\Delta_j^2|{w_1}]|w_1\leq B-N_1]+\frac{2C_p^2\varrho^{2N_1}}{(1-\varrho)^2}+(1-\frac{1-\Gamma}{2})^{\lfloor\frac{B-N_1}{N_2}\rfloor}(C_p\kappa_0+N_1)^2\\
        \overset{(a)}{\leq}&6\left(\frac{N_1^2}{\pi_i(a_i|s)}+2C_p^2\kappa_0^2\pi_i(a_i|s)+2N_1^2\pi_i(a_i|s)+\frac{C_p^2\varrho^{2N_1}\pi_i(a_i|s)}{(1-\varrho)^2}\right)\mathbb{E}'[\frac{1}{k}|w_1\leq B-N_1]\\
        &+\frac{2C_p^2\varrho^{2N_1}}{(1-\varrho)^2}+(1-\frac{1-\Gamma}{2})^{\lfloor\frac{B-N_1}{N_2}\rfloor}(C_p\kappa_0+N_1)^2.\\
    \end{align*}
    
    In (a) we use $\mathbb{E}'[\Delta_j^2|{w_1}]\leq \pi_i(a_i|s)(3\frac{N_1^2}{\pi_i(a_i|s)^2}+3(2C_p^2\kappa_0^2+2N_1^2)+3\frac{C_p^2\varrho^{2N_1}}{(1-\varrho)^2})$. 

    If trajectory length $B$ is large enough, there exists a waiting period length $n'N_2$ such that $k_0=\lfloor\frac{B}{2N_1+n'N_2}\rfloor>1$, $\mathbb{P}(k\leq k_0)\leq\mathbb{P}(w_1\geq n'N_2)\leq(1-\frac{1-\Gamma}{2})^{n'}$ is small enough. Then we can bound the average visiting time and variance by:
    \begin{align*}
        \mathbb{E}'[\frac{1}{k}|w_1\leq B-N_1]\leq&\frac{\mathbb{P}(k\leq k_0)\times1+\mathbb{P}(k>k_0)\frac{1}{k_0}}{\mathbb{P}(w_1\leq B-N_1)}\\
        \leq&\frac{(1-\frac{1-\Gamma}{2})^{n'}+\frac{2N_1+n'N_2}{B}}{1-(1-\frac{1-\Gamma}{2})^{\lfloor\frac{B-N_1}{N_2}\rfloor}},
    \end{align*}
    \begin{align*}
        \mathbb{E}'[(\hat\beta_i^{\pi}(s,a)-\beta_i^{\pi}(s,a_i))^2]\leq&
        6\left(\frac{N_1^2}{\pi_i(a_i|s)}+2C_p^2\kappa_0^2\pi_i(a_i|s)+2N_1^2\pi_i(a_i|s)+\frac{C_p^2\varrho^{2N_1}\pi_i(a_i|s)}{(1-\varrho)^2}\right)\frac{(1-\frac{1-\Gamma}{2})^{n'}+\frac{2N_1+n'N_2}{B}}{1-(1-\frac{1-\Gamma}{2})^{\lfloor\frac{B-N_1}{N_2}\rfloor}}\\
        &+\frac{2C_p^2\varrho^{2N_1}}{(1-\varrho)^2}+(1-\frac{1-\Gamma}{2})^{\lfloor\frac{B-N_1}{N_2}\rfloor}\frac{C_p}{1-\varrho}.\\
    \end{align*}
    
    \textbf{Step2: bound the difference between imaginary world and real world}
    
    Since $\hat{q}_i^{\pi}(s,a_i)$ is determined by $X=(k,\tau_1,\mathcal{T}_1,\tau_2,\mathcal{T}_2,...,\tau_k,\mathcal{T}_k)$, let $\hat{q}_i^{\pi}(s,a_i)=f(X)$. Then $\frac{\mathbb{E}[\hat{q}_i^{\pi}(s,a_i)]}{\mathbb{E}'[\hat{q}_i^{\pi}(s,a_i)]}=\frac{\sum_x f(x)\mathbb{P}(X=x)}{\sum_x f(x)\mathbb{P}'(X=x)}\leq \max_x\frac{\mathbb{P}(X=x)}{\mathbb{P}'(X=x)}$. We can bound $ \frac{\mathbb{P}(X=x)}{\mathbb{P}'(X=x)}$, $\forall x$ as follow:
    \begin{align*}
        \frac{\mathbb{P}(X=x)}{\mathbb{P}'(X=x)}=&\frac{\mathbb{P}(\tau_2|\tau_1,\mathcal{T}_1)...\mathbb{P}(\tau_{k}|\tau_{k-1},\mathcal{T}_{k-1})}{\mathbb{P}'(\tau_2|\tau_1,\mathcal{T}_1)...\mathbb{P}'(\tau_{k}|\tau_{k-1},\mathcal{T}_{k-1})}\\
        \overset{(a)}{\leq}&(\max_{s'}\frac{\mathbb{P}(s^{\tau_1+2N_1}=s'|\tau_1)}{\nu^{\pi}(s')})\dots(\max_{s'}\frac{\mathbb{P}(s^{\tau_{k-1}+2N_1}=s'|\tau_{k-1})}{\nu^{\pi}(s')})\\
        \leq&(1+\frac{C_p\varrho^{2N_1}}{1-\Gamma})^{\frac{B}{2N_1}}\leq e^{\frac{C_p\varrho^{2N_1}}{1-\Gamma}\frac{B}{2N_1}}\leq 1+\frac{2C_p\varrho^{2N_1}}{1-\Gamma}\frac{B}{2N_1}.
    \end{align*}

    Here $(a)$ is derived by that:
    \begin{equation*}
        \begin{aligned}
            \mathbb{P}(\tau_{j+1}|\tau_{j},\mathcal{T}_{j}) &= \sum_{s'\neq s} \mathbb{P}(\tau_j+2N_1=s'|\tau_{j},\mathcal{T}_{j})\mathbb{P}(s^t\neq s, \forall\ t\in[\tau_j+2N_1+1, \tau_{j+1}-1], s^{\tau_2}=s|\tau_j+2N_1=s'),\\
            \mathbb{P}'(\tau_{j+1}|\tau_{j},\mathcal{T}_{j}) &= \sum_{s'\neq s} \nu^{\pi}(s')\mathbb{P}(s^t\neq s, \forall\ t\in[\tau_j+2N_1+1, \tau_{j+1}-1], s^{\tau_2}=s|\tau_j+2N_1=s'),
        \end{aligned}
    \end{equation*}
    for $\tau_{j+1}\neq\tau_j+2N_1$. When $\tau_{j+1}=\tau_j+2N_1$, we have:
    \begin{equation*}
            \mathbb{P}(\tau_{j+1}|\tau_{j},\mathcal{T}_{j}) = \mathbb{P}(\tau_j+2N_1=s|\tau_{j},\mathcal{T}_{j}), \ \mathbb{P}'(\tau_{j+1}|\tau_{j},\mathcal{T}_{j}) = \nu^{\pi}(s).
    \end{equation*}

    Therefore, the following result can be derived:
    \begin{align*}
        \mathbb{E}[(\hat{q}_i^{\pi}(s,a_i)-q_i^{\pi}(s,a_i))^2]\leq&\mathbb{E}'[(\hat{q}_i^{\pi}(s,a_i)-q_i^{\pi}(s,a_i))^2](1+\frac{C_p\varrho^{2N_1}}{\nu^{\pi}(s)}\frac{B}{N_1})\\
        \leq&6(1+\frac{C_p\varrho^{2N_1}}{1-\Gamma}\frac{B}{N_1})\left(\frac{N_1^2}{\alpha}+\frac{2C_p^2}{(1-\varrho)^2}+2N_1^2+\frac{C_p^2\varrho^{2N_1}}{(1-\varrho)^2}\right)\frac{(1-\frac{1-\Gamma}{2})^{n'}+\frac{2N_1+n'N_2}{B}}{1-(1-\frac{1-\Gamma}{2})^{\lfloor\frac{B-N_1}{N_2}\rfloor}}\\
        &+(1+\frac{C_p\varrho^{2N_1}}{1-\Gamma}\frac{B}{N_1})(\frac{2C_p^2\varrho^{2N_1}}{(1-\varrho)^2}+(1-\frac{1-\Gamma}{2})^{\lfloor\frac{B-N_1}{N_2}\rfloor}\frac{C_p}{1-\varrho}).
    \end{align*}
\end{proof}  

Let $n'=N_1=O(\log\frac{1}{\alpha\delta})$, $N_2=O(\log\frac{1}{1-\Gamma})$, $B=\tilde O(\frac{1}{\alpha\delta})$. For any agent $i$, time $t$, state $s$, and action $a_i$, $\mathbb{E}_t[(\hat{q}_i^{\pi}(s,a_i)-q_i^{\pi}(s,a_i))^2]\leq \delta$.

\begin{lemma}(Policy improvement)\label{lm:23}
\begin{align*}
    \Phi(\pi^{t+1})-\Phi(\pi^t)\geq&\left(\frac{1}{\beta}-\frac{1}{2}-\frac{L_{\Phi}}{1-\Gamma}\right)\sum_{i=1}^N\mathbb{E}_{s\sim\nu^{\pi_i^{t+1}, \pi^t_{-i}}} \lVert \pi_i^{t+1}(\cdot|s)-\pi_i^t(\cdot|s)\rVert_2^2\\
    &-\frac{1}{2}\sum_{i=1}^N\mathbb{E}_{s\sim\nu^{\pi_i^{t+1}, \pi^t_{-i}}}[\sum_{a_i} (\hat{q}_i^{\pi^t}(s,a_i)-q_i^{\pi^t}(s,a_i))^2].
\end{align*}
\end{lemma}
\begin{proof}    
    We use the same decomposition as \cref{eq:12}.

    Similar as \cref{lm:22}, $
        \text{Diff}_2\geq-\frac{L_{\Phi}}{1-\Gamma}\sum_{i=1}^N\mathbb{E}_{s\sim\nu^{\pi_i^{t+1}, \pi^t_{-i}}} \lVert\pi_i^{t+1}(\cdot|s)-\pi_i^{t}(\cdot|s)\rVert_2^2$.

    Note that $\pi_i^{t+1}(\cdot|s)=\underset{p(\cdot|s)\in\triangle_{\alpha}(\mathcal{A}_i)}{\arg\max} \{\beta\langle\hat{q}^{\pi^t}_i(s,\cdot), p(\cdot|s)\rangle_{\mathcal{A}_i} - \frac{1}{2}\lVert p(\cdot|s)-\pi_i^{t}(\cdot|s)\rVert_2^2) \}$. Deriving from the optimality we have:
    \begin{equation}\label{eq:18}
        \langle\beta\hat{q}^{\pi_t}_i(s,\cdot)-\pi_i^{t+1}(\cdot|s)+\pi_i^t(\cdot|s),p(\cdot|s)-\pi_i^{t+1}(\cdot|s)\rangle_{\mathcal{A}_i}\leq 0,\ \forall p(\cdot|s)\in\Delta_{\alpha}(\mathcal{A}_i).
    \end{equation}

    To bound $\text{Diff}_1$,
    \begin{equation*}
    \begin{aligned}
        \text{Diff}_1 =& \sum_{i=1}^N\rho_i^{\pi_i^{t+1},\pi_{-i}^t} - \rho_i^{\pi_i^{t},\pi_{-i}^t}\\
        =& \sum_{i=1}^N\mathbb{E}_{s\sim\nu^{\pi_i^{t+1}, \pi^t_{-i}}}[\langle\overline{Q}_i^{\pi^t}(s,\cdot), \pi^{t+1}_i(\cdot|s)-\pi^t_i(\cdot|s)\rangle_{\mathcal{A}_i}] \\
        =&\sum_{i=1}^N\mathbb{E}_{s\sim\nu^{\pi_i^{t+1}, \pi^t_{-i}}}[\langle \overline{Q}_i^{\pi^t}(s,\cdot)+N_1\rho_i^{\pi^t}, \pi^{t+1}_i(\cdot|s)-\pi^t_i(\cdot|s)\rangle_{\mathcal{A}_i}] \\
        =& \sum_{i=1}^N\mathbb{E}_{s\sim\nu^{\pi_i^{t+1}, \pi^t_{-i}}}[\langle\hat{q}_i^{\pi^t}(s,\cdot), \pi^{t+1}_i(\cdot|s)-\pi^t_i(\cdot|s)\rangle_{\mathcal{A}_i}]\\
        &+\sum_{i=1}^N\mathbb{E}_{s\sim\nu^{\pi_i^{t+1}, \pi^t_{-i}}}[\langle q^{\pi^t}_i(s,a_i)-\hat{q}_i^{\pi^t}(s,\cdot),\pi^{t+1}_i(\cdot|s)-\pi^t_i(\cdot|s)\rangle_{\mathcal{A}_i}] \\
        \overset{(a)}{\geq}& \frac{1}{\beta}\sum_{i=1}^N\mathbb{E}_{s\sim\nu^{\pi_i^{t+1}, \pi^t_{-i}}} \lVert \pi_i^{t+1}(\cdot|s)-\pi_i^t(\cdot|s)\rVert_2^2-\sum_{i=1}^N\mathbb{E}_{s\sim\nu^{\pi_i^{t+1}, \pi^t_{-i}}} \frac{\lVert\hat{q}_i^{\pi^t}(s,\cdot)-q_i^{\pi^t}(s,\cdot)\rVert_{2}^2+\lVert\pi_i^{t+1}(\cdot|s)-\pi_i^t(\cdot|s)\rVert_2^2}{2}\\
        =&(\frac{1}{\beta}-\frac{1}{2})\sum_{i=1}^N\mathbb{E}_{s\sim\nu^{\pi_i^{t+1}, \pi^t_{-i}}} \lVert \pi_i^{t+1}(\cdot|s)-\pi_i^t(\cdot|s)\rVert_2^2-\frac{1}{2}\sum_{i=1}^N\mathbb{E}_{s\sim\nu^{\pi_i^{t+1}, \pi^t_{-i}}}[\sum_{a_i} (\hat{q}_i^{\pi^t}(s,a_i)-q_i^{\pi^t}(s,a_i))^2].\\
    \end{aligned}
    \end{equation*}
    
    (a) is derived by applying $p=\pi^t_i$ to equation (27).
\end{proof}

\begin{theorem}
    If all players run \cref{alg:6} independently and synchronously, $\beta\leq (1+\frac{2L_{\Phi}}{1-\Gamma})^{-1}$, the Nash regret will be bounded by: 
    \begin{equation*}
        \mathbb{E}[\textnormal{Nash-regret}(T)]\leq(\frac{2}{\sqrt{\beta}}+\kappa\sqrt{A_{\max}\beta})\sqrt{\frac{2ND}{T}}+(\frac{2}{\sqrt{\beta}}+\kappa\sqrt{\beta})D\sqrt{ NA_{\max}\delta }+2\sqrt{A_{\max}\delta}+2\kappa\alpha.
    \end{equation*}
\end{theorem}
\begin{proof}
    \begin{equation}
    \begin{aligned}\label{eq:19}
        &\text{Nash-gap}(t) \\
        =& \max_i(\max_{\pi_i'}\rho_i^{\pi_i',\pi_{-i}^t} - \rho_i^{\pi_i^t,\pi_{-i}^t}) \\
        \overset{(a)}{=}&\mathbb{E}_{s\sim\nu^{\pi_i, \pi^t_{-i}}}[\langle\overline{Q}_i^{\pi^t}(s,\cdot), \pi_i(\cdot|s)-\pi^t_i(\cdot|s)\rangle_{\mathcal{A}_i}] \\
        =&\mathbb{E}_{s\sim\nu^{\pi_i, \pi^t_{-i}}}[\langle\overline{Q}_i^{\pi^t}(s,\cdot), (1-\alpha)\pi_i(\cdot|s)+\alpha u_i(\cdot|s)-\pi^t_i(\cdot|s)\rangle_{\mathcal{A}_i}]+\alpha\mathbb{E}_{s\sim\nu^{\pi_i, \pi^t_{-i}}}[\langle\overline{Q}_i^{\pi^t}(s,\cdot), \pi_i(\cdot|s)-u_i(\cdot|s)\rangle_{\mathcal{A}_i}] \\
        \leq&\mathbb{E}_{s\sim\nu^{\pi_i, \pi^t_{-i}}}[\langle\overline{Q}_i^{\pi^t}(s,\cdot), (1-\alpha)\pi_i(\cdot|s)+\alpha u_i(\cdot|s)-\pi^t_i(\cdot|s)\rangle_{\mathcal{A}_i}]+2\alpha\kappa\\
        =&\mathbb{E}_{s\sim\nu^{\pi_i, \pi^t_{-i}}}[\langle\overline{Q}_i^{\pi^t}(s,\cdot), (1-\alpha)\pi_i(\cdot|s)+\alpha u_i(\cdot|s)-\pi^{t+1}_i(\cdot|s)\rangle_{\mathcal{A}_i}]+\mathbb{E}_{s\sim\nu^{\pi_i, \pi^t_{-i}}}[\langle\overline{Q}_i^{\pi^t}(s,\cdot), \pi_i^{t+1}(\cdot|s)-\pi^{t}_i(\cdot|s)\rangle_{\mathcal{A}_i}]\\
        &+2\alpha\kappa\\
        =&\mathbb{E}_{s\sim\nu^{\pi_i, \pi^t_{-i}}}[\langle\hat{q}^{\pi^t}_i(s,\cdot), (1-\alpha)\pi_i(\cdot|s)+\alpha u_i(\cdot|s)-\pi^{t+1}_i(\cdot|s)\rangle_{\mathcal{A}_i}]\\
        &+\mathbb{E}_{s\sim\nu^{\pi_i, \pi^t_{-i}}}[\langle q^{\pi^t}_i(s,\cdot)-\hat{q}^{\pi^t}_i(s,\cdot), (1-\alpha)\pi_i(\cdot|s)+\alpha u_i(\cdot|s)-\pi^{t+1}_i(\cdot|s)\rangle_{\mathcal{A}_i}]\\
        &+\mathbb{E}_{s\sim\nu^{\pi_i, \pi^t_{-i}}}[\langle\overline{Q}_i^{\pi^t}(s,\cdot), \pi_i^{t+1}(\cdot|s)-\pi^{t}_i(\cdot|s)\rangle_{\mathcal{A}_i}]+2\alpha\kappa.\\
    \end{aligned}   
    \end{equation}
    In $(a)$, we assume agent $i$ achieved the $\max_i$ and $\pi_i$ achieved $\max_{\pi_i'}$.
    
    By \cref{eq:18}:
    \begin{equation*}
        \langle\beta\hat{q}^{t}_i(s,\cdot)-\pi_i^{t+1}(\cdot|s)+\pi_i^t(\cdot|s),(1-\alpha)\pi_i(\cdot|s)+\alpha u_i(\cdot|s)-\pi_i^{t+1}(\cdot|s)\rangle_{\mathcal{A}_i}\leq 0,
    \end{equation*}
    we can bound the first term of \cref{eq:19} as:
    \begin{equation*}
    \begin{split}
        \langle\hat{q}^{\pi^t}_i(s,\cdot), (1-\alpha)\pi_i(\cdot|s)+\alpha u_i(\cdot|s)-\pi^{t+1}_i(\cdot|s)\rangle_{\mathcal{A}_i}\leq &\frac{1}{\beta}\langle \pi_i^{t+1}(\cdot|s)-\pi_i^t(\cdot|s), (1-\alpha)\pi_i(\cdot|s)+\alpha u_i(\cdot|s)-\pi_i^{t+1}(\cdot|s)\rangle_{\mathcal{A}_i}\\
        \leq &\frac{2}{\beta}\lVert \pi_i^{t+1}(\cdot|s)-\pi_i^t(\cdot|s)\rVert_2.
    \end{split}
    \end{equation*}
    The last inequality comes from $\lVert \pi'(\cdot|s)-\pi(\cdot|s)\rVert_2\leq\lVert \pi'(\cdot|s)-\pi(\cdot|s)\rVert_1\leq2$.

    Therefore, the Nash gap can be bounded as:
    \begin{align*}
        &\text{Nash-gap}(t)\\
        \overset{(a)}{\leq}& \mathbb{E}_{s\sim\nu^{\pi_i, \pi^t_{-i}}}[\frac{2}{\beta}\lVert \pi_i^{t+1}(\cdot|s)-\pi_i^t(\cdot|s)\rVert_2+2\lVert q^{\pi^t}(s,\cdot)-\hat{q}^{\pi^t}(s,\cdot)\rVert_{2}+\kappa\sqrt{A_{\max}}\lVert\pi_i^{t+1}(\cdot|s)-\pi^t_i(\cdot|s)\rVert_2+2\alpha\kappa]\\
        =&(\frac{2}{\beta}+\kappa\sqrt{A_{\max}})\mathbb{E}_{s\sim\nu^{\pi_i, \pi^t_{-i}}}\lVert\pi_i^{t+1}(\cdot|s)- \pi_i^t(\cdot|s) \rVert_2+2\mathbb{E}_{s\sim\nu^{\pi_i, \pi^t_{-i}}}\lVert q^{\pi^t}(s,\cdot)-\hat{q}^{\pi^t}(s,\cdot)\rVert_{2}+2\alpha \kappa\\   
        \leq&(\frac{2}{\beta}+\kappa\sqrt{A_{\max}})\sqrt{\lVert\frac{\nu^{\pi_i, \pi^t_{-i}}}{\nu^{\pi^t}}\rVert_{\infty}}\sum_{s}\sqrt{\nu^{\pi_i,\pi_{-i}^t}(s)\nu^{\pi^{t+1}_i,\pi_{-i}^t}(s)}\sqrt{\lVert\pi_i^{t+1}(\cdot|s)- \pi_i^t(\cdot|s) \rVert_2^2}\\
        &+2\mathbb{E}_{s\sim\nu^{\pi_i, \pi^t_{-i}}}\lVert q^{\pi^t}(s,\cdot)-\hat{q}^{\pi^t}(s,\cdot)\rVert_{2}+2\alpha \kappa\\
        \leq&(\frac{2}{\beta}+\kappa\sqrt{A_{\max}})\sqrt{D}\sum_{s}\sqrt{\nu^{\pi_i,\pi_{-i}^t}(s)}\sqrt{\nu^{\pi^{t+1}_i,\pi_{-i}^t}(s)\lVert\pi_i^{t+1}(\cdot|s)- \pi_i^t(\cdot|s) \rVert_2^2}\\
        &+2\mathbb{E}_{s\sim\nu^{\pi_i, \pi^t_{-i}}}\lVert q^{\pi^t}(s,\cdot)-\hat{q}^{\pi^t}(s,\cdot)\rVert_{2}+2\alpha \kappa.\\
    \end{align*}
    
    Applying \cref{lm:23} and the approximation error bound leads to
    \begin{align*}
         \mathbb{E}[\mathbb{E}_{s\sim\nu^{\pi_i, \pi^t_{-i}}}\lVert q^{\pi^t}(s,\cdot)-\hat{q}^{\pi^t}(s,\cdot)\rVert_{2}]\leq&\sqrt{\mathbb{E}[\mathbb{E}_{s\sim\nu^{\pi_i, \pi^t_{-i}}}\lVert q^{\pi^t}(s,\cdot)-\hat{q}^{\pi^t}(s,\cdot)\rVert_{2}^2]}\\
         \overset{(a)}{=}&\sqrt{\mathbb{E}[\sum_s \nu^{\pi_i, \pi^t_{-i}}(s)\mathbb{E}_t[ \sum_{a_i}(q^{\pi^t}(s,a_i)-\hat{q}^{\pi^t}(s,a_i))^2]] }\\
         \leq& \sqrt{A_{\max}\delta}.
    \end{align*}
    
    In (a) we can exchange $\mathbb{E}_t$ and $\sum_s\nu^{\pi_i, \pi^t_{-i}}(s)$ since $\pi_i$ only depends on $\pi^t$, i.e. $\pi_i$ only depends on $\mathcal{F}_{t-1}$.
    
    Sum over $t$:
    \begin{equation*}
        \begin{aligned}
        &\mathbb{E}\sum_{t=0}^{T-1}\text{Nash-gap}(t)\\
        \leq&(\frac{2}{\beta}+\kappa\sqrt{A_{\max}})\sqrt{D} \mathbb{E}[\sum_{t=0}^{T-1}\sum_{s}\sqrt{\nu^{\pi_i,\pi_{-i}^t}(s)}\sqrt{\nu^{\pi^{t+1}_i,\pi_{-i}^t}(s)\lVert\pi_i^{t+1}(\cdot|s)- \pi_i^t(\cdot|s) \rVert_2^2}]+2\sqrt{A_{\max}\delta}T+2\alpha\kappa T\\
        \overset{(a)}{\leq}& (\frac{2}{\beta}+\kappa\sqrt{A_{\max}})\sqrt{D} \mathbb{E}\sqrt{\sum_{t=0}^{T-1}\sum_{s}\nu^{\pi_i,\pi_{-i}^t}(s)}\sqrt{\sum_{t=0}^{T-1}\sum_{i}\sum_{s}\nu^{\pi^{t+1}_i,\pi_{-i}^t}(s)\lVert\pi_i^{t+1}(\cdot|s)- \pi_i^t(\cdot|s) \rVert_2^2}\\
        &+2\sqrt{A_{\max}\delta}T+2\alpha\kappa T\\
        \overset{(b)}{\leq}&(\frac{2}{\beta}+\kappa\sqrt{A_{\max}})\sqrt{DT} \mathbb{E}\sqrt{\left(2\beta\sum_{t=0}^{T-1}(\Phi(\pi^{t+1}) - \Phi(\pi^t) )+\beta\sum_{t=0}^{T-1}\sum_{i=1}^N\sum_{a_i} \mathbb{E}_{s\sim\nu^{\pi_i^{t+1}, \pi^t_{-i}}}(\hat q_i^{\pi^t}(s,a_i)-q_i^{\pi^t}(s,a_i))^2\right)}\\
        &+2\sqrt{A_{\max}\delta} T+2\alpha\kappa T\\
        \leq&(\frac{2}{\beta}+\kappa\sqrt{A_{\max}})\sqrt{DT}(\sqrt{2\beta C_{\Phi}}+\mathbb{E}\sqrt{\beta\sum_{t=0}^{T-1}\sum_{i=1}^N\sum_{a_i} \mathbb{E}_{s\sim\nu^{\pi_i^{t+1}, \pi^t_{-i}}}(\hat q_i^{\pi^t}(s,a_i)-q_i^{\pi^t}(s,a_i))^2})\\
        &+2\sqrt{A_{\max}\delta}T+2\alpha\kappa T\\ 
        \overset{(c)}{\leq}&(\frac{2}{\beta}+\kappa\sqrt{A_{\max}})\sqrt{DT}(\sqrt{2\beta C_{\Phi}}+\sqrt{\beta\mathbb{E}\sum_{t=0}^{T-1}\sum_{i=1}^N\sum_{a_i} \mathbb{E}_{s\sim\nu^{\pi_i^{t+1}, \pi^t_{-i}}}(\hat q_i^{\pi^t}(s,a_i)-q_i^{\pi^t}(s,a_i))^2})\\
        &+2\sqrt{A_{\max}\delta}T+2\alpha\kappa T\\ 
        \leq& (\frac{2}{\beta}+\kappa\sqrt{A_{\max}})\sqrt{DT}(\sqrt{2\beta C_{\Phi}}+\sqrt{\beta TNA_{\max}\delta })+2\sqrt{A_{\max}\delta}T+2\alpha\kappa T,\\ 
        \end{aligned}
    \end{equation*}
    where (a) is due to Cauchy-Schwarz inequality. (b) is derived by \cref{lm:23} and learning rate $\beta\leq (1+\frac{2L_{\Phi}}{1-\Gamma})^{-1}$. Note that $\Phi(\pi^T)-\Phi(\pi^0)\leq C_{\Phi}$. (c) is Jensen's inequality.
    \begin{align*}
        \mathbb{E}[\text{Nash-regret}(T)]\leq(\frac{2}{\sqrt{\beta}}+\kappa\sqrt{A_{\max}\beta})\sqrt{\frac{2C_{\Phi}D}{T}}+(\frac{2}{\sqrt{\beta}}+\kappa\sqrt{A_{\max}\beta })\sqrt{ NA_{\max}\delta }+2\sqrt{A_{max}\delta}+2\kappa\alpha.
    \end{align*}
\end{proof}

If $\beta=(1+\frac{2L_{\Phi}}{1-\Gamma})^{-1}$, $\mathbb{E}[\text{Nash-regret}(T)]=O(\sqrt{\frac{C_{\Phi}NDA_{\max}S^{3/2}\kappa_0^2}{(1-\Gamma)T}}+\sqrt{\frac{N^2A^2_{\max}S^{3/2}\kappa_0^2\delta}{1-\Gamma}}+\kappa\alpha)$. To obtain an $\epsilon$-Nash equilibrium, let $T=O(\frac{C_{\Phi}NDA_{\max}S^{3/2}\kappa_0^2}{(1-\Gamma)\epsilon^2})$, $\delta=O(\frac{(1-\Gamma)\epsilon^2}{N^2A^2_{\max}S^{3/2}\kappa_0^2})$ and $\alpha=O(\frac{\epsilon}{\kappa})$. By \cref{lm:wei}, $B=\tilde O(\frac{1}{\alpha\delta})=\tilde{O}(\frac{N^2A^2_{\max}S^2\kappa_0^3}{(1-\Gamma)^{3/2}\epsilon^3})$. Therefore, the sample complexity for \cref{alg:6} is $TB=\tilde{O}(\frac{C_{\Phi}N^3DA^3_{\max}S^{7/2}\kappa_0^5}{(1-\Gamma)^{5/2}\epsilon^5})$. Comparing with sample complexity of \cref{alg:2} in \cref{thm:2}, the sample complexity of \cref{alg:6} is $O(\frac{A_{\max}^2S^{7/2}\kappa_0^4}{1-\Gamma})$ smaller.

\section{PROOFS OF SECTION 5}\label{D}
\begin{lemma}
    $\frac{\partial\rho_i^{\pi_{\theta}}}{\partial\theta_{s,a_i}}=\nu^{\pi_{\theta}}(s)\pi_{\theta_i}(a_i|s)\overline{A}_i^{\pi_{\theta}}(s,a_i)$.
\end{lemma}
\begin{proof}
    From the policy gradient theorem \citep{sutton1999policy}, $ \frac{\partial\rho_i^{\pi_{\theta}}}{\partial\theta_{s,a_i}}=\sum_{s',\mathbf{a}'}\nu^{\pi_{\theta}}(s)\pi_{\theta}(\mathbf{a}|s)\frac{\partial\log\pi_{\theta}(\mathbf{a}'|s')}{\partial\theta_{s,a_i}}Q_i^{\pi_{\theta}}(s,\mathbf{a})$, while $\pi_{\theta}(\mathbf{a}|s)=\Pi_{i=1}^N\pi_{\theta_i}(a_i|s)$ and $\frac{\partial \log\pi_{\theta}(\mathbf{a}'|s)}{\partial \theta_{s,a_i}}=\mathbf{1}\{a_i'=a_i,s'=s\}-\mathbf{1}\{s'=s\}\pi_{\theta_i}(a_i|s)$. Therefore, $\frac{\partial\rho_i^{\pi_{\theta}}}{\partial\theta_{s,a_i}}=\nu^{\pi_{\theta}}(s)\pi_{\theta_i}(a_i|s)\overline{A}_i^{\pi_{\theta}}(s,a_i)$.
\end{proof}

\begin{lemma}[Lemma 9-12 \cite{zhang2022global}]
    The update rule $\theta_i^{t+1}=\theta_i^t+\beta F_i(\theta^t)^{\dagger}\nabla_{\theta_i}\rho_i^{\pi_t}$ is equivalent to $\pi_i^{t+1}(a_i|s)\varpropto \pi_i^t(a_i|s)\exp\left( \beta \overline{A}_i^{\pi^t}(s,a_i)\right) $, where $F_i(\theta)=\mathbb{E}_{s\sim\nu^{\pi_{\theta}}}\mathbb{E}_{a_i\sim\pi_{\theta_i}(\cdot|s)}[\nabla_{\theta_i}\log\pi_{\theta_i}(a_i|s)\nabla_{\theta_i}\log\pi_{\theta_i}(a_i|s)^T]$ and $\pi_i^t=\pi_{\theta_i^t}$
\end{lemma}

\begin{lemma}[Policy improvement, \cref{lm:8}]\label{lm:26}
If $\beta\leq\max\{\frac{1-\Gamma}{(N-1)(\kappa_Q+S\kappa^2)},\frac{1-\Gamma}{2L_{\Phi}} \}$, the policies updated by \cref{alg:5} have:
    \begin{equation*}
    \begin{aligned}
        \Phi(\pi^{t+1}) - \Phi(\pi^t) \geq \frac{1}{\beta}\sum_i\mathbb{E}_{s\sim\nu^{\pi_i^{t+1},\pi_i^t}}\log Z_i^{t,s}.
    \end{aligned}
    \end{equation*}
\end{lemma}
\begin{proof}
    \begin{equation*}
    \begin{aligned}
        \Phi^{t+1}-\Phi^t=&\underbrace{\sum_{i=1}^N (\Phi(\pi^{t+1}_i,\pi^t_{-i})-\Phi(\pi^{t}_i,\pi^t_{-i}))}_{\text{Diff}_1}\\
        &+ \underbrace{\sum_{i=1}^N\sum_{j=i+1}^N (\Phi(\pi_{-(i,j)}^{t,t+1},\pi_i^{t+1},\pi_j^{t+1})-\Phi(\pi_{-(i,j)}^{t,t+1},\pi_i^t,\pi_j^{t+1}) - \Phi(\pi_{-(i,j)}^{t,t+1},\pi_i^{t+1},\pi_j^t)+\Phi(\pi_{-(i,j)}^{t,t+1},\pi_i^t,\pi_j^t))}_{\text{Diff}_2},
    \end{aligned}
    \end{equation*}
    (where $\pi_{-(i,j)}^{t,t+1}:=(\pi_{1: (i-1)}^t,\pi_{(i+1): (j-1)}^t, \pi_{(j+1): N}^{t+1}$)\ ).\\
    
    Similar to \cref{lm:21},
    \begin{equation}
    \begin{aligned}
        &\Phi(\pi_{-(i,j)}^{t,t+1},\pi_i^{t+1},\pi_j^{t+1})-\Phi(\pi_{-(i,j)}^{t,t+1},\pi_i^t,\pi_j^{t+1}) - \Phi(\pi_{-(i,j)}^{t,t+1},\pi_i^{t+1},\pi_j^t)+\Phi(\pi_{-(i,j)}^{t,t+1},\pi_i^t,\pi_j^t) \\
        \geq & -\frac{\kappa_Q+S\kappa^2}{2}(\lVert\pi_j^{t+1}-\pi_j^t\rVert_{1,\infty}^2+\lVert\pi_i^{t+1}-\pi_i^t\rVert_{1,\infty}^2).\\
    \end{aligned}
    \end{equation} 
    
    Let $s_i\in \arg\max_{s\in\mathcal{S}}\lVert\pi_i^{t+1}(\cdot|s)-\pi^t_i(\cdot|s)\rVert_1$. By Pinsker's inequality, we get:
    \begin{equation}
        \text{Diff}_2\geq -\frac{(N-1)(\kappa_Q+S\kappa^2)}{2}\sum_{i=1}^N\lVert\pi_i^{t+1}-\pi_i^t\rVert_{1,\infty}^2\geq -(N-1)(\kappa_Q+S\kappa^2)\sum_{i=1}^ND^{\pi_i^{t+1}}_{\pi_i^t}(s_i).
    \end{equation}
    
    Define $Z_t^{i,s}=\sum_{a_i}\pi_i^t(a_i|s)\exp\left(\beta \overline{A}_i^t(s,a_i)\right)$. By the update rule $\pi_i^{t+1}(a_i|s)=\frac{\pi_i^t(a_i|s)\exp\left(\beta \overline{A}_i^t(s,a_i)\right)}{Z_t^{i,s}}$, $\overline{A}_i^t(s,a_i)=\frac{1}{\beta}(\log(\frac{\pi_i^{t+1}(a_i|s)}{\pi_i^t(a_i|s)})+\log Z_i^{t,s})$. Hence
    
    \begin{align*}
        \Phi(\pi^{t+1}_i,\pi^t_{-i})-\Phi(\pi^{t}_i,\pi^t_{-i})=&\rho_i^{\pi_i^{t+1},\pi_i^t}-\rho_i^{\pi_i^t,\pi_{-i}^{t}}\\
        =&\mathbb{E}_{s\sim\nu^{\pi_i^{t+1},\pi_{-i}^t}}\sum_{a_i}\pi_i^{t+1}(a_i|s)\overline{A}_{i}^{\pi^t}(s,a_i)\\
        =&\mathbb{E}_{s\sim\nu^{\pi_i^{t+1},\pi_{-i}^t}}\sum_{a_i}\pi_i^{t+1}(a_i|s)\frac{1}{\beta}(\log\frac{\pi_i^{t+1}(a_i|s)}{\pi_i^t(a_i|s)}+\log Z_i^{t,s})\\
        =&\frac{1}{\beta}\mathbb{E}_{s\sim\nu^{\pi_i^{t+1},\pi_{-i}^t}}(D_{\pi_i^t}^{\pi_i^{t+1}}(s)+\sum_{a_i}\pi_i^{t+1}(a_i|s)\log Z_i^{t,s})\\
        \overset{(a)}{\geq}&\frac{1-\Gamma}{\beta} D_{\pi_i^t}^{\pi_i^{t+1}}(s_i)+\frac{1}{\beta}\mathbb{E}_{s\sim\nu^{\pi_i^{t+1},\pi_{-i}^t}}\log Z_i^{t,s}.
        \end{align*}
        Therefore, 
        \begin{align*}
        \text{Diff}_1\geq&\frac{1-\Gamma}{\beta}\sum_iD_{\pi_i^t}^{\pi_i^{t+1}}(s_i)+\frac{1}{\beta}\sum_i\mathbb{E}_{s\sim\nu^{\pi_i^{t+1},\pi_{-i}^t}}\log Z_i^{t,s},\\
    \end{align*}
    where (a) comes from $1-\Gamma\leq\nu^{\pi_i^{t+1},\pi_{-i}^t}(s_i)\leq\Gamma$ for any $i$.
    
    If the learning rate is chosen as $\beta\leq\frac{1-\Gamma}{(N-1)(\kappa_Q+S\kappa^2)}$, then combining the above results:
    \begin{equation}\label{eq:20}
        \begin{aligned}
        \Phi(\pi^{t+1})-\Phi(\pi^t)\geq&(\frac{1-\Gamma}{ \beta}-(N-1)(\kappa_Q+S\kappa^2))\sum_{i=1}^N D_{\pi_i^{t+1}}^{\pi_i^t}(s_i)+\frac{1}{\beta}\sum_i\mathbb{E}_{s\sim\nu^{\pi_i^{t+1},\pi_i^t}}\log Z_i^{t,s}\\
        \geq&\frac{1}{\beta}\sum_i\mathbb{E}_{s\sim\nu^{\pi_i^{t+1},\pi_i^t}}\log Z_i^{t,s}.\\
        \end{aligned}
    \end{equation}

    If we use \cref{lm:22} to bound each term in $\text{Diff}_2$, then by $\lVert x\rVert_2^2\leq\lVert x\lVert_1^2$ and Pinsker's inequality:
    \begin{equation*}
        \text{Diff}_2 \geq -L_{\Phi}\sum_i \sum_s\lVert\pi_i^{t+1}(\cdot|s)-\pi_i^t(\cdot|s)\rVert_2^2\geq-\frac{2L_{\Phi}}{1-\Gamma}\sum_i\mathbb{E}_{s\sim\nu^{\pi_i^{t+1}, \pi^t_{-i}}} D^{\pi_i^{t+1}}_{\pi_i^{t}}(s).
    \end{equation*}
    A similar result for \cref{eq:20} can be derived:
    \begin{align*}
        \Phi(\pi^{t+1})-\Phi(\pi^t)=&\text{Diff}_1+\text{Diff}_2\\
        \geq&\frac{1}{\beta}\sum_i\mathbb{E}_{s\sim\nu^{\pi_i^{t+1},\pi_{-i}^t}}(D_{\pi_i^t}^{\pi_i^{t+1}}(s)+\log Z_i^{t,s})-\frac{2L_{\Phi}}{1-\Gamma}\sum_i\mathbb{E}_{s\sim\nu^{\pi_i^{t+1}, \pi^t_{-i}}} D^{\pi_i^{t+1}}_{\pi_i^{t}}(s)\\
        \geq&\left(\frac{1}{\beta}-\frac{2L_{\Phi}}{1-\Gamma}\right)\sum_i\mathbb{E}_{s\sim\nu^{\pi_i^{t+1},\pi_{-i}^t}}D_{\pi_i^t}^{\pi_i^{t+1}}(s)+\frac{1}{\beta}\sum_i\mathbb{E}_{s\sim\nu^{\pi_i^{t+1},\pi_{-i}^t}}\log Z_i^{t,s}\\
        \geq&\frac{1}{\beta}\sum_i\mathbb{E}_{s\sim\nu^{\pi_i^{t+1},\pi_i^t}}\log Z_i^{t,s}.\\
    \end{align*}
    Therefore, if $\beta\leq\max\{\frac{1-\Gamma}{(N-1)(\kappa_Q+S\kappa^2)},\frac{1-\Gamma}{2L_{\Phi}} \}$, $\Phi(\pi^{t+1})-\Phi(\pi^t)\geq \frac{1}{\beta}\sum_i\mathbb{E}_{s\sim\nu^{\pi_i^{t+1},\pi_i^t}}\log Z_i^{t,s}$.
\end{proof}

With the above monotone improvement, the sufficient condition for asymptotic convergence established by \cite{zhang2022global} is satisfied.
\begin{lemma}[Section 12.0.2 \cite{zhang2022global}]\label{lm:27}
    If all stationary points of the potential function $\Phi(\theta)=\Phi(\pi(\theta))$ are isolated, $\beta\lVert A^{\pi}\rVert_{\infty}\leq 1$ for any $\pi$, and non-negative improvement $\Phi(\pi^{t+1})-\Phi(\pi^t)\geq \frac{1}{\beta}\sum_i\mathbb{E}_{s\sim\nu^{\pi_i^{t+1},\pi_i^t}}\log Z_i^{t,s}$ exists, algorithm \ref{alg:5} asymptotically converges to a Nash equilibrium. Also $c>0$.
\end{lemma}

Note that $\lVert A^{\pi}\rVert_{\infty}\leq\lVert Q^{\pi}\rVert_{\infty}+\lVert V^{\pi}\rVert_{\infty}\leq 2\kappa$ for any $\pi\in\Pi$. With \cref{lm:26} and \cref{lm:27} the following lemma holds.
\begin{lemma}[Restatement of \cref{lm:9}]
    If all stationary points of the potential function $\Phi(\theta)=\Phi(\pi(\theta))$ are isolated, $\beta\leq\max\{\frac{1-\Gamma}{(N-1)(\kappa_Q+S\kappa^2)},\frac{1-\Gamma}{2L_{\Phi}} \}$ and $\beta\leq\frac{1}{2\kappa}$, \cref{alg:5} asymptotically converges to a Nash equilibrium. Also $c>0$.
\end{lemma}

\begin{lemma}[Lemma 21 \cite{zhang2022global}]
    When $\beta\lVert A^{\pi}\rVert_{\infty}\leq 1$ for any $\pi$, $\log Z_t^{i,s}\geq \frac{c}{3}(\beta \max_{a_i}\overline{A}_i^{\pi_t}(s,a_i))^2$.
\end{lemma}

\begin{theorem}[Restatement of \cref{thm:4}]
    If all stationary points of the potential function $\Phi(\theta)=\Phi(\pi(\theta))$ are isolated, when $\beta\leq\max\{\frac{1-\Gamma}{(N-1)(\kappa_Q+S\kappa^2)},\frac{1-\Gamma}{2L_{\Phi}} \}$ and $\beta\leq\frac{1}{2\kappa}$, the regret of \cref{alg:5} is bounded 
    $$\textnormal{Nash-regret}^*(T)\leq\frac{3C_{\Phi}}{c\beta(1-\Gamma) T}.$$
\end{theorem}
\begin{proof}
    \begin{align*}
        \text{Nash-gap}(t) &= \max_i(\max_{\pi_i'}\rho_i^{\pi_i',\pi_{-i}^t} - \rho_i^{\pi_i^t,\pi_{-i}^t}) \\
        &\overset{(a)}{=}\mathbb{E}_{s\sim\nu^{\pi_i, \pi^t_{-i}}}\sum_{a_i}\pi_i(a_i|s)\overline{A}_i^{\pi^t}(s,a_i) \\
        &\leq \max_{s,a_i}\overline{A}_i^{\pi^t}(s,a_i).
    \end{align*}
    In $(a)$, assume $i$ attains the maximum in $\max_i$ and $\pi_i$ belongs $\arg\max_{\pi'_i}$. Apply the result in \cref{lm:26}:
    \begin{align*}
        \Phi(\pi^{t+1})-\Phi(\pi^t)\geq&\frac{1}{\beta}\sum_i\mathbb{E}_{s\sim\nu^{\pi_i^{t+1},\pi_i^t}}\log Z_i^{t,s}\\
        \geq&\frac{1-\Gamma}{\beta}\max_s\frac{c}{3}(\beta \max_{a_i}\overline{A}_i^{\pi^t}(s,a_i))^2
        \geq\frac{c\beta(1-\Gamma)}{3}\text{Nash-gap}(t)^2.
        \end{align*}        
    Therefore, we have
    \begin{align*}
    \frac{\Phi(\pi^T)-\Phi(\pi^0)}{T}\geq&\frac{c\beta(1-\Gamma)}{3}\text{Nash-regret}^*(T).
    \end{align*}
    Note that $\Phi(\pi^T)-\Phi(\pi^0)\leq C_{\Phi}$, the desired result can be shown.
\end{proof}

\end{document}